\newif\ifarxiv
\arxivfalse

\documentclass[final]{cvpr}

\usepackage{epsfig}
\usepackage{graphicx}
\usepackage{amsmath}
\usepackage{amssymb}

\usepackage{times}
\usepackage{epsfig}
\usepackage{graphicx}
\usepackage{amsmath}
\usepackage{amssymb}
\usepackage{booktabs}
\usepackage{multirow}
\usepackage{color,colortbl}
\usepackage{framed}
\usepackage{expl3}
\usepackage{caption}
\usepackage{placeins} 
\usepackage{palatino}

\usepackage{collcell}

\usepackage[pagebackref=true,breaklinks=true,colorlinks,bookmarks=false]{hyperref}

\def\cvprPaperID{5395} 
\def\confYear{CVPR 2021}

\definecolor{mygray}{gray}{0.6}

\newcommand{\stdminus}[1]{ \scalebox{0.65}{$\pm #1$}}
\def \ours   {Grafit\xspace}
\def \knn   {kNN\xspace}
\definecolor{orange}{rgb}{1.0, 0.49, 0.0}

\usepackage{enumitem}
\setlist[itemize]{%
labelsep=5pt,%
labelindent=0.4\parindent,%
itemindent=0pt,%
leftmargin=*,%
itemsep=-1pt, 
}

\makeatletter
\renewcommand{\paragraph}{%
  \@startsection{paragraph}{4}%
  {\z@}{2.0ex \@plus 1ex \@minus .2ex}{-1em}%
  {\normalfont\normalsize\bfseries\itshape}%
}

\newcommand*{\MinNumber}{0.0}%
\newcommand*{\MidNumber}{80.0} %
\newcommand*{\MidNumberB}{95.0} %
\newcommand*{\MaxNumber}{100.0}%

\definecolor{MinColor}{rgb}{1.0,1.0,1.0}
\definecolor{MidColor}{rgb}{0.7,0.7,0.7}
\definecolor{MidColorB}{rgb}{1.0,0.5,0.5}
\definecolor{MaxColor}{rgb}{1.0,0.2,0.2}

\newcommand{\ApplyGradient}[1]{\hspace{-0.33em}%
        \ifdim #1 pt > \MidNumberB pt
            \pgfmathsetmacro{\PercentColor}{max(min(100.0*(#1 - \MidNumberB)/(\MaxNumber-\MidNumberB),100.0),0.00)} %
            \colorbox{MaxColor!\PercentColor!MidColorB}{\makebox[3.5em]{#1}}%
        \else \ifdim #1 pt > \MidNumber pt
            \pgfmathsetmacro{\PercentColor}{max(min(100.0*(#1 - \MidNumber)/(\MidNumberB-\MidNumber),100.0),0.00)} %
            \colorbox{MidColorB!\PercentColor!MidColor}{\makebox[3.5em]{#1}}%
        \else
            \pgfmathsetmacro{\PercentColor}{max(min(100.0*(\MidNumber - #1)/(\MidNumber-\MinNumber),100.0),0.00)} %
            \colorbox{MinColor!\PercentColor!MidColor}{\makebox[3.5em]{#1}}%
        \fi \fi
}

\newcolumntype{R}{>{\collectcell\ApplyGradient}c<{\endcollectcell}}

\begin{document}

\title{\ours: Learning fine-grained image representations with coarse labels}

\author{Hugo Touvron$^{*,\dagger}$ 
\hspace{0.3cm} 
Alexandre Sablayrolles $^*$ 
\hspace{0.3cm} 
Matthijs Douze $^*$
\hspace{0.3cm} 
Matthieu Cord $^\dagger$
\hspace{0.3cm} 
Herv\'e J\'egou $^*$ \\[0.5cm]

\scalebox{1.}{$^*$ Facebook AI Research \hspace{0.6cm} $^\dagger$Sorbonne University}\\[18pt]
\hspace{-10pt} %
\begin{minipage}{1.00\linewidth}
{\centering \includegraphics[trim={0 20pt 0 0},clip,width=1.0\linewidth]{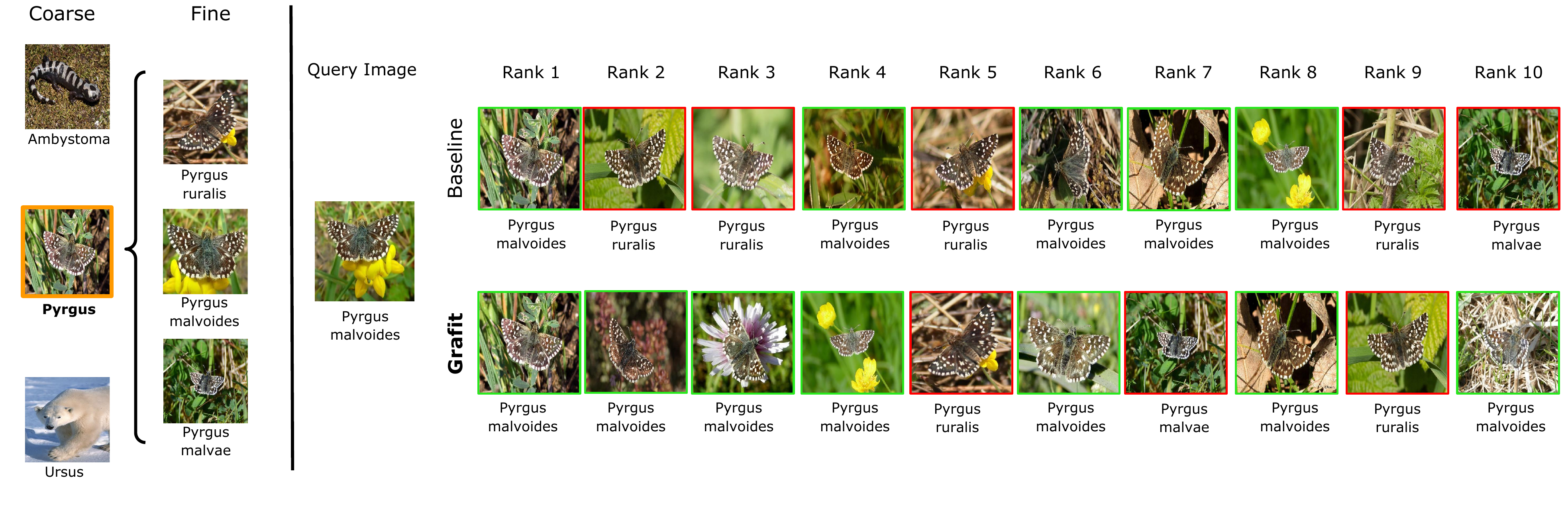}%

\vspace{-5pt}
\captionof{figure}{%
Category-level retrieval orders images based on their semantic similarity to a query. 
The \ours method, although it has used only coarse labels (like \textcolor{orange}{'pyrgus'}) at training time, produces a ranking consistent with fine-grained labels. 
Unsupervised learning is a particular case of this task, in which the set of coarse labels is reduced to a singleton. Image credit: \cite{inatimages_main}.
\label{fig:splash}}}
\vspace{-10pt}
\end{minipage}

}

\maketitle

\begin{abstract}
\vspace{-8pt}
This paper tackles the problem of learning a finer representation than the one provided by training labels. 
This enables fine-grained category retrieval of images in a collection annotated with coarse labels only.

Our network is learned with a nearest-neighbor classifier objective,
and an instance loss inspired by self-supervised learning.
By jointly leveraging the coarse labels and the underlying fine-grained latent space, it significantly improves the accuracy of category-level retrieval methods.  

Our strategy  outperforms all competing methods for retrieving or classifying images at a finer granularity than that available at train time. 
It also improves the accuracy for transfer learning tasks to fine-grained datasets, thereby establishing the new state of the art on five public benchmarks, like iNaturalist-2018. 
\end{abstract}

\def \coarsetocoarse {coarse-to-coarse\xspace}
\def \finetofine     {fine to fine\xspace}
\def \coarsetofine   {coarse to fine\xspace}
\def \finetocoarse   {fine to coarse\xspace}
\newcommand{\iminatevol}[1]{\includegraphics[width=0.08\linewidth]{figs/inat18_evolution/#1}}

\definecolor{kingdom}{rgb}{1.0, 0.0, 0.0}
\definecolor{phylum}{rgb}{0.0, 0.0, 0.0}
\definecolor{class}{rgb}{0.83, 0.37, 0.0}
\definecolor{order}{rgb}{0.9, 0.62, 0.0}
\definecolor{family}{rgb}{0.8, 0.47, 0.65}
\definecolor{genus}{rgb}{0.0, 0.45, 0.72}
\definecolor{species}{rgb}{0.0, 0.62, 0.45}

\vspace{-10pt}

\def \Sb {\textbf{\S}}

\section{Introduction}
\label{sec:introduction}

Image classification now achieves a performance that meets many application needs~\cite{He2016ResNet,Krizhevsky2012AlexNet,Touvron2020FixingTT}. 
In practice however, the dataset and labels available at training time do not necessarily correspond to those needed in subsequent applications~\cite{damour2020underspecification}.  
The granularity of the training-time concepts may not suffice for fine-grained downstream tasks. %

This has encouraged the development of specialized classifiers offering a more precise representation.
Fine-grained classification datasets~\cite{Horn2017INaturalist} have been developed for specific domains, for instance to distinguish different plants~\cite{Kiat2019Herbarium}  or bird species~\cite{WahCUB_200_2011}. 

Gathering a sufficiently large collection with fine-grained labels is difficult by itself, as it requires to find enough images of rare classes, and annotating them precisely requires domain specialists with in-domain expertise.
This is evidenced by the Open Images construction annotation protocol~\cite{Kuznetsova2020TheOI} that states that:
``\emph{Manually labeling a large number of images with the presence or absence of 19,794 different classes is not feasible}''.
For this reason they resorted to computer-assisted annotation, at the risk of introducing biases due to the assisting algorithm.
Being able to get strong classification and image retrieval performance on fine concepts using only coarse labels at training time can circumvents the issue, liberating the data collection process from the quirks of a rigid fine-grained taxonomy.

In this paper, our objective is to learn a finer-grained representation than that available at training time. 
This approach addresses the following use-cases:

    \paragraph{Category-level Retrieval.} Given a collection of images annotated with coarse labels, like a product catalog, we aim at ranking these images according to their fine-grained semantic similarity to a new query image outside the collection, as illustrated by Figure~\ref{fig:splash}.%
    
    \paragraph{On-the-fly classification.} For this task the fine-grained labels are available at test time only, and we use a non-parametric \knn classifier~\cite{wu2018improving} for on-the-fly classification, \ie without training on the fine-grained labels. \medskip

Our work leverages two intuitions. 
First, in order to improve the granularity beyond the one provided by image labels, we need to exploit another signal than just the labels. 
For this purpose, we build upon recent works~\cite{berman2019multigrain,Xie2019UnsupervisedDA} that exploits \emph{two} losses to address both image classification and instance recognition,
leveraging the ``free'' annotations provided by multiple data augmentations of a same instance, in the spirit of self-supervised learning~\cite{Bojanowski2017UnsupervisedLB,Caron2020UnsupervisedLO,Chen2020ASF,Grill2020BootstrapYO}.  

The second intuition is that it is best to explicitly infer coarse labels even when classifying for a finer granularity. 
For this purpose, we propose a simple method that exploits both a coarse classifier and image embeddings to improve fine-grained category-level retrieval. 
This strategy outperforms existing works that exploit coarse labels at training time but do not explicitly rely on them when retrieving finer-grained concepts~\cite{wu2018improving}.

In summary, in this context of coarse-to-fine representation learning, our paper makes the following contributions: 
\begin{itemize}
\itemsep0.1em 

\item We propose a method that learns a representation at a finer granularity than the one offered by the annotation at training time. 
It exhibits a significant accuracy improvement on all the  coarse-to-fine tasks that we consider. For instance, we improve by \textbf{+16.3\%} the top-1 accuracy for on-the-fly classification on ImageNet. 
This improvement is still +9.5\% w.r.t. our own  stronger baseline, everything being equal otherwise.

\item Our approach performs similarly or better at the coarse level. 
A byproduct of our study is a very strong \knn-classifier on Imagenet: \ours with ResNet-50 trunk reaches \textbf{79.6\%} top-1 accuracy at resolution $224$$\times$$224$. 

\item \ours improves \textbf{transfer learning}: our experiments show that our representation discriminates better at a finer granularity. 
Everything being equal otherwise, fine-tuning our model for fine-grained benchmarks significantly improves the accuracy. 

\item As a result we establish the new state of the art on five public  benchmarks for transfer learning: Oxford Flowers-102~\cite{Nilsback08}, Stanford Cars~\cite{Cars2013}, Food101~\cite{bossard14Food101}, iNaturalist 2018~\cite{Horn2018INaturalist} \& 2019~\cite{Horn2019INaturalist}. 
 
\end{itemize}

This paper is organized as follows.
After reviewing related works in Section~\ref{sec:related}, we present our method in
Section~\ref{sec:method}.
Section~\ref{sec:experiments} compares our approach against baselines on various datasets, 
and presents an extensive ablation. %
Section~\ref{sec:conclusion} concludes the paper. 

\paragraph{In the supplemental material,}  Appendix~\ref{sec:disc_gran} summarizes two experiments that show how an instance-level loss improves the granularity beyond the one learned by a vanilla cross-entropy loss.  
Appendix~\ref{app:experiment} complements our experimental section~\ref{sec:experiments} with more detailed results. Appendix~\ref{app:visu} provides visual results associated with different levels of training/testing granularities. 


\section{Related work}
\label{sec:related}
\paragraph{Label granularity in image classification.} 
In computer vision, the concept of granularity underlies several tasks, such as fine-grained~\cite{Kiat2019Herbarium,Horn2017INaturalist} or hierarchical image classification~\cite{Deng2012HedgingYB,Wu2019AHL,Xie2015HyperclassAA}.  
Some authors consider a formal definition of granularity, see for instance Cui et al.~\cite{Cui2019MeasuringDG}. 
In our paper, we only consider levels of granularity relative to each other, where each coarse class is partitioned into a set of finer-grained classes.   

In some works on hierarchical image classification~\cite{Eshratifar2019Coarse2FineAT,Guo2017CNNRNNAL,Ristin2015FromCT,Taherkhani_2019_ICCV}, a coarse annotation is available for all training images, but only a subset of the training images are labelled at a fine granularity.
In this paper we consider the case where no fine labels at all are available at training time.

\paragraph{Train-Test granularity discrepancy.}
A few works consider the case where the test-time labels are finer than those available at training time and where each fine label belongs to one coarse label.
Approaches to this task are based on  clustering~\cite{wu2018improving}  or transfer learning~\cite{Huh2016WhatMI}. 
Huh et al.~\cite{Huh2016WhatMI} address the question:
``is the feature embedding induced by the coarse classification task capable of separating  finer labels (which it never saw at training)?''
To evaluate this, they consider the 1000 ImageNet classes as fine, and group them into 127 coarse classes with the WordNet~\cite{fellbaum98wordnet} hierarchy.
Wu et al.~\cite{wu2018improving} 
evaluate on the 20 coarse classes of CIFAR-100~\cite{Krizhevsky2009LearningML} and on the same subdivision of ImageNet into 127 classes.
They evaluate their method, Scalable Neighborhood Component Analysis (SNCA), with a \knn classifier  applied on features extracted from a network trained with coarse labels. 
Note that this work departs from the popular framework of object/category discovery~\cite{Cho2015UnsupervisedOD,Galleguillos2011FromRS,Hsu2016DeepIC,Vo2019UnsupervisedIM,Vo2020TowardUM}, which is completely unsupervised. 

In our work we mainly compare to the few works that consider coarse labels at train time, therefore SNCA~\cite{wu2018improving} is one of our baseline. We adopt their coarse labels definition and evaluation procedure for on-the-fly classification.

\paragraph{Unified embeddings for classes and instances.} 
Similar to Wu et al.~\cite{wu2018improving}, several Distance Metric Learning (DML) approaches like the Magnet loss~\cite{Rippel2016MetricLW} or ProxyNCA~\cite{MovshovitzAttias2017NoFD,Teh2020ProxyNCAplus} jointly take into account intra- and inter-class variability. 
This improves transfer learning performance and favors in some cases the emergence of finer hierarchical concepts.
Berman et al. proposed  Multigrain~\cite{berman2019multigrain}, which simply adds to the classification objective a triplet loss that pulls together different data-augmentations of a same image. 
Recent works on semi-supervised learning ~\cite{Berthelot2019ReMixMatchSL,Berthelot2019MixMatchAH,Sohn2020FixMatchSS,Xie2019UnsupervisedDA,Yalniz2019BillionscaleSL,Zhai2019S4LSS} rely on both supervised and self-supervised losses to get information from unlabelled data. 
For instance the approach of Xie et al.~\cite{Xie2019UnsupervisedDA} is similar to Multigrain, except that the Kullback-Leibler divergence replaces the triplet loss.
Matching embeddings of the same images under different data-augmentations is the main signal in current works on self-supervised learning, which we discuss now. 

\paragraph{Unsupervised and Self-Supervised Learning.} 
In unsupervised and self-supervised approaches~\cite{Caron2020UnsupervisedLO,Chen2020ASF,gidaris2020,Grill2020BootstrapYO,Ji2019InvariantIC,wvangansbeke2020learning} the model is trained on unlabeled data.
Each image instance is considered as a distinct class and the methods aim at making the embeddings of different data-augmentations of a same instance more similar than those of other images.
To deal with finer semantic levels than those provided by the labels, we use an approach similar to BYOL~\cite{Grill2020BootstrapYO}. 
BYOL only requires pairs of positive elements (no negatives), more specifically different augmentations of the same image.
A desirable consequence is that this limits contradictory signals on the classification objective. 

\paragraph{Transfer Learning.}
Transfer learning datasets~\cite{bossard14Food101,Cars2013,Nilsback08} are often fine grained and rely on a feature extractor pre-trained on another set of classes.
However, the fine labels are not a subset of the pre-training labels, so we consider transfer learning as a generalization of our coarse-to-fine task.
It is preferable to pre-train on a domain similar to the target~\cite{Cui2018LargeSF}, \textit{e.g.,} pre-training on iNaturalist~\cite{Horn2017INaturalist}  is preferable to pre-training on ImageNet if the final objective is to discriminate between species of birds.
The impact of pre-training granularity is discussed in prior works \cite{Cui2019MeasuringDG,Yan2019ClusterFitIG}. 
In Section~\ref{subsec:transfer_learning_task} we investigate how \ours pre-training performs on fine-grained transfer learning datasets.

%
\section{\ours: Fitting a finer Granularity}
\label{sec:method}

\begin{figure}[t]
\centering
\includegraphics[width=1.0\linewidth]{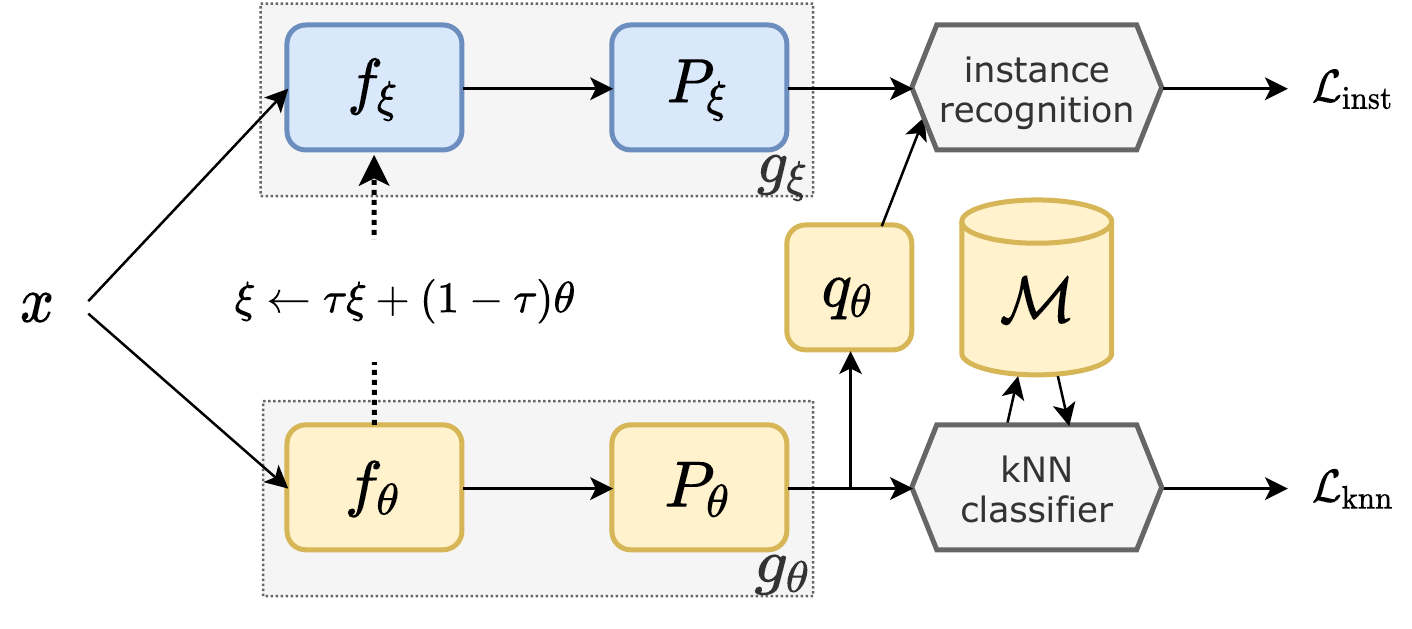}

\vspace{-5pt}
\caption{\label{fig:our_method}
 Illustration of our method at train time. 
 The convnet trunk that receives gradient is $f_\theta$ and is used to update the target network $f_\xi$ as a moving average. 
The database of neighbors is updated by averaging embedding in each mini-batch with corresponding embeddings in the database. 
}
\end{figure}

Figure~\ref{fig:our_method} depicts our approach at training time. 
In this section we discuss the different components and training losses. 
Then we detail how we produce the category-level ranking, and how we perform on-the-fly classification.

\subsection{Training procedure: \ours and \ours FC}

We first introduce an instance loss inspired by  BYOL~\cite{Grill2020BootstrapYO} that favors fine-grained recognition. 
The \ours model includes a trunk network $f_{\theta}$, to which we add two multi-layers perceptrons (MLP): a ``projector'' $P_{\theta}$ and a predictor $q_{\theta}$. 
In the \ours FC variant, $P_{\theta}$ is linear for a more direct fair comparison with Wu et al.~\cite{wu2018improving}'s  projector. %
The learnable parameters are represented by the vector $\theta$. 
As in BYOL we define a ``target network'' $f_{\xi}$ as an exponential moving average of the main network $f_{\theta}$:
the parameters $\xi$ are not learned, but computed as $\xi \leftarrow \tau \xi + (1-\tau)\theta$, with a target decay rate $\tau \in [0, 1]$.

\paragraph{Instance loss.}  
Each image $x$ is transformed by $T$ data augmentations $(t_1,\dots, t_T)$.
Denoting $\cos$ the cosine similarity and $g_{\theta}(x) = P_{\theta} ( f_{\theta}  (x))$, 
the instance loss is:
\begin{equation}
\mathcal{L}_\mathrm{inst}(x) = -
    \sum_{1 \leq i \neq j \leq T}
    \frac{\cos\big( 
        q_{\theta} \circ g_{\theta} (t_{i}(x)). 
        g_{\xi}(t_{j}(x))}{T(T-1)}
    \big),
\end{equation}

\paragraph{\knn loss.} 
A parametric classifier with softmax yields a representation
that does not generalize naturally to new classes~\cite{wu2018improving} and is not adapted for \knn classification.
Therefore, inspired by the neighborhood component analysis~\cite{Goldberger2004NeighbourhoodCA,Min2010DeepST,Salakhutdinov2007LearningAN}, Wu et al.~\cite{wu2018improving} propose a loss function optimized directly for \knn evaluation, that we adopt and denote by  $\mathcal{L}_\mathrm{knn}$.
Let $x_i$ be a training image with coarse label $y_i$ and $\sigma$ a temperature hyper-parameter.
For each image $x_i$ we select $x_j(j \neq i)$ as its neighbor with probability $p_{i,j}$, computed as
\begin{equation}
p_{i,j} \propto \exp\big(\cos(g_{\theta}(x_i),g_{\theta}(x_j))/\sigma
\big), 
\label{eq:probaij}
\end{equation}
where the $p_{i,j}$ are normalized so that $\sum_{j\neq i}p_{i,j} = 1$. 
The loss is then defined as:
\begin{equation}
\mathcal{L}_\mathrm{knn}(x_i,y_i) = - \log{\sum_{j , y_j=y_i,j \neq i}p_{i,j}}. 
\label{eq:knnloss}
\end{equation}

We $\ell_2$-normalize after the $P_{\theta}$ projection. 
The $\mathcal{L}_\mathrm{knn}$ scores all classes with Equation~\ref{eq:knnloss}. 

\paragraph{Memory of embeddings.} 
One of the limitations of the kNN approach is that it requires to use all the features of the training set. 
 To avoid recomputing all the embeddings of the training set, we use a memory $\mathcal{M}=\{m_1,\dots,m_i,\dots\}$. 
It is updated as follows: when the image $x_i$ in the training set is in the current mini-batch, we update its embedding $m_i$ as follows: $m_i \leftarrow \frac{1}{2}(m_i + g_{\theta}(x_i))$. 
In order to limit the memory space needed, we apply the $\mathcal{L}_\mathrm{knn}$ loss  on the space of the projected features, which allows us to store smaller embedding and hence requires less memory. 
For instance for ImageNet we have to store 1.2M training images. Without the projection with ResNet-50 architecture for $f_\theta$, the memory size is $2048 \times 1.2M$ but with a projection on a space of size 256 the memory size is $256 \times 1.2M$ what is $\times 8$ smaller.

\paragraph{Combined loss.}
Our method is summarized in Figure~\ref{fig:our_method}.
The total loss at training time for an image $x$ with label $y$ is:
\begin{equation}
\mathcal{L}_\mathrm{tot}(x)= \mathcal{L}_\mathrm{knn}(g_{\theta}(x),y) +  \mathcal{L}_\mathrm{inst}(x).
\label{equ:addlosses}
\end{equation}

Appendix~\ref{app:experiment} empirically shows that weighting differently the losses does not bring much performance.

\paragraph{Adapting the architecture at test-time.} 
The training parameters include the model weights ($f_\theta, P_{\theta}$) and the parameters related to $\mathcal{L}_\mathrm{inst}$ ($f_\xi, P_\xi$ and $q_\theta$) as described previously. 
At test time we remove the $\mathcal{L}_\mathrm{inst}$ branch, keeping only $f_{\theta}$ and $P_{\theta}$. 
In order to have consistent representations of all the training images with the final weights, we re-compute $m_i=g_\theta(x_i)$ for each training image $x_i$ and store it in $\mathcal M$. 

\subsection{Category-level retrieval}
\label{sec:categoryretrieval}

For a given test image $x'$ the task is to order by semantic relevance all images from the training collection. 
In our coarse-to-fine case, a search result is deemed correct if it has the same fine label as the query. 

\paragraph{Cosine-based ranking. }
The standard strategy to order the images is to compute $g_\theta(x')$, and to order all images $x_i$ in the collection by they cosine similarity score $\cos(g_\theta(x_i),g_\theta(x'))$ to the query (the $g_\theta(x_i)$ are pre-computed in $\mathcal M$).
The experiments in Section~\ref{sec:experiments} show that the way \ours embeddings are trained already improves the ranking with that method. 

\paragraph{Ranking conditioned by coarse prediction. } 
Let $x'$ be a test image and $x$ a training image with coarse class $y$.
Let $p_\mathrm{c}(x,y)$ be the probability that the image $x$ has coarse label $y$ according to our classifier.
Our conditional score $\psi_\mathrm{cond}$ is a compromise between the embedding similarity and the coarse classification, in spirit of the loss in Equation~\ref{equ:addlosses}: 

\begin{equation}
\label{eq:cond_score}
    \psi_\mathrm{cond}(x',x) = 
    \cos\left(g_{\theta}(x'),g_{\theta}(x)\right)
    + 
    \log\left(\frac{p_\mathrm{c}(x',y)}{1-p_\mathrm{c}(x',y)}\right).
\end{equation}

Note that, in that case, we rely on the fact that the collection in which we search is the training set, so that the coarse labels associated with the collection are known. 
In Section~\ref{sec:experiments} we show experimentally that $\psi_\mathrm{cond}$ improves the category-level retrieval performance in the coarse-to-fine context.

\paragraph{Conditional ranking: Oracle.}
If we assume that the coarse label of the query test image is known (given by an oracle), then we can set $p_\mathrm{c}(x',y)=1_{y=y'}$ with $y'$ the coarse class of the test image $x'$.
This boils down to systematically putting images with the same coarse class as the test image first in the ranking. 
Experimentally, this shows the impact of test label prediction on the score, and provides an upper bound on the performance of the conditional ranking strategy. 
It is also relevant in practice in a scenario where the user provides this coarse labeling, for instance by selecting it from an interface. 

\subsection{On-the-fly classification}

In on-the-fly classification, a \knn classifier ``knows'' about the fine classes of the training images only at test time~\cite{wu2018improving}. 
Such a non-parametric classification does not require any training or fine-tuning. 
As a side note, this flexible classifier can handle settings with evolving datasets, including dynamic additions of new classes, although such setups are outside the scope of this paper.
For a test image $x$ we compute the embedding $g_\theta(x)$ and compare it to the training image embeddings stored in $\mathcal M$. 
We select the $k$ embeddings maximizing the cosine similarity to the query, $(x_1, ..., x_k)$, with labels $(y_1, ..., y_k)$. 
For a direct comparison with Wu et al.~\cite{wu2018improving} 
and  consistently with Equation~\ref{eq:knnloss}, 
we apply an exponentially decreasing neighbour weighting that computes the probability that $x$ belongs to class $y$ as
\begin{equation}
    p_\mathrm{\knn}(x, y) 
    \propto
    \sum_{j=1, y_j = y}^k
    \exp\left(
    \cos(g_\theta(x),g_\theta(x_j))/\sigma
    \right)
    .
\end{equation}
We normalize the probabilities so that $\sum_y p_\mathrm{\knn}(x, y)=1$.


\section{Experiments}
\label{sec:experiments} 

We consider evaluation scenarios where it is beneficial to learn at a finer granularity than that provided by the training labels.
The first two tasks are coarse-to-fine tasks (category-level retrieval and on-the-fly classification), where we measure the capacity of the network to discriminate fine labels without having seen them at training time. 
The third protocol is vanilla transfer learning, where we transfer from Imagenet to a fine-grained dataset.

\subsection{Datasets and evaluation metrics} 
\label{sec:dataset}

We carry out our evaluations on public benchmarks, which statistics are detailed in Table~\ref{tab:dataset_coarse_to_fine}.

\paragraph{ CIFAR-100~\cite{Krizhevsky2009LearningML}} has 100 classes grouped into 20 coarse concepts of 5 fine classes each. 
For instance the coarse class \emph{large carnivore} includes fine classes \emph{bear}, \emph{leopard}, \emph{lion}, \emph{tiger} and \emph{wolf}.
In all experiments, we use the coarse concepts to train our models and evaluate the trained model using the fine-grained labels.

\paragraph{ImageNet~\cite{Russakovsky2015ImageNet12}} classes follow the WordNet~\cite{fellbaum98wordnet} hierarchy. 
We use the 127 coarse labels defined in Huh et al.~\cite{Huh2016WhatMI} in order to allow for a direct comparison with their method.

\paragraph{iNaturalist-2018} offers 7 granularity levels from the most general to the most specific, that follow the biological taxonomy:
Kingdom (6 classes), Phylum (25 classes), Class (57 classes), Order (272 classes), Family (1,118 classes), Genus (4,401 classes) and Species (8,142 classes).  
We consider pairs of (coarse,fine) granularity levels in our experiments. 
\textbf{iNaturalist-2019} is similar to iNaturalist-2018 with fewer classes and images, and yields similar conclusions. 

\paragraph{Flowers-102, Stanford Cars and Food101} are fine-grained benchmarks with no provided coarse labelling. Therefore we can use them for the transfer learning task.  
\begin{table}[t]
\caption{Datasets used for our different tasks. The four top datasets offer two or more levels of granularity, we use them for all coarse-to-fine tasks. 
The bottom three are fine-grained datasets employed to evaluate transfer learning. \label{tab:dataset_coarse_to_fine}}

\centering
{\small
\begin{tabular}{l|rrr}
\toprule
Dataset & Train size & Test size & \#classes   \\
\midrule
CIFAR-100~\cite{Krizhevsky2009LearningML}  & 50,000    & 10,000 & 20/100   \\ 
ImageNet \cite{Russakovsky2015ImageNet12}  & 1,281,167 & 50,000 & 127/1000  \\ 
iNaturalist 2018~\cite{Horn2018INaturalist}& 437,513   & 24,426 & 6/\dots/8,142 \\ 
iNaturalist 2019~\cite{Horn2019INaturalist}& 265,240   & 3,003  & 6/\dots/1,010  \\ 
\midrule
Flowers-102~\cite{Nilsback08}& 2,040 & 6,149 & 102  \\ 
Stanford Cars~\cite{Cars2013}& 8,144 & 8,041 & 196  \\  
Food101~\cite{bossard14Food101}& 75,750 & 25,250 & 101  \\ 
\bottomrule
\end{tabular}}
\end{table}

\paragraph{Evaluation metrics.} For category-level retrieval we report the mean average precision (mAP), as commonly done for retrieval tasks~\cite{Babenko2014NeuralCF,Philbin2007Objectretrieval}. For on-the-fly classification we report the top-1 accuracy.

\subsection{Baselines}
\label{subsec:baseline}

We use existing baselines and introduce stronger ones:  

\paragraph{Wu's baselines~\cite{wu2018improving}}
    use activations of a network learned with cross-entropy loss, but evaluated with a \knn classifier.
    Huh et al.~\cite{Huh2016WhatMI} evaluate how a network trained on the 127 ImageNet coarse classes transfers on the 1000 fine labels\footnote{They fine-tune a linear classifier with fine labels. We do not consider this task in the body of the paper, but refer to Appendix~\ref{subsec:eval_fc}: our approach provides a significant improvement in this case as well. }.
    
 \paragraph{Our main baseline:} we learn a network with cross-entropy loss, and perform retrieval or \knn-classification with the $\ell_2$-normalized embedding produced by the model trunk. 
    We point out that, thanks to our strong optimization strategy borrowed from recent works~\cite{Tong2018BagofTricks,tan2019efficientnet}, this baseline by itself outperforms all published results in several settings, for instance our ResNet-50 baseline without extra training data outperforms on ImageNet a ResNet-50 pretrained on YFCC100M~\cite{Yalniz2019BillionscaleSL} (see Table~\ref{tab:resnet_50_baseline} in Appendix~\ref{app:experiment} for a comparison).

\paragraph{SNCA.} Wu et al.~\cite{wu2018improving} proposed SNCA, a model optimized with a \knn loss.
    In our implementation, we add a linear operator $P_{\theta}$ to  the network trunk $f_{\theta}$ when training the supervised loss $\mathcal{L}_\mathrm{knn}$.

\paragraph{SNCA+.} We improve SNCA with our stronger optimization procedure. 
The retrieval or \knn evaluation uses features from a MLP instead of a simple linear projector, which means that its number of parameters is on par with \ours (and larger than \ours FC). 

\paragraph{ClusterFit+.} Same as for SNCA, we improve ClusterFit~\cite{Yan2019ClusterFitIG} with our training procedure, and cross-validate the number of clusters (15000 for Imagenet and 1500 for CIFAR-100). As a result we improve its performance and have a fair comparison, everything being equal otherwise.

\definecolor{rowcolorgray}{gray}{0.92}
    
\begin{table}[t]
\caption{\textbf{Coarse-to-fine: comparison with the state of the art} for category-level retrieval (mAP, \%) and kNN classification (top-1, \%), with the ResNet50 architecture. 
We compare \ours with the state of the art~\cite{wu2018improving} and our stronger baselines. We \colorbox{rowcolorgray}{highlight} methods that use more parameters (32.9M vs $\sim$23.5M), see Table~\ref{tab:Ablation_baselines_coarse_to_fine} for details.\label{tab:coarse_to_fine_metrics}}
\centering

{\small
\begin{tabular}{l|cc|cc}
\toprule
  
\multirow{2}{4em}{Method}  & \multicolumn{2}{c}{CIFAR-100} & \multicolumn{2}{c}{ImageNet-1k}\\
& \ \knn\  & \  mAP\  & \ \knn\  & \  mAP\\

\midrule

 Baseline, Wu et al.~\cite{wu2018improving}  & 54.2  & \_ & 48.1  & \_\\
 SNCA, Wu et al.~\cite{wu2018improving}  & 62.3  & \_ & 52.8  & \_\\
 Baseline (ours) & 71.8 & 42.5 & 54.7 & 22.7 \\
 ClusterFit+ & 72.5  & 23.0 & 59.5  & 12.7 \\
 \rowcolor{rowcolorgray} SNCA+ & 72.2  & 35.9 & 55.4  & 31.8 \\
\midrule
\ours FC  & 75.6 & 55.0  & \textbf{69.1} & \textbf{44.4} \\
 \rowcolor{rowcolorgray} \ours & \textbf{77.7}& \textbf{55.7}  & \textbf{69.1} & 42.9 \\

\bottomrule
\end{tabular}}
\end{table}

\newcommand{\thh}{\textsuperscript{th}\ }
\subsection{Experimental details} 

\paragraph{Architectures.} 
Most experiments are carried out using  the ResNet-50 architecture~\cite{He2016ResNet} except for Section~\ref{subsec:transfer_learning_task} where we also use RegNet~\cite{Radosavovic2020RegNet} and ResNeXt~\cite{Xie2017AggregatedRT}. 

\paragraph{Training settings.}
Our training procedure borrows from the bag of tricks~\cite{Tong2018BagofTricks}:
 we use SGD with Nesterov momentum and cosine learning rates decay.
 We follow  Goyal et al.'s~\cite{Goyal2017AccurateLM} recommendation for the  learning rate magnitude: $\mathrm{lr}=\frac{0.1}{256} \times \mathrm{batch size}$.
 The data augmentation consists of random resized crop, RandAugment~\cite{Cubuk2019RandAugmentPA} and Erasing~\cite{Zhong2020RandomED}.
 We train for 600 epochs with batches of 1024 images at resolution $224\times 224$ pixels  (except for CIFAR-100: $32\times 32$).
 We set the temperature $\sigma$ to $0.05$  in all our experiments following Wu et al.~\cite{wu2018improving}.
 Appendix~\ref{app:train_settings} gives more details.

For the on-the-fly classification task, the unique hyper-parameter $k$ is cross-validated in $k \in \{10,15,20,25,30\}$.

\subsection{Coarse-to-fine experiments} 
\label{sec:sub_category_disc}

\paragraph{CIFAR and ImageNet experiments.}
Table~\ref{tab:coarse_to_fine_metrics} compares \ours results for \coarsetofine tasks with the baselines from Section~\ref{subsec:baseline}.
On CIFAR-100, \ours outperforms other methods by $\textbf{+5.5\%}$ top-1 accuracy. On ImageNet the gain over other methods is $\textbf{+13.7\%}$.

\ours also outperforms other methods on category-level retrieval, by $\textbf{13.2\%}$ on CIFAR and $\textbf{11.1\%}$ on ImageNet.
Table~\ref{tab:coarse_to_fine_metrics} shows that \ours not only provides a better on-the-fly classification (as evaluated by the \knn metric), but that the ranked list is more relevant to the query (results for mAP).

\begin{table}[t]
\caption{
\knn evaluation on iNaturalist-2018 with different semantic levels. 
The symbol $\varnothing$ refers to the unsupervised case (a unique class). 
We compare with the best competing method according to Table~\ref{tab:coarse_to_fine_metrics}. 
\label{tab:inat_18_knn}}
\centering
\scalebox{0.64}{
{\small
\begin{tabular}{c}
    \hspace{-8pt}
\begin{tabular}{@{}l|r|cccccccc@{\ }}
\toprule
  & Train $\rightarrow$ & $\varnothing$ & Kingdom & Phylum & class & Order & Family & Genus & Species \\
  & $\downarrow$Test / \#classes $\rightarrow$ & \small{1} & \small{6} & \small{25} & \small{57} & \small{272} & \small{1,118} & \small{4,401} & \small{8,142} \\
\midrule

\multirow{7}{1em}{\rotatebox{90}{ClusterFit+}}
&Kingdom   & 70.9   & 94.7      & 95.0     & 95.3     & 95.6      & 96.2     & 96.3       & 96.1  \\
&Phylum    & 48.8   & 87.4      & 90.3     & 90.7     & 91.1      & 92.6     & 92.6       & 92.2  \\
&Class     & 40.4   & 80.2      & 83.8     & 85.7     & 86.7      & 88.8     & 88.8       & 88.2  \\
&Order     & 17.1   & 54.5      & 59.0     & 61.4     & 70.8      & 73.9     & 74.3       & 72.3  \\
&Family    & 5.6   & 38.3      & 42.1     & 44.4     & 54.3      & 63.0    & 64.2       & 61.9  \\
&Genus     & 0.9   & 26.7      & 29.5     & 31.5     & 40.1      & 49.4     & 53.9       & 51.7  \\
&Species  & 0.3   & 21.8      & 23.7     & 25.2     & 32.7      & 40.3     & 44.7       & 43.4  \\
\midrule
\multirow{7}{1em}{\rotatebox{90}{\ours}}
&Kingdom   & 95.5   & 98.1      & 98.2     & 98.2     & 98.2      & 98.2     & 98.4       & 98.3       \\
&Phylum    & 90.0   & 94.1      & 96.6     & 96.7     & 96.8      & 96.7     & 96.9       & 96.7       \\
&Class     & 82.2   & 87.5      & 90.9     & 94.5     & 94.9      & 94.9     & 95.0       & 95.0       \\
&Order     & 54.0   & 61.7      & 66.9     & 72.7     & 87.1      & 87.5     & 87.6       & 87.3       \\
&Family    & 33.7   & 42.1      & 48.7     & 55.1     & 70.9      & 81.8     & 82.4       & 82.1       \\
&Genus     & 20.5   & 27.0      & 33.5     & 39.5     & 54.2      & 64.6     & 75.6       & 75.5       \\
&Species   & 15.9   & 20.4      & 25.5     & 30.8     & 42.7      & 51.2     & 61.9       & 67.7       \\

\bottomrule
\end{tabular}
\end{tabular}}}
\end{table}
\begin{table}[t]
\caption{
Category-retrieval evaluation (mAP, \%) on iNaturalist-2018 with different semantics levels.
We compare with the best baseline according to Table~\ref{tab:coarse_to_fine_metrics}.\label{tab:inat_18_map}}

\centering
\scalebox{0.66}{
{\small
\begin{tabular}{c}
    \hspace{-8pt}
\begin{tabular}{@{}l|r|ccccccc}
\toprule

  & Train $\rightarrow$ & Kingdom & Phylum & class & Order & Family & Genus & Species \\
  & $\downarrow$Test / \#classes $\rightarrow$ & \small{6} & \small{25} & \small{57} & \small{272} & \small{1,118} & \small{4,401} & \small{8,142} \\

\midrule
\multirow{7}{1em}{\rotatebox{90}{SNCA+}}
&Kingdom   & 97.6      & 83.3     & 75.9     & 59.2      & 56.0     & 54.9       & 55.0  \\
&Phylum    & 59.8      & 91.7     & 79.4     & 49.1      & 35.0     & 32.3       & 32.2  \\
&Class     & 41.3      & 73.1     & 89.9     & 49.2      & 28.1     & 23.6       & 23.0  \\
&Order     & 9.09      & 24.9     & 35.7     & 77.9      & 35.3     & 18.0       & 15.0  \\
&Family    & 2.24      & 6.43     & 11.2     & 35.7      & 68.4     & 29.1       & 21.7  \\
&Genus     & 0.39      & 2.47     & 5.03     & 18.1      & 36.6     & 60.5       & 46.0  \\
&Species   & 0.19      & 1.86     & 3.80     & 12.8      & 26.4     & 46.0       & 54.9  \\
\midrule
\multirow{7}{1em}{\rotatebox{90}{\ours}}
&Kingdom   & 98.6      & 88.3     & 79.7     & 60.8      & 58.0     & 55.9       & 55.5  \\
&Phylum   & 67.8      & 97.2     & 82.1     & 50.9      & 38.9     & 34.2       & 33.0  \\
&Class    & 50.1      & 74.9     & 95.4     & 51.2      & 32.3     & 25.9       & 24.1  \\
&Order    & 17.7      & 30.7     & 42.7     & 88.3      & 42.3     & 21.1       & 16.2  \\
&Family   & 8.70      & 13.2     & 18.0     & 43.9      & 83.1     & 34.8       & 24.2  \\
&Genus    & 6.78      & 9.72     & 13.5     & 29.0      & 46.9     & 77.2       & 53.9  \\
&Species  & 6.45      & 9.02     & 12.1     & 23.6      & 35.6     & 55.4       & 70.0  \\
\bottomrule
\end{tabular}
\end{tabular}
}}
\end{table}

\paragraph{Coarse-to-Fine with different taxonomic ranks.}
We showcase \ours on various levels of coarse granularity by training one model on each annotation level of iNaturalist-2018 and evaluating on all levels with \knn classification (Table~\ref{tab:inat_18_knn}) and retrieval (Table~\ref{tab:inat_18_map}).

\begin{figure}[t]
    \includegraphics[trim={0 160pt 0 0}, clip,scale=0.07]{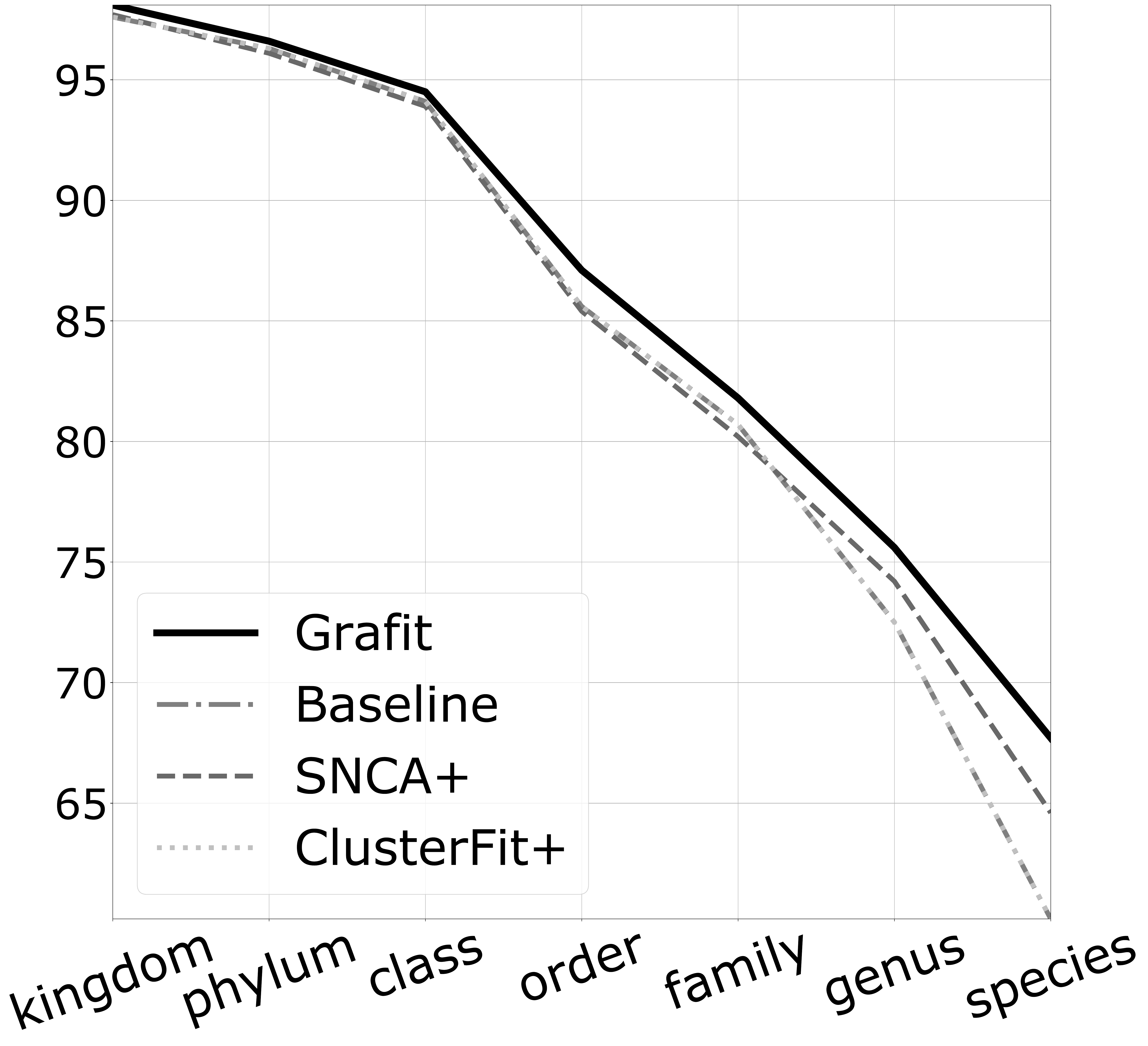} \hfill
    \includegraphics[trim={0 160pt 0 0}, clip,scale=0.07]{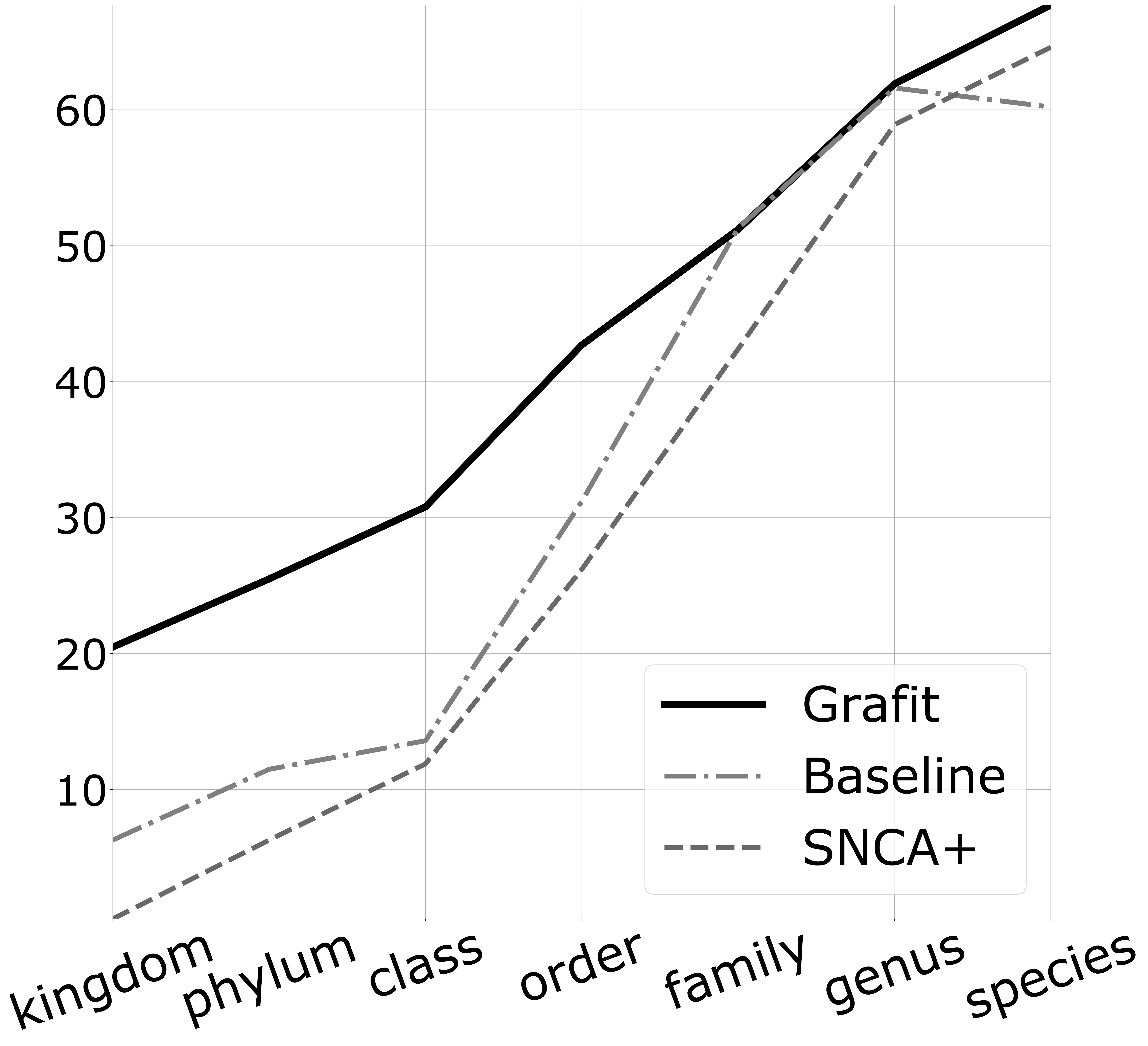}\\
    \ \includegraphics[scale=0.07]{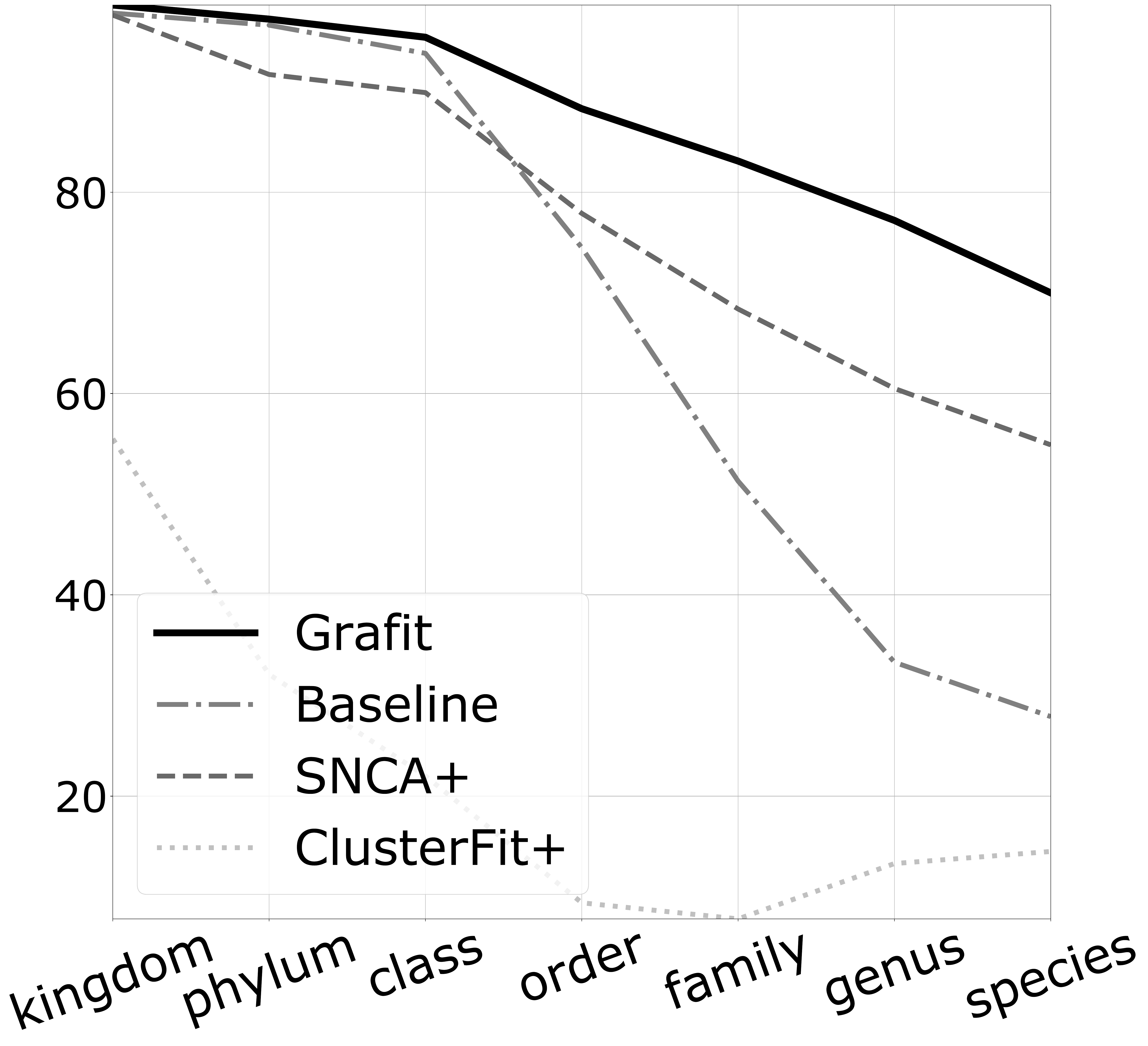} \hfill \includegraphics[trim={0pt 0pt 0 0}, clip,scale=0.07]{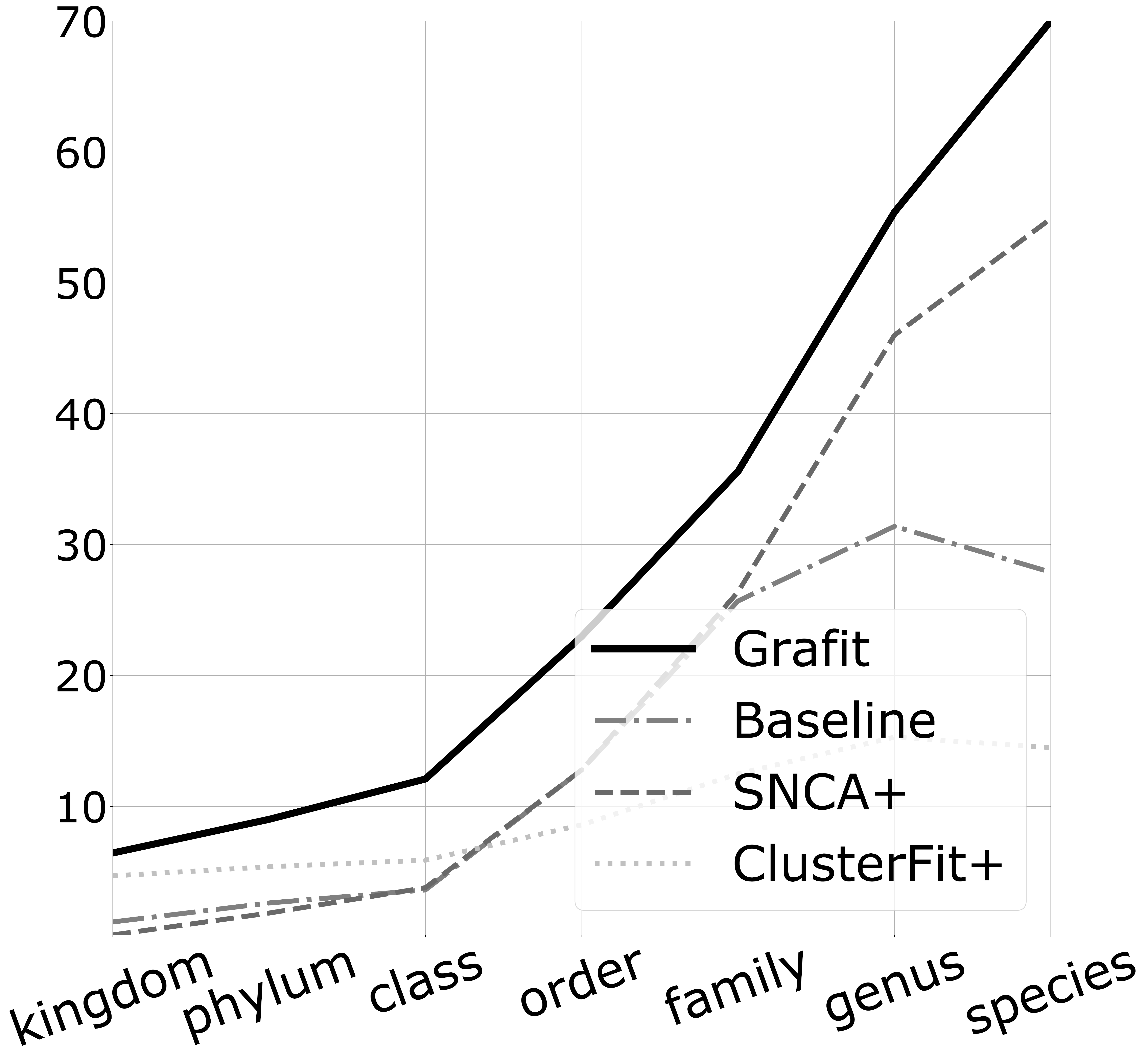}

\caption{
    Evaluation on iNaturalist-2018~\cite{Horn2018INaturalist} with 
    and
    \emph{left:} train=test granularity 
   \emph{right:} test at finest granularity (species). 
    We compare our method \ours, SNCA+, ClusterFit+ and  Baseline. 
    \emph{Top:} on-the-fly \knn classification (top-1 accuracy); 
    \emph{bottom:} category-level retrieval (mAP). 
   \label{fig:inat_knn_map_evaluation}}
\end{figure}
\begin{figure*}[t]
\begin{minipage}{0.72\linewidth}
 {\hspace{1.5cm} Train granularity: \emph{Family}  \hspace{2.5cm}  Train granularity: \emph{Genus}  }

\raisebox{2.4cm}{\rotatebox{90}{baseline}}
 \includegraphics[width=0.48\linewidth]{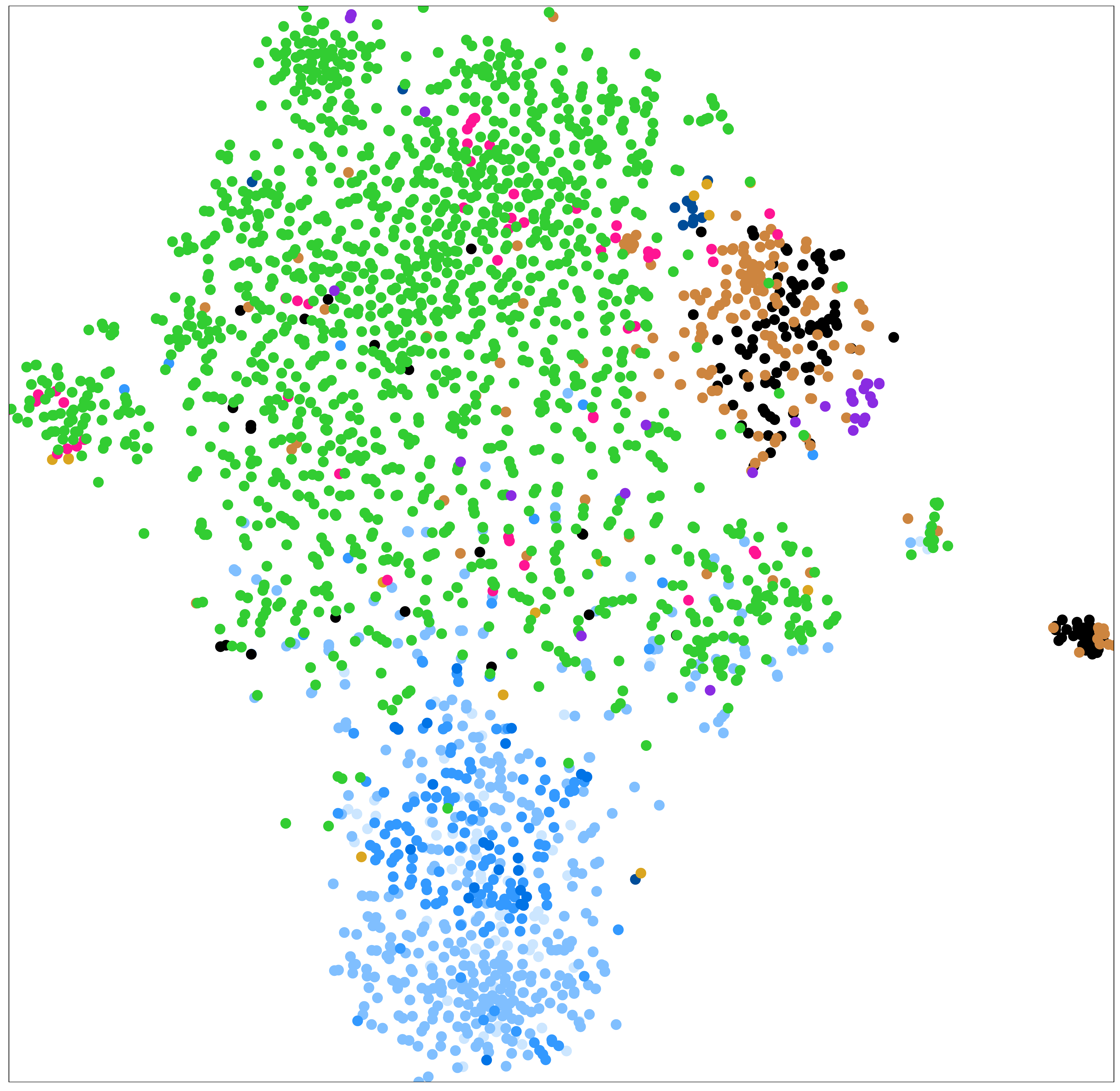} \hfill
 \includegraphics[width=0.48\linewidth]{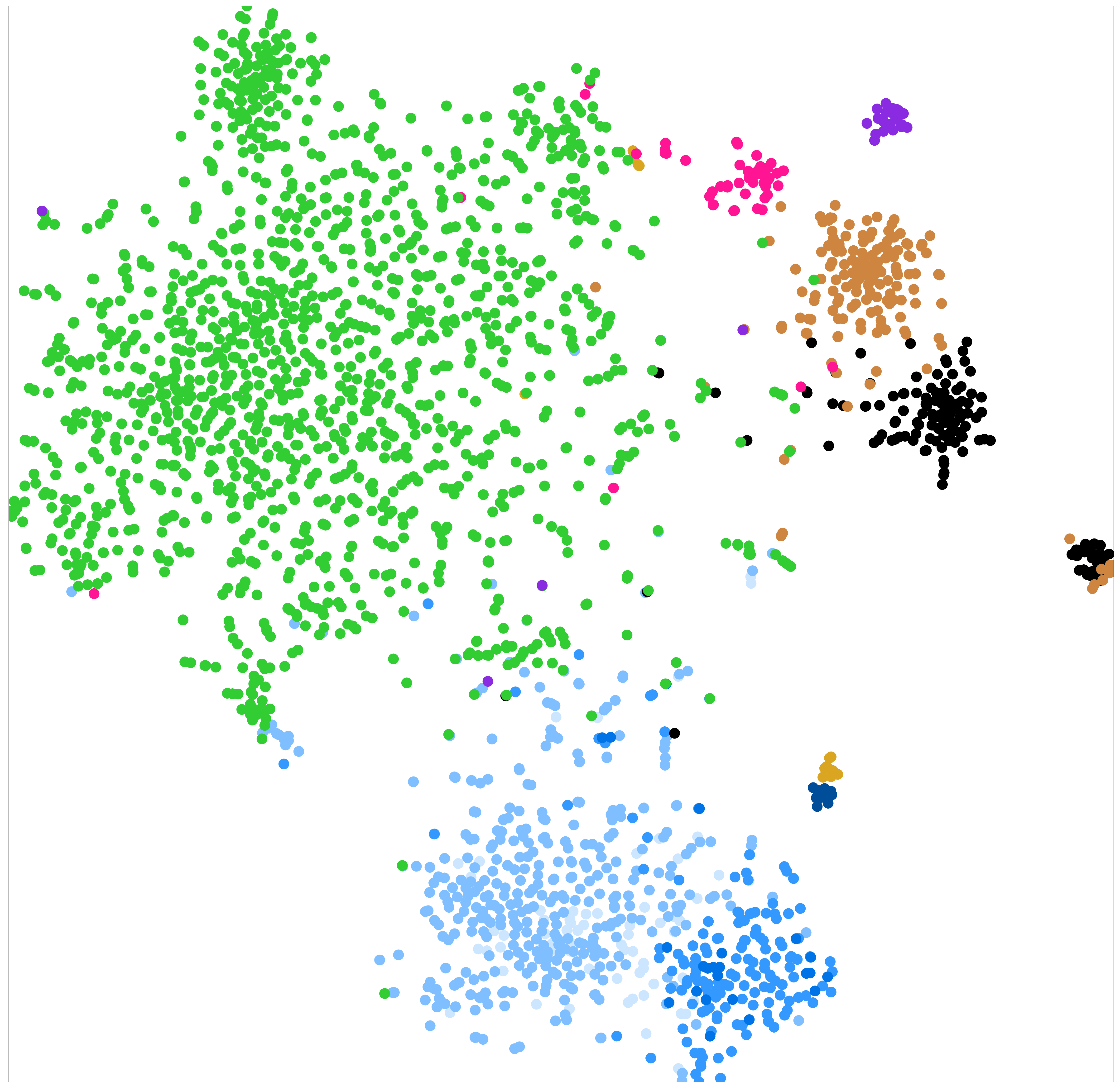} 

\raisebox{2.5cm}{\rotatebox{90}{\ours}}
 \includegraphics[width=0.48\linewidth]{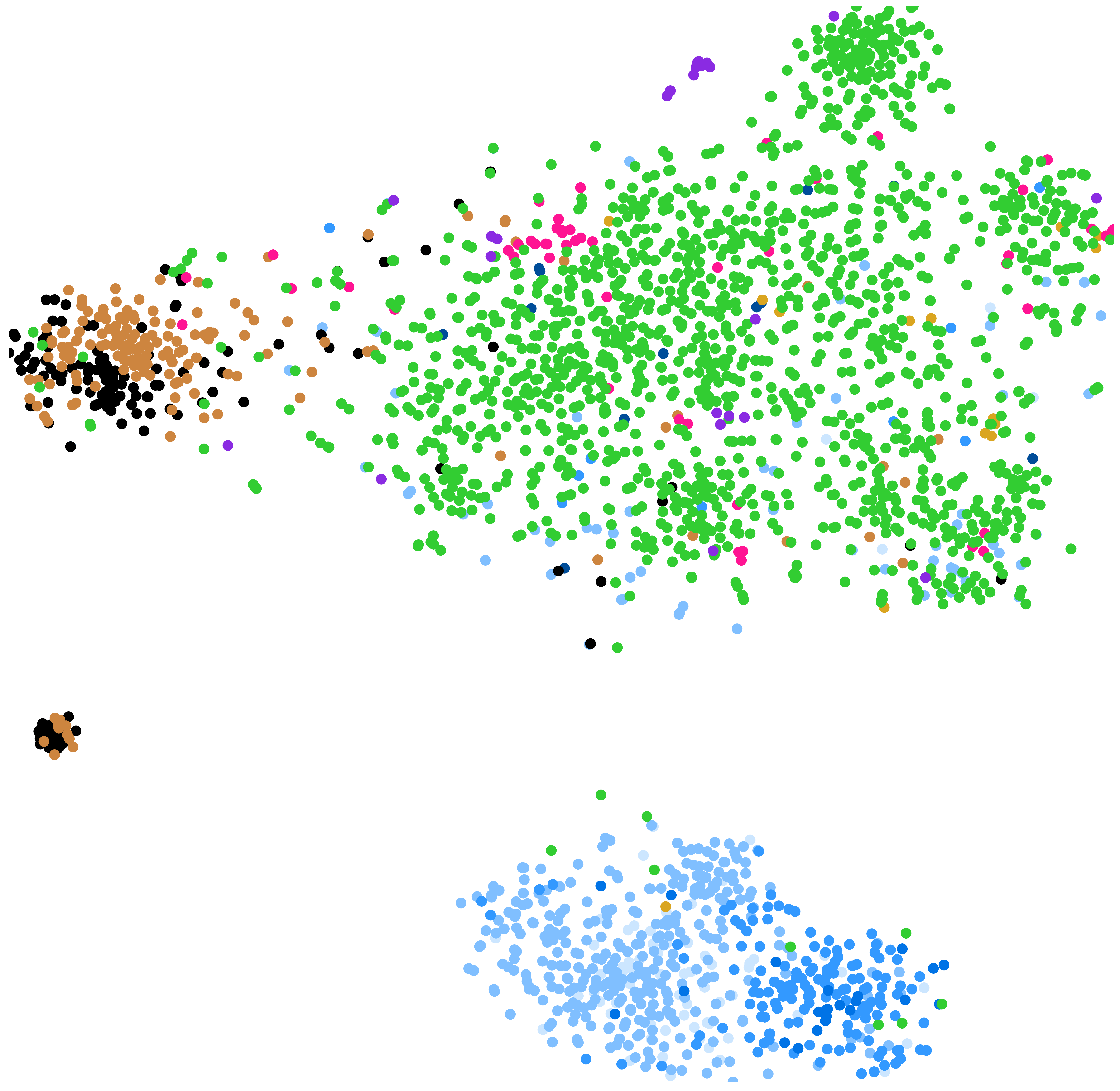} \hfill
 \includegraphics[width=0.48\linewidth]{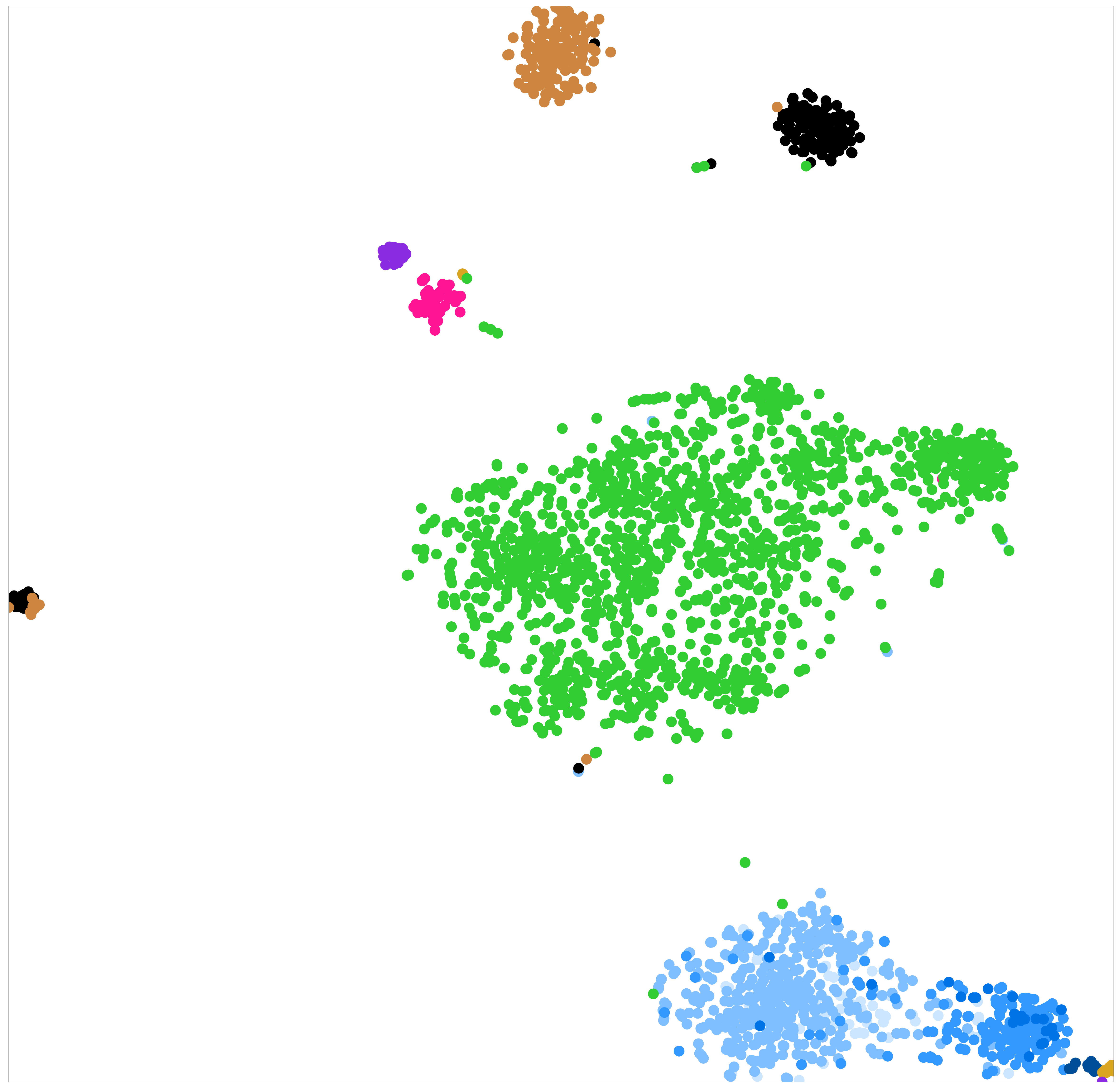}
\end{minipage}
\hfill
\begin{minipage}{0.27\linewidth}
~

    \caption{t-SNE representations of features from images of the family \emph{paridae}, focusing on the genus \emph{baeolophus} (in blue). 
    When trained with granularity Family, all depicted points have the same coarse label, while granularity Genus means that the network has seen 7 distinct labels. 
    Visually, \ours offers a better separation of the images than the baseline w.r.t. the two finest  level 'Genus' and 'Species'.  
    \label{fig:inat_tsne}
    }

\vspace{10pt}
\centering \includegraphics[width=0.9\linewidth]{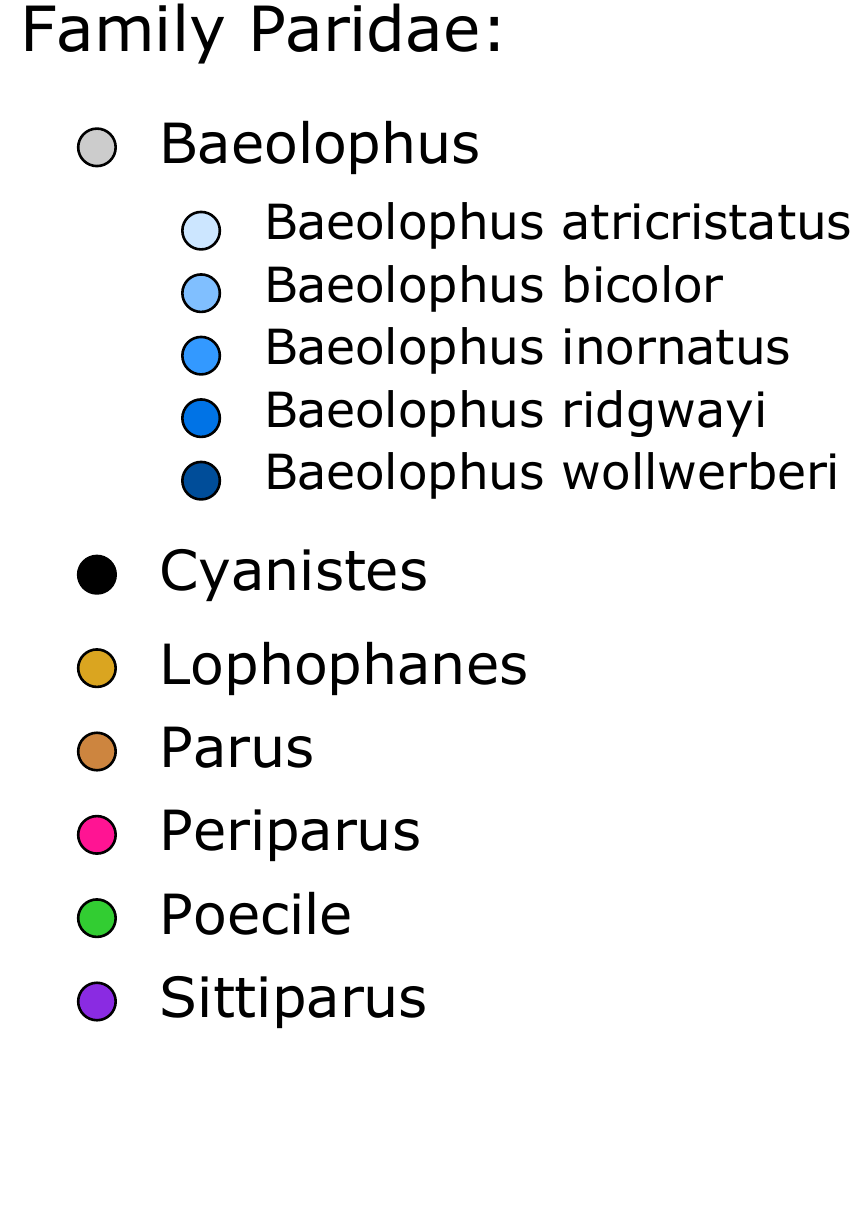} 
\end{minipage}
\end{figure*}
Figure~\ref{fig:inat_knn_map_evaluation} presents results with retrieval and \knn classification for two of the most interesting cases: when the  train and test granularities are the same (left), and on the finest test level (Species) with varying granularities at training time (right). 
On the left, the accuracy of all methods decreases as the granularity increases: this is expected as the task moves from coarse classification to fine, as it is more difficult to discriminate amongst a larger number of classes. 

We observe that the performance drop of \ours for category-level retrieval is reduced in comparison with  other methods. 
On the right figures, the accuracy of all methods increases as the level of annotation increases (keeping evaluation at Species).
\ours also stands out in this context, outperforming other methods.

We report comprehensive results with \ours and the  baselines from Section ~\ref{subsec:baseline} on iNaturalist-2019 \& 2018 in the supplemental material (Appendix~\ref{sec:table_inat}).

\paragraph{Visualizations.} 
Figure~\ref{fig:splash} shows visual results for the category-level retrieval task with \ours.
All the results for the baseline and \ours have the correct coarse label, but our method is better at a finer granularity.  
In Appendix~\ref{app:visu} we show that the improvement is even more evident when the granularity level at training time is coarser. 

Figure~\ref{fig:inat_tsne} presents t-SNE  visualizations~\cite{vandermaaten08tsne} of the latent spaces associated with the baseline and \ours for images associated with a sub-hierarchy of iNaturalist-2018. 

\subsection{Ablation studies}
\label{sec:ablation}

\paragraph{Losses, architectural choice and conditioning.} 
Table~\ref{tab:Ablation_baselines_coarse_to_fine} presents a study on CIFAR-100 and ImageNet-1k, where we ablate several components of our method.
A large improvement stems from the instance loss when it supplements the supervised kNN loss. 
It is key for discriminating at a finer grain. 
The category-level retrieval significantly benefits from our approach, rising from 22.7\% to 44.4\% in the best case. 
Coarse conditioning also has a consistent measurable impact on performance, yielding around $3$ mAP points across the various settings.

\definecolor{rowcolorgray}{gray}{0.92}
\definecolor{rowcolorred}{rgb}{1, 0.93, 0.92} 
\definecolor{rowcolorblue}{rgb}{0.9, 0.95, 1.0}
\definecolor{rowcolordarkerblue}{rgb}{0.8, 0.9, 1.0}

\begin{table}[t]
\caption{Ablation study on CIFAR-100 and ImageNet with ResNet50 architecture. 
We report results both for on-the-fly classification (\knn classifier, top-1 accuracy, \%) and category-level retrieval (mAP, \%). 
The rows corresponding to the main baselines and methods discussed through our paper are highlighted: our \colorbox{rowcolorgray}{baseline} and improved \colorbox{rowcolorred}{SNCA+} in grey and red, and our two variants \colorbox{rowcolorblue}{\ours-FC} and \colorbox{rowcolordarkerblue}{\ours} in blue. 
The last row is the result that \ours would obtain with a perfect coarse classification. 
\label{tab:Ablation_baselines_coarse_to_fine}}

\vspace{-5pt}
\centering
\scalebox{0.78}{
{\small
\begin{tabular}{c@{\ \ }c@{\ \ }c@{\ }c@{\ }c@{\ }|c@{\quad}c|c@{\quad}c@{\quad}c}
\toprule
\multicolumn{3}{@{\ }c@{\ }@{\ }}{Loss} & 
\multirow{2}{1.2cm}{\centering knn head proj. $P_\theta$}  & 
\multirow{2}{1.2cm}{\centering coarse cond.} & \multicolumn{2}{c|}{CIFAR100} & \multicolumn{3}{c@{\ \ }}{Imagenet}\\
\cmidrule(lr){1-3}
$\mathcal{L}_\mathrm{CE}$ & $\mathcal{L}_\mathrm{knn}$ & $\mathcal{L}_\mathrm{inst}$ &  & & \knn & mAP  & \knn & mAP & \#Params  \\
\midrule
\rowcolor{rowcolorgray} 
\checkmark & \_         & \_         & \_         & \_         & 71.8          & 42.5          & 54.7          & 22.7          & 23.5M \\
\checkmark & \_         & \_         & \_         & \checkmark & 71.8          & 43.1          & 54.7          & 24.4          & 23.5M \\
\_         & \_         & \checkmark & \_         & \_         & 54.3          & 14.3          & 41.7          & 3.47          & 23.5M \\
\checkmark & \_         & \checkmark & \_         & \_         & 76.9          & 51.0          & 65.0          & 26.0          & 23.5M \\
\_         & \checkmark & \_         & FC         & \_         & 70.0          & 39.7          & 57.8          & 30.7          & 23.8M \\
\_         & \checkmark & \checkmark & FC         & \_         & 75.6          & 53.6          & \textbf{69.1}          & 41.7          & 23.8M \\
\rowcolor{rowcolorblue}
\_         & \checkmark & \checkmark & FC         & \checkmark & 75.6          & 55.0          & \textbf{69.1}          & \textbf{44.4}          & 23.8M \\
\midrule
\rowcolor{rowcolorred}
\_         & \checkmark & \_         & MLP        & \_         & 72.2          & 35.9          & 55.4          & 31.8          & 32.9M \\
\_         & \checkmark & \_         & MLP        & \checkmark & 72.2          & 41.4          & 55.4          & 32.9          & 32.9M \\
\_         & \checkmark & \checkmark & MLP        & \_         & \textbf{77.7} & 52.9          & \textbf{69.1} & 39.4          & 32.9M \\
\rowcolor{rowcolordarkerblue}
\_         & \checkmark & \checkmark & MLP        & \checkmark & \textbf{77.7} & \textbf{55.7} & \textbf{69.1} & 42.9 & 32.9M \\
\midrule
\_         & \checkmark & \checkmark & MLP        & oracle     & 77.7          & \textit{59.3} & 69.1          & \textit{47.2} & 32.9M \\
\bottomrule
\end{tabular}}}
\medskip
\end{table}

\paragraph{Sanity check: training with coarse vs fine labels. } 
Table~\ref{tab:coarse_to_fine_fine_to_fine} compares the performance gap of several methods when training with coarse labels vs fine labels. 
The performance improvement of \ours over competing methods on Imagenet is quite sizable: with fine-tuning, \ours with coarse labels is almost on par with the baseline on fine labels. 
For on-the-fly classification, \ours with coarse labels reaches $69.1\%$ performance on Imagenet, significantly decreasing the gap with fine-grained labels settings. The \knn classification performance is $79.3\%$.
This concurs with our prior observations in Section~\ref{sec:sub_category_disc} on iNaturalist-2018. 

Overall, in this setting \ours provides some slight yet systematic improvement over the baseline. 
With a ResNet-50 architecture  at image resolution $224 \times 224$ pixels, \ours  reaches \textbf{79.6\%} top-1 accuracy with a \knn classifier on ImageNet, which is competitive with classical cross-entropy results published for this architecture. See Appendix~\ref{app:experiment} for a comparison (Table~\ref{tab:resnet_50_baseline}) and more results on Imagenet.

\begin{table}[t]
\caption{We compare coarse-to-fine and fine-to-fine context with mAP (\%), kNN (top-1, \%) and fine-tuning (FT) of a linear classifier with fine labels (top-1, \%)  on ImageNet.\label{tab:coarse_to_fine_fine_to_fine}}

\vspace{-5pt}
\centering
\scalebox{0.95}{
{\small
\begin{tabular}{l|ccc|ccc}
\toprule
Method & 
\multicolumn{3}{c}{Train Coarse}&
\multicolumn{3}{c}{Train Fine} \\

(with ResNet50) & mAP & kNN    & FT & mAP & kNN    & FT \\

\midrule

Baseline  & 22.7         & 54.7           & 78.1          & 51.5          & 78.0           & \textbf{79.3}  \\
SNCA+     & 31.8         &  55.4          & 77.9          & 72.0          & 79.1           & 77.4 \\
\ours FC  & \textbf{44.4}& \textbf{69.1}  & \textbf{78.3} & \textbf{72.4} & 79.2           & 78.5 \\
\ours     & 42.9         & \textbf{69.1}  & 77.9          & 71.2          & \textbf{79.6}  & 78.0  
 \\
\bottomrule
\end{tabular}}}
\end{table}

\subsection{Transfer Learning to fine-grained datasets} 
\label{subsec:transfer_learning_task}

We now evaluate \ours for transfer learning on fine-grained datasets (See) Table~\ref{fig:our_method}, with ImageNet pre-training. 

\paragraph{Settings.}
We initialize the network trunk with ImageNet pre-trained weights and fine-tune only the classifier. For our method, the network trunk $f_\theta$ remains fixed and  
the projector $P_\theta$ is discarded. 
For all methods we fine-tune during 240 epochs with a cosine learning rate schedule starting at 0.01 and batches of 512 images (details in Appendix~\ref{app:tl}).

\paragraph{Classifier.}
We experiment with two types of classifiers: a standard linear classifier (FC) and a multi-layer perceptron (MLP) composed of two linear layers separated by a batch-normalization and a ReLU activation.
We introduce this MLP because, during training, both \ours and SNCA+ employ an MLP projector, so their feature space is not learned to be linearly separable. In contrast, the baseline is trained with a cross-entropy loss associated with a linear classifier.

\paragraph{Tasks.}
We evaluate on five classical transfer learning datasets:
Oxford Flowers-102~\cite{Nilsback08},
Stanford Cars~\cite{Cars2013},
Food101~\cite{bossard14Food101},
iNaturalist 2018~\cite{Horn2018INaturalist} \& 2019~\cite{Horn2019INaturalist}.
Table~\ref{tab:dataset_coarse_to_fine} summarizes some statistics associated with each dataset.

\paragraph{Results.}
Table~\ref{tab:transfer_learning_tasks_max} compares a ResNet-50 pretrained on ImageNet with \ours, SNCA+, ClusterFit~\cite{Yan2019ClusterFitIG} and our baseline on five transfer learning benchmarks. 
Our method outperforms all methods. The table also shows the relatively strong performance of SNCA+. %

Table~\ref{tab:transfer_learning_tasks_sota} compares \ours with the RegNetY-8.0GF~\cite{Radosavovic2020RegNet} architecture against the state of the art, on the same benchmarks. Note that this architecture is significantly faster than the EfficientNet-B7 and ResNet-152 employed in other papers, and that we use a lower resolution in most settings. %

In Table~\ref{tab:transfer_learning_tasks_sota} we consider  models pre-trained on ImageNet with a classifier fine-tuned on the fine-grained target dataset.
In each case we report results with \ours (with a MLP for the projector $P_\theta$) and \ours FC. 
See more detailed results in  Appendix~\ref{app:experiment} Table~\ref{tab:transfer_learning_tasks}.

In summary, \ours establishes the new state of the art. 
We point out that we have used a consistent training scheme across all datasets, and a single architecture that is more efficient than in competing methods.

    \begin{table}[t]
    \caption{\label{tab:transfer_learning_tasks_max}
     Comparison of transfer learning performance for different pre-training methods.  
     All methods use a ResNet-50 pre-trained on Imagenet. 
     The training procedues are the same (except the result reported for ClusterFit~\cite{Yan2019ClusterFitIG}).
     We report the top-1 accuracy (\%) with a single center crop evaluation at resolution $224 \times 224$.
     See Table~\ref{tab:transfer_learning_task_diverse_arch} of Appendix~\ref{app:tl} for additional results with other architectures. 
    }
    
    \vspace{-5pt}
    \centering
\newcommand{\vtabheadB}[2]{%
\rotatebox{90}{\begin{minipage}{12mm}
#1 \\ #2 
\end{minipage}%
}}
 \hspace{-8pt}  
    \small
    \begin{tabular}{l|cccc|c|c}
      \toprule
       Dataset & \rotatebox{90}{Baseline} &  \vtabheadB{ClusterFit}{\cite{Yan2019ClusterFitIG}} &\rotatebox{90}{ClusterFit+ }& \rotatebox{90}{SNCA+}   & \rotatebox{90}{\ours} & \rotatebox{90}{\ours FC}\\
   
    \midrule
    	Flowers-102       & 96.2 & \_ & 96.2 & \textbf{98.2}  & \textbf{98.2}  & 97.6\\ 
    	Stanford Cars     & 90.0 & \_  & 89.4  & 92.5      & 92.5  & \textbf{92.7} \\ 
    	Food101           & 88.9 & \_  & 88.9  & 88.8      & \textbf{89.5} & 88.7 \\ 
    	iNaturalist 2018  & 68.4 & 49.7  &  67.5    & 69.2   &  \textbf{69.8} & 68.5 \\ 
    	iNaturalist 2019  & 73.7 & \_  & 73.8  & 74.5     & \textbf{75.9}  & 74.6  \\
      \bottomrule
    \end{tabular}
    \end{table}

\begin{table}[t]
\caption{\label{tab:transfer_learning_tasks_sota}
 State of the art for transfer learning with pretrained ImageNet-1k models.
We report top-1 accuracy (\%) with a single center crop.
For \ours we use a 39M-parameter RegNetY-8.0GF~\cite{Radosavovic2020RegNet} with resolution $384 \times 384$ pixels that is $4\times$ faster than EfficientNetB7 at inference.
``Res'' is the inference resolution in pixels. 
}
\centering

\vspace{-7pt}
\hspace{-8pt} \scalebox{0.84}{
\small
\begin{tabular}{@{\ }l|lrc|c|c@{\ }}
  \toprule
  & \multicolumn{4}{c|}{Best reported results (\%)} & \ours\\
  Dataset & State of the art  & \# Params & Res &  Top-1 & Top-1  \\
\midrule
	Flowers-102 & EfficientNet-B7~\cite{tan2019efficientnet}         & 64M   & 600   & 98.8 & \textbf{99.1} \\
	Stanford Cars        & EfficientNet-B7~\cite{tan2019efficientnet}  & 64M   & 600   & \textbf{94.7} & \textbf{94.7} \\
	Food101  &  EfficientNet-B7~\cite{tan2019efficientnet}     & 64M   & 600   & 93.0 & \textbf{93.7} \\    
	iNaturalist 2018 & ResNet-152~\cite{Chu2020FeatureSA}    & 60M   & 224   & 69.1 & \textbf{81.2} \\    
	iNaturalist 2019 & --                                   & --    & --    & --   & \textbf{84.1} \\

  \bottomrule
\end{tabular}}
\vspace{-7pt}
\end{table}


\section{Conclusion}
\vspace{-3pt}
\label{sec:conclusion}
\vspace{-3pt}
This paper has introduced a procedure to learn a neural network that offers a finer granularity than the one provided by the annotation. 
It  improves the performance for fine-grained category retrieval within a coarsely annotated collection. 
For on-the-fly \knn classification, \ours significantly reduces the gap with a network trained with fine labels. 
It also translates into better transfer learning to fine-grained datasets, outperforming the current state of the art with a more efficient network.

\clearpage

{\small
\bibliographystyle{ieee_fullname}
\bibliography{egbib}
}
\clearpage \newpage 
\appendix\newpage

\twocolumn[
  \begin{@twocolumnfalse}
\begin{center}
{\LARGE
\textbf{Supplementary material for Paper ID 5395} \\[0.2cm]
\scalebox{0.95}{
``\textbf{\ours: Learning fine-grained image representations with coarse labels}'' }

\vspace{0.5cm} %
}
\end{center}
\end{@twocolumnfalse}
]

In this supplementary material we report additional analyses, results and examples that complement our paper.  
Appendix~\ref{sec:disc_gran} presents two experiments on the impact of self-supervised losses on the granularity and distribution of embeddings.
Appendix~\ref{app:experiment} contains a more accurate description of the experimental settings and more detailed account of the experiments conducted for this paper.
Appendix~\ref{app:visu} presents additional visualizations of the ranking obtained with \ours.

\newcommand{\norm}[1]{\left\lVert#1\right\rVert_2}

\newtheorem{definition}{Definition}

\section{About granularity}
\label{sec:disc_gran}

Is it possible to create representations that discriminate between classes finer than the available coarse labels?  
Considering that we have seen only coarse labels at training time, how can we exploit the coarse classifier for fine-grained classification, if useful at all? 
In this section we discuss these two questions and construct an experiment to analyze the role of the losses and of the coarse classifier. 
We then provide empirical observations. 

\paragraph{Practical setup. } 

In the following two experiments, we consider the CIFAR-100 benchmark that has two granularity levels with 20 and 100 classes (see Section \ref{sec:dataset}).

We denote by $f$ the Resnet-18 trunk mapping from the image space to an embedding space. 
We train the neural network trunk $f$ with three possible losses: 
\begin{itemize}
    \item 
    Baseline: regular cross-entropy classification training $\mathcal{L}_\mathrm{CE}$ with coarse or fine classes; 
    \item
    Triplet loss: training a triplet loss $\mathcal{L}_\mathrm{Triplet}$ to differentiate between image instances (does not use the labels);
    \item
    $\mathcal{L}_\mathrm{CE} + \mathcal{L}_\mathrm{Triplet}$: sum of the two losses. 
    This is intended to be a simple proxy of \ours.
\end{itemize}

\subsection{Experiment: separating arbitrary fine labels}  
\label{sec:analysis}

This experiment is inspired both by the Rademacher complexity \cite{Boucheron2005Rademacher} and by the self-supervised learning (SSL) literature \cite{Berthelot2019ReMixMatchSL}.  
In SSL, the standard way to evaluate the quality of a feature extractor $f$ is to measure the accuracy of the network after learning a linear classifier $l$ for the target classes on top of $f$. 
The Rademacher complexity measures how a  class of functions (\textit{i.e.} $l\circ f$, with $f$ fixed and $l$ learned) is able to classify a set of images with random binary labels.

For this experiment we train the trunk $f$ jointly with a (coarse class) classifier with $\mathcal{L}_\mathrm{CE}$ using coarse labels.
We hope to improve the granularity of $f$, i.e. improve the network trunk such that a (finer-grained) classifier $l$ trained on top of $f$ performs better at discriminating between instances that have the same coarse label. 

    \begin{table}[t]
    \caption{Separability experiment on CIFAR-100.  
    The trunk is trained with coarse labels only. 
    Images with the same coarse label are randomly grouped into two distinct fine-grain labels (40 distinct labels in total).  
    Then we fine-tune a linear classifier on this synthetic labels and measure the top-1 accuracy on fine-labels.  
    When conditioning, the estimator exploits the hierarchy: we first predict the coarse class and condition on it to make the final prediction. We report results with three training losses.\label{tab:rademacher_complexity_v2}}
    \centering
    \begin{tabular}{l|c|c}
    \toprule
    Training &  \multicolumn{2}{c}{Top-1 (\%)} \\
    loss & no cond. & coarse cond. \\
    \midrule 
    $\mathcal{L}_\mathrm{CE}$ & 
        53.7~\stdminus{0.3} & 54.5~\stdminus{0.3} \\
    $\mathcal{L}_\mathrm{Triplet}$ & 
        26.4~\stdminus{0.3} & --- \\
    $\mathcal{L}_\mathrm{CE} +\mathcal{L}_\mathrm{Triplet}$   &
        57.1~\stdminus{0.2} & \textbf{58.5}~\stdminus{0.3} \\
    \midrule
    Random network & 8.7~\stdminus{0.3} & --- \\
    \bottomrule
    \end{tabular}
    \end{table}

\paragraph{Random labels.}

We generate synthetic fine labels by the following process:
for each coarse label, we randomly and evenly split the training images into two subcategories, yielding 40 classes in total. 
Inspired by the empirical Rademacher estimation, we sample 10 distinct splits of random labels. 
For each split,  we learn a linear classifier $l$ on top of $f_i$.  
We then compute the mean accuracy (top-1, \%) of $l \circ f_i$ on the training examples for the three losses.  %
By evaluating to what extent one can fit a linear classifier $l$ on top of $f$,
this experiment measures how well the data are spread in the representation spaces. 

\paragraph{Impact of the loss terms.}

We report the results in Table~\ref{tab:rademacher_complexity_v2}.
We can see that, to distinguish between our synthetic fine labels, training with the triplet loss $\mathcal{L}_\mathrm{Triplet}$ in combination with the classification loss $\mathcal{L}_\mathrm{CE}$ is essential: the sum of losses outperforms each individual loss.

\paragraph{Conditioning.}

We also measure the impact of \emph{conditioning on coarse classes}: we first predict the coarse label with the coarse classifier, and leverage its softmax output to classify the fine class. 
This clearly improves the accuracy, which motivates our fusion strategy inspired by this conditioning in Section~\ref{sec:categoryretrieval}.

\subsection{Experiment: varying the training granularity}

In this section we make empirical observations related to the training granularity in the embedding space. 

We train $f$ with one of the three losses and either coarse or fine labels as supervision. 
In a second stage, we train a linear classifer $l$ on the Resnet-18 trunk with fine class supervision, and evaluate its accuracy on the test set.

\paragraph{Accuracies.}

We first quantify the quality of the representation space. 
The accuracies are reported in Table~\ref{tab:analysis_training_schemes}. 
We observe that the coarse labels are almost as good as the fine labels as a pre-training. 
The unsupervised $\mathcal{L}_\mathrm{Triplet}$ loss performs significantly worse, which concurs with our previous separability experiment. 
Combining this loss with the $\mathcal{L}_\mathrm{CE}$ loss improves is, both with coarse and fine supervisions.

\begin{table}[t]
\caption{
Top-1 accuracy of a ResNet-18 on CIFAR-100 for different training schemes. 
We report the results after finetuning of the linear classifier on the fine labels (see Section~\ref{sec:disc_gran}).
The Triplet training is unsupervised, so the two columns are the same.
\label{tab:analysis_training_schemes}}
\centering
\scalebox{1.0}{
{
\begin{tabular}{|l|cc|}
\toprule
  Method & Train Coarse   &  Train Fine \\

\midrule

$\mathcal{L}_\mathrm{CE}$ & 80.4~\stdminus{0.2} & 80.6~\stdminus{0.2} \\
$\mathcal{L}_\mathrm{Triplet}$ & 76.5~\stdminus{0.2} & 76.5~\stdminus{0.2} \\
$\mathcal{L}_\mathrm{CE} + \mathcal{L}_\mathrm{Triplet}$ & \textbf{80.9~\stdminus{0.2}} & \textbf{81.3~\stdminus{0.2}} \\

\bottomrule
\end{tabular}}}
\end{table}

\paragraph{Size of the representation space.}
\label{sec:size_space}

We quantify the information content of embedding vectors by computing their principal components analysis (PCA).
This is a reasonable proxy for information content given that the features are separated by linear classifiers afterwards. 
We observe the cumulative energy of the PCA components ordered by decreasing energy.
We assume that a more uniform energy distribution (and thus lower curves) means that the representation is richer, since a few vector components cannot summarize it. 

\begin{figure}[t]
\centering

\begin{tabular}{c}
\includegraphics[width=1.0\linewidth]{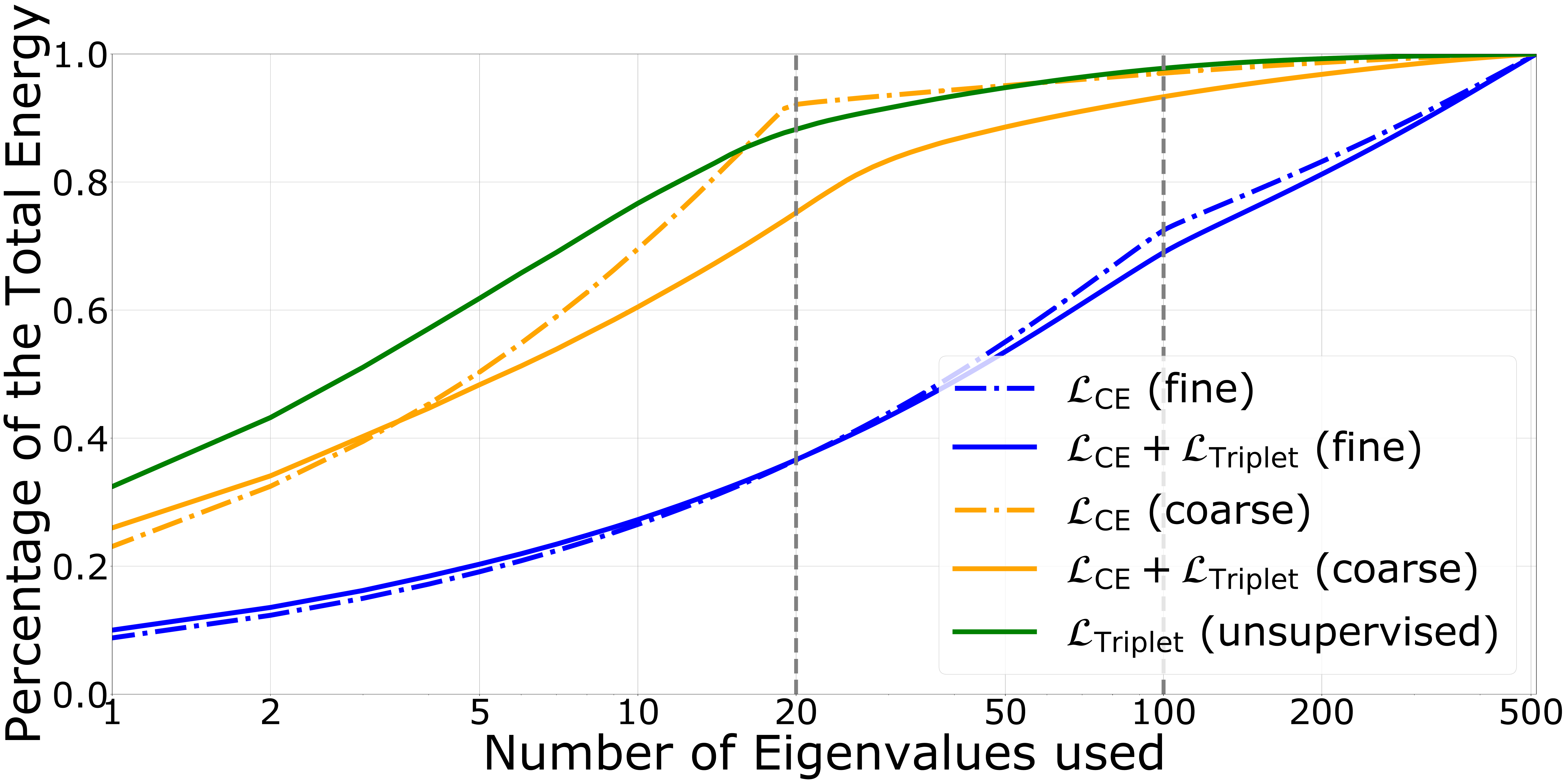}\\
   
\end{tabular}

\caption{\label{fig:pca_energy}
 Cumulative energy of the PCA decomposition of CIFAR-100 image embeddings, depending on the granularity of the training labels (20 or 100  classes). 
}
\end{figure}

Figure~\ref{fig:pca_energy} shows the results.
When training with $\mathcal{L}_\mathrm{CE}$ loss, the most uniform distribution for the principal components is obtained for the fine supervision. This is expected since it is a finer-grain separation of entities and that can not be summarized with a  subspace as small as the one associated with a relatively small number of categories.
Notice that the training granularity (20 or 100 classes) can be read as an inflexion point on the PCA decomposition curves.
The loss $\mathcal{L}_\mathrm{Triplet}$ is not very informative on its own but does improve the cross-entropy representation when combined with it.

\paragraph{Discussion.}

This simple preliminary experiment shows that the label granularity has a strong impact on very basic statistics of the embedding distribution. 
It is the basic intuition behind \ours: a rich representation can be obtained using just coarse labels, if we combine them with a self-supervised loss.

\section{Additions to the experiments}
\label{app:experiment}

This section details the training procedure of \ours and provides more extensive experimental results.

\subsection{Training settings}

\label{app:train_settings}
As described in the main part, 
our training procedure is inspired by Tong et al.~\cite{Tong2018BagofTricks}:
 we use SGD with Nesterov momentum and cosine learning rates decay.
 We follow  Goyal et al.'s~\cite{Goyal2017AccurateLM} recommendation for the  learning rate magnitude: $\mathrm{lr}=\frac{0.1}{256} \times \mathrm{batch size}$.
 The augmentations include random resized crop, RandAugment~\cite{Cubuk2019RandAugmentPA} and Erasing~\cite{Zhong2020RandomED}.
 We train for 600 epochs with batches of 1024 images of resolution $224\times 224$ pixels  (except for CIFAR-100 where the resolution is $32\times 32$).
 For \ours with $\mathcal{L}_\mathrm{inst}$ we use $T=4$ different data-augmentations on ImageNet and $T=8$ on CIFAR-100.
 For the supervised loss we use one data-augmentation in order to have the same training procedure as our supervised baseline.

\paragraph{Weighting of the losses.}
\label{sec:lossweight}

We investigate the impact of weighting the losses  $\mathcal{L}_\mathrm{knn}$ and $\mathcal{L}_\mathrm{inst}$. 
For example, on CIFAR-100 classification, Table~\ref{tab:weighting_comparison} shows that an equal weighting gives the best or near-best results. 
Therefore, to avoid adding a hyper-parameter and in order to simplify the method, we chose to not use weighting, \emph{i.e.} we just sum up the two losses.

   \begin{table}[t]
    \caption{Category-level (mAP, \%) and one-the-fly kNN classification (top-1, \%)  performance in a coarse-to-fine setting on CIFAR-100 with different loss weighting. Our total loss is $\mathcal{L}_\mathrm{tot}(x)= \mathcal{L}_\mathrm{knn}(g_{\theta}(x),y) +  \lambda \mathcal{L}_\mathrm{inst}(x)$ with $\lambda$ being a real-valued coefficient.
    \label{tab:weighting_comparison}}
    \centering
    \scalebox{0.96}{
    {\small
    \hspace{-8pt}
    \begin{tabular}{@{\ }l|ccccccccc@{\ }}
    \toprule
    $\lambda$ & 0.0   & 0.2 & 0.4 & 0.6 &  0.8 & 1.0  & 1.2 & 1.4\\
    \midrule
    mAP       & 35.9  & 46.3   & 49.6   & 51.4   &  52.4   & 52.9 & 52.8   & 52.4 \\
    \midrule
    kNN       & 72.2  & 70.0   & 73.2   & 74.8   & 75.8    & 77.7 & 77.4   & 77.7 \\
    \bottomrule
    \end{tabular}}}
    \end{table}

\paragraph{A strong Baseline.}

Our training procedure improves the ResNet-50 performance and thus is a strong baseline against which we can compare \ours.
Therefore, Table~\ref{tab:resnet_50_baseline} compares our baseline on ImageNet with other ResNet-50 training procedures.
We observe that our training procedure gives better results than many other approaches. 
This makes it possible to isolate the contribution of our improved training practices and that of the \ours loss.

    \begin{table}[t]
    \caption{Performance comparison (top-1 accuracy) with our ResNet-50 baseline and state of the art ResNet-50 on ImageNet. All results are with single center crop evaluation with image resolution $224 \times 224$.
    \label{tab:resnet_50_baseline}}
    \centering
    \scalebox{0.99}{
    {\small
    \begin{tabular}{l|l|l}
    \toprule
    Method & Extra-data & Top-1 (\%) \\
    \midrule
    ResNet-50~\cite{He2016ResNet} PyTorch & \_ & 76.2\\
    RandAugment~\cite{Cubuk2019RandAugmentPA} & \_ & 77.6\\
    CutMix~\cite{Yun2019CutMix} & \_ & 78.6\\
    Noisy-Student~\cite{Xie2019SelftrainingWN} & JFT-300M~\cite{Xie2019SelftrainingWN} & 78.9\\
    Billion Scale~\cite{Yalniz2019BillionscaleSL} & YFCC100M~\cite{Thomee2016YFCC100MTN} & 79.1\\
    \midrule
    Our Baseline & \_ & \textbf{79.3}\\
    \bottomrule
    \end{tabular}}}
    \end{table}

\subsection{\{coarse,fine\}-to-\{coarse,fine\}: evaluation}
\label{subsec:eval_fc}

We compare our main baselines and \ours's performance in the 4 following scenarios: coarse-to-coarse, coarse-to-fine, fine-to-fine and fine-to-coarse.
The evaluations are performed with two classifiers: a \knn classifier (kNN) and a linear classifier fine-tuned (FT) with a cross-entropy loss on top of the embeddings. 

The results in Table~\ref{tab:baselines_full} show that \ours training improves the accuracy in almost all settings, including the fine-to-fine setting, which is just regular classification with the vanilla labels for Imagenet.

    \begin{table}[t]
    \caption{Performance comparison (top-1 accuracy) when learning and testing at different granularities (ResNet50). 
    For CIFAR-100, there are 100 fine and 20 coarse concepts. ImageNet covers 1000 fine and 127 coarse concepts. 
    We report the results of both the kNN classifier and of a linear classifier fine-tuned with the target granularity (FT).
    \label{tab:baselines_full}}
    
    \centering
    \scalebox{0.82}{
    {\small
    \begin{tabular}{|l|l|l|cccc|}
    \toprule
    \multirow{2}{*}{} & 
    \multirow{2}{3em}{Method} & 
    \multirow{2}{2.5em}{$\downarrow$ Test} & 
    \multicolumn{2}{c}{Train Coarse}&
    \multicolumn{2}{c|}{Train Fine} \\

     & &  & \multicolumn{1}{c}{kNN}    &   \multicolumn{1}{c}{FT} & \multicolumn{1}{c}{kNN}    &   \multicolumn{1}{c|}{FT} \\
    
    \midrule
    \multirow{6}{2em}{\rotatebox{90}{CIFAR-100}} & Baseline & \multirow{3}{2em}{Coarse}
    & 89.3\stdminus{0.1} & 89.4\stdminus{0.2} & 90.3\stdminus{0.1} & 90.5\stdminus{0.2}  \\
    & SNCA+ &  & 88.4\stdminus{0.3} & 88.9\stdminus{0.3} & 88.8\stdminus{0.1} & 90.2\stdminus{0.1}  \\
    & \ours&   & \textbf{90.6}\stdminus{0.1} & \textbf{90.6}\stdminus{0.1}    & \textbf{90.6}\stdminus{0.3} & \textbf{90.9}\stdminus{0.2}    \\
    \cmidrule{2-7}
    & Baseline  & \multirow{3}{2em}{Fine}
    &  71.8\stdminus{0.3} &  82.3\stdminus{0.2} & 82.7~\stdminus{0.2} & 82.7~\stdminus{0.2}   \\
    & SNCA+ &  & 72.2~\stdminus{0.3} & 82.0~\stdminus{0.4} & 81.7~\stdminus{0.1} & 82.9~\stdminus{0.1}  \\
     & \ours &   &  \textbf{77.7}\stdminus{0.2} & \textbf{83.7}\stdminus{0.2} & \textbf{83.2}\stdminus{0.3} & \textbf{83.7}\stdminus{0.2}  \\
    \bottomrule
    \bottomrule
    
   \multirow{6}{2em}{\rotatebox{90}{ImageNet-1k}} &  Baseline  & \multirow{3}{2em}{Coarse}
    & 87.0\stdminus{0.1} & \textbf{87.6}\stdminus{0.1} &  87.4\stdminus{0.1} & 87.9\stdminus{0.1}   \\
    & SNCA+ &  & 87.7\stdminus{0.1}  & 87.5\stdminus{0.1} & 88.9\stdminus{0.1} & 87.2\stdminus{0.1}  \\
   &  \ours  &  & \textbf{88.4}\stdminus{0.1}  & 87.3\stdminus{0.1} & \textbf{89.2}\stdminus{0.1} & \textbf{87.7}\stdminus{0.1}  \\
    \cmidrule{2-7}
   &  Baseline  & \multirow{3}{2em}{Fine}
    & 54.7\stdminus{0.2} & \textbf{78.1}\stdminus{0.1} & 78.0\stdminus{0.1}  & \textbf{79.3}\stdminus{0.1}  \\
   & SNCA+ &  & 55.4\stdminus{0.2} & 77.9\stdminus{0.1} & 79.1\stdminus{0.1} & 77.4\stdminus{0.1}  \\
   &  \ours   & & \textbf{69.1}\stdminus{0.2} & 77.9\stdminus{0.1} & \textbf{79.6}\stdminus{0.1} & 78.0\stdminus{0.1}  \\
    \bottomrule
    \end{tabular}}}
    \end{table}

\subsection{Coarse-to-Fine with different taxonomic rank}
\label{sec:table_inat}

\paragraph{Datasets.} We carry out evaluations on iNaturalist-2018, and with iNaturalist-2019~\cite{Horn2019INaturalist}, which is a subset of iNaturalist-2018~\cite{Horn2018INaturalist} where  classes with too few images have been removed. 
iNaturalist 2019 dataset is thus composed of 268,243 images divided into 1,010 classes at the finest level. From the coarse to the finest level, we have 3 classes for Kingdom, 4 classes for Phylum, 9 classes for Class, 34 classes for Order, 57 classes for Family, 72 classes for Genus and 1,010 classes for Species.

\paragraph{Results.}
We report exhaustive results with our two coarse-to-fine evaluation protocols with all our baselines on iNaturalist-2018~\cite{Horn2018INaturalist} and iNaturalist-2019~\cite{Horn2019INaturalist} in Table~\ref{tab:INaturalist_Results}.

We comment more specifically the \knn classification accuracy (left) because for retrieval, \ours outperform all the baselines by a large margin. 
The table on the right is divided in 10 matrices each containing results for one combination of a method and a dataset (iNaturalist 2018 or 2019).

The diagonal values in the matrices correspond to a traditional setting where the training and the test granularity are the same.
Even in this case, the \ours descriptors outperforms the baseline methods most often. 
On iNaturalist 2018, for Species, the finest and most challenging level, the additional \ours loss improves the top-1 accuracy by 7\% absolute.
The gain is more marginal for iNaturalist 2019 (+0.9\%), which shows that \ours is especially useful for unbalanced class distributions where some classes are in a low-shot training regime.

The lower triangle of each matrix reports the coarse-to-fine results, which is the setting in which we focus in our paper. 
\ours obtains the best results for most combinations, with accuracy gains of around 10 points with respect to the baseline and by a few points for ClusterFit+. 
It is interesting to look at the $\varnothing$ column, which is the  unsupervised case. 
In that case, the baseline training reduces to a random network, but \ours is able to extract signal from the \knn loss.

The upper triangle is the fine-to-coarse setting, where finer labels are available for the training images than what is used at test time. 
This is obviously not the setting of the paper but it is worth discussing these results. 
A natural baseline for fine-to-coarse is to discard the fine labels and train only with the coarse labels induced by the fine annotation. 
This would yield the same accuracy as on the corresponding entry of the diagonal of the matrix. 
Irrespective of the method, the fine-to-coarse training does not necessarily outperform this simple strategy.

\begin{table*}
\caption{
Evaluation on iNaturalist-2018/2019 with all combinations of training / testing semantic levels. 
\emph{Left:} on-the-fly k-NN classification accuracy (top-1, \%)
\emph{Right:} category-level retrieval (mAP, \%).
We highlight the \colorbox{green}{best} and \colorbox{yellow}{second-best} result across methods for each train-test granularity combination. 
The diagonals (test = train granularity) are in {\bf bold}. 
Lower triangles are coarse-to-fine combinations, handled in the paper. 
Upper triangles (fine-to-coarse) are reported for reference but not formally addressed by our approach: 
better strategies would exploit the hierarchy of concepts more explicitly.\label{tab:INaturalist_Results}}

\newcommand{\inLrA}[1]{\cellcolor{green}#1}
\newcommand{\inLrB}[1]{\cellcolor{yellow}#1}
\newcommand{\inLrC}[1]{#1}
\newcommand{\inLrD}[1]{#1}
\newcommand{\inLrE}[1]{#1}
\newcommand{\inDrA}[1]{\cellcolor{green}{\bf #1}}
\newcommand{\inDrB}[1]{\cellcolor{yellow}{\bf #1}}
\newcommand{\inDrC}[1]{{\bf #1}}
\newcommand{\inDrD}[1]{{\bf #1}}
\newcommand{\inDrE}[1]{{\bf #1}}

\newcommand{\inUrA}[1]{\textcolor{mygray}{#1}}
\newcommand{\inUrB}[1]{\textcolor{mygray}{#1}}
\newcommand{\inUrC}[1]{\textcolor{mygray}{#1}}
\newcommand{\inUrD}[1]{\textcolor{mygray}{#1}}
\newcommand{\inUrE}[1]{\textcolor{mygray}{#1}}

\centering
\scalebox{0.65}{
{\small
\begin{tabular}{|ll|cccccccc|}
\toprule

&Test $\backslash$ Train & $\varnothing$ & King. & Phyl. & Class & Order & Fam. & Gen. & Spec. \\
\cmidrule{1-10} 
\multicolumn{10}{|c|}{iNaturalist-2018} \\
& \# classes: & 1 & 6 & 25 & 57 & 272 & 1118 & 4401 & 8142 \\
\cmidrule{1-10}

\multirow{7}{1em}{\rotatebox{90}{Baseline}}
&  Kingdom	& \inLrD{70.9}	& \inDrD{97.6}	& \inUrC{98.0}	& \inUrC{98.1}	& \inUrB{98.2}	& \inUrB{98.2}	& \inUrD{97.9}	& \inUrD{97.5}	\\
&  Phylum	& \inLrC{48.8}	& \inLrC{88.0}	& \inDrC{96.3}	& \inUrC{96.4}	& \inUrC{96.6}	& \inUrB{96.7}	& \inUrD{96.2}	& \inUrD{95.2}	\\
&  Class	& \inLrC{40.4}	& \inLrD{77.1}	& \inLrC{86.7}	& \inDrC{94.1}	& \inUrC{94.7}	& \inUrC{94.8}	& \inUrD{94.1}	& \inUrD{92.9}	\\
&  Order	& \inLrC{17.1}	& \inLrD{43.6}	& \inLrD{55.0}	& \inLrD{61.0}	& \inDrC{85.6}	& \inUrC{86.6}	& \inUrD{85.5}	& \inUrD{82.6}	\\
&  Family	& \inLrC{5.6}	& \inLrD{23.0}	& \inLrD{32.8}	& \inLrD{36.7}	& \inLrC{62.0}	& \inDrC{80.7}	& \inUrD{79.7}	& \inUrD{76.1}	\\
&  Genus	& \inLrC{0.9}	& \inLrD{10.0}	& \inLrD{17.3}	& \inLrD{20.1}	& \inLrC{41.7}	& \inLrC{63.0}	& \inDrD{72.5}	& \inUrD{68.3}	\\
&  Species	& \inLrC{0.3}	& \inLrD{6.3}	& \inLrD{11.5}	& \inLrD{13.6}	& \inLrD{31.2}	& \inLrA{51.3}	& \inLrB{61.6}	& \inDrD{60.2}	\\
\cmidrule{1-10}
\multirow{7}{1em}{\rotatebox{90}{SNCA+}}
&  Kingdom	& \inLrC{71.2}	& \inDrC{97.7}	& \inUrD{97.9}	& \inUrC{98.1}	& \inUrD{97.9}	& \inUrD{98.0}	& \inUrC{98.2}	& \inUrB{98.3}	\\
&  Phylum	& \inLrE{48.0}	& \inLrE{68.7}	& \inDrD{96.1}	& \inUrC{96.4}	& \inUrD{96.4}	& \inUrD{96.5}	& \inUrC{96.7}	& \inUrB{96.7}	\\
&  Class	& \inLrE{39.4}	& \inLrE{56.7}	& \inLrD{84.8}	& \inDrD{93.9}	& \inUrD{94.3}	& \inUrD{94.6}	& \inUrC{94.7}	& \inUrC{94.7}	\\
&  Order	& \inLrE{16.2}	& \inLrE{23.3}	& \inLrE{47.4}	& \inLrE{59.0}	& \inDrD{85.4}	& \inUrD{86.2}	& \inUrC{86.7}	& \inUrC{86.7}	\\
&  Family	& \inLrE{5.2}	& \inLrE{7.8}	& \inLrE{23.2}	& \inLrE{33.2}	& \inLrD{57.8}	& \inDrD{80.2}	& \inUrC{81.1}	& \inUrC{81.3}	\\
&  Genus	& \inLrC{0.9}	& \inLrE{1.3}	& \inLrE{10.4}	& \inLrE{17.6}	& \inLrE{36.9}	& \inLrD{56.8}	& \inDrB{74.2}	& \inUrC{74.1}	\\
&  Species	& \inLrC{0.3}	& \inLrE{0.5}	& \inLrE{6.3}	& \inLrE{11.9}	& \inLrE{26.2}	& \inLrD{42.4}	& \inLrD{58.9}	& \inDrC{64.6}	\\
\cmidrule{1-10}
\multirow{7}{1em}{\rotatebox{90}{ClusterFit+}}
&  Kingdom	& \inLrD{70.9}	& \inDrE{94.7}	& \inUrE{95.0}	& \inUrE{95.3}	& \inUrE{95.6}	& \inUrE{96.2}	& \inUrE{96.3}	& \inUrE{96.1}	\\
&  Phylum	& \inLrC{48.8}	& \inLrD{87.4}	& \inDrE{90.3}	& \inUrE{90.7}	& \inUrE{91.1}	& \inUrE{92.6}	& \inUrE{92.6}	& \inUrE{92.2}	\\
&  Class	& \inLrC{40.4}	& \inLrC{80.2}	& \inLrE{83.8}	& \inDrE{85.7}	& \inUrE{86.7}	& \inUrE{88.8}	& \inUrE{88.8}	& \inUrE{88.2}	\\
&  Order	& \inLrC{17.1}	& \inLrC{54.5}	& \inLrC{59.0}	& \inLrC{61.4}	& \inDrE{70.8}	& \inUrE{73.9}	& \inUrE{74.3}	& \inUrE{72.3}	\\
&  Family	& \inLrC{5.6}	& \inLrC{38.3}	& \inLrC{42.1}	& \inLrC{44.4}	& \inLrE{54.3}	& \inDrE{63.0}	& \inUrE{64.2}	& \inUrE{61.9}	\\
&  Genus	& \inLrC{0.9}	& \inLrB{26.7}	& \inLrC{29.5}	& \inLrC{31.5}	& \inLrD{40.1}	& \inLrE{49.4}	& \inDrE{53.9}	& \inUrE{51.7}	\\
&  Species	& \inLrC{0.3}	& \inLrA{21.8}	& \inLrC{23.7}	& \inLrC{25.2}	& \inLrC{32.7}	& \inLrE{40.3}	& \inLrE{44.7}	& \inDrE{43.4}	\\

\cmidrule{1-10}
\multirow{7}{1em}{\rotatebox{90}{\ours FC}}
&  Kingdom	& \inLrB{91.1}	& \inDrB{97.8}	& \inUrB{98.1}	& \inUrA{98.4}	& \inUrA{98.3}	& \inUrA{98.4}	& \inUrA{98.5}	& \inUrA{98.4}	\\
&  Phylum	& \inLrB{81.7}	& \inLrB{93.0}	& \inDrB{96.4}	& \inUrA{96.9}	& \inUrA{97.0}	& \inUrA{96.9}	& \inUrA{97.1}	& \inUrA{96.8}	\\
&  Class	& \inLrB{71.9}	& \inLrB{86.0}	& \inLrB{90.7}	& \inDrA{94.8}	& \inUrA{95.0}	& \inUrA{95.1}	& \inUrA{95.3}	& \inUrA{95.0}	\\
&  Order	& \inLrB{41.8}	& \inLrB{58.5}	& \inLrB{66.8}	& \inLrB{72.2}	& \inDrB{86.8}	& \inUrB{87.1}	& \inUrB{87.3}	& \inUrB{87.2}	\\
&  Family	& \inLrB{22.4}	& \inLrB{38.8}	& \inLrB{48.4}	& \inLrB{54.4}	& \inLrB{70.4}	& \inDrB{81.1}	& \inUrB{81.6}	& \inUrB{81.7}	\\
&  Genus	& \inLrB{11.4}	& \inLrC{24.6}	& \inLrB{33.1}	& \inLrB{38.6}	& \inLrB{53.0}	& \inLrB{63.9}	& \inDrC{73.8}	& \inUrB{74.2}	\\
&  Species	& \inLrB{8.13}	& \inLrC{18.8}	& \inLrA{25.6}	& \inLrB{29.9}	& \inLrB{41.5}	& \inLrC{50.9}	& \inLrC{60.9}	& \inDrB{65.9}	\\

\cmidrule{1-10}
\multirow{7}{1em}{\rotatebox{90}{\ours}}
&  Kingdom	& \inLrA{95.5}	& \inDrA{98.1}	& \inUrA{98.2}	& \inUrB{98.2}	& \inUrB{98.2}	& \inUrB{98.2}	& \inUrB{98.4}	& \inUrB{98.3}	\\
&  Phylum	& \inLrA{90.0}	& \inLrA{94.1}	& \inDrA{96.6}	& \inUrB{96.7}	& \inUrB{96.8}	& \inUrB{96.7}	& \inUrB{96.9}	& \inUrB{96.7}	\\
&  Class	& \inLrA{82.2}	& \inLrA{87.5}	& \inLrA{90.9}	& \inDrB{94.5}	& \inUrB{94.9}	& \inUrB{94.9}	& \inUrB{95.0}	& \inUrA{95.0}	\\
&  Order	& \inLrA{54.0}	& \inLrA{61.7}	& \inLrA{66.9}	& \inLrA{72.7}	& \inDrA{87.1}	& \inUrA{87.5}	& \inUrA{87.6}	& \inUrA{87.3}	\\
&  Family	& \inLrA{33.7}	& \inLrA{42.1}	& \inLrA{48.7}	& \inLrA{55.1}	& \inLrA{70.9}	& \inDrA{81.8}	& \inUrA{82.4}	& \inUrA{82.1}	\\
&  Genus	& \inLrA{20.5}	& \inLrA{27.0}	& \inLrA{33.5}	& \inLrA{39.5}	& \inLrA{54.2}	& \inLrA{64.6}	& \inDrA{75.6}	& \inUrA{75.5}	\\
&  Species	& \inLrA{15.9}	& \inLrB{20.4}	& \inLrB{25.5}	& \inLrA{30.8}	& \inLrA{42.7}	& \inLrB{51.2}	& \inLrA{61.9}	& \inDrA{67.7}	\\

\midrule

\multicolumn{10}{|c|}{iNaturalist-2019} \\
& \# classes: & 1 & 3 & 4 & 9 & 34 & 57 & 72 & 1010 \\
\cmidrule{1-10}

\multirow{7}{1em}{\rotatebox{90}{Baseline}}
&  Kingdom	& \inLrC{77.0}	& \inDrB{98.9}	& \inUrC{98.9}	& \inUrD{99.0}	& \inUrA{99.3}	& \inUrA{99.4}	& \inUrB{99.3}	& \inUrD{98.9}	\\
&  Phylum	& \inLrC{73.8}	& \inLrC{97.1}	& \inDrD{98.7}	& \inUrD{98.9}	& \inUrA{99.2}	& \inUrB{99.2}	& \inUrA{99.2}	& \inUrC{98.7}	\\
&  Class	& \inLrC{63.3}	& \inLrD{87.6}	& \inLrC{90.3}	& \inDrD{98.0}	& \inUrB{98.5}	& \inUrB{98.6}	& \inUrB{98.6}	& \inUrB{98.0}	\\
&  Order	& \inLrC{17.9}	& \inLrD{49.6}	& \inLrD{56.4}	& \inLrD{70.8}	& \inDrC{95.6}	& \inUrB{95.5}	& \inUrA{96.0}	& \inUrB{95.2}	\\
&  Family	& \inLrC{12.4}	& \inLrD{42.1}	& \inLrD{50.4}	& \inLrD{65.0}	& \inLrC{89.4}	& \inDrB{94.8}	& \inUrB{95.1}	& \inUrB{94.4}	\\
&  Genus	& \inLrC{9.6}	& \inLrD{39.2}	& \inLrD{46.5}	& \inLrD{62.1}	& \inLrC{86.1}	& \inLrC{91.5}	& \inDrB{94.8}	& \inUrB{93.9}	\\
&  Species	& \inLrC{1.5}	& \inLrD{9.8}	& \inLrD{13.5}	& \inLrD{20.6}	& \inLrD{34.5}	& \inLrD{39.9}	& \inLrC{42.4}	& \inDrC{75.0}	\\
\cmidrule{1-10}
\multirow{7}{1em}{\rotatebox{90}{SNCA+}}
&  Kingdom	& \inLrE{76.9}	& \inDrD{98.6}	& \inUrC{98.9}	& \inUrA{99.2}	& \inUrB{99.2}	& \inUrC{99.3}	& \inUrC{99.1}	& \inUrB{99.0}	\\
&  Phylum	& \inLrE{73.3}	& \inLrE{87.1}	& \inDrC{98.8}	& \inUrA{99.1}	& \inUrB{99.1}	& \inUrC{99.1}	& \inUrC{98.9}	& \inUrA{99.0}	\\
&  Class	& \inLrE{62.3}	& \inLrE{74.9}	& \inLrE{84.1}	& \inDrA{98.2}	& \inUrA{98.6}	& \inUrC{98.3}	& \inUrC{98.1}	& \inUrC{97.8}	\\
&  Order	& \inLrE{17.6}	& \inLrE{19.7}	& \inLrE{30.2}	& \inLrE{55.4}	& \inDrD{95.3}	& \inUrD{95.2}	& \inUrD{95.2}	& \inUrD{94.2}	\\
&  Family	& \inLrE{12.2}	& \inLrE{12.7}	& \inLrE{20.7}	& \inLrE{45.5}	& \inLrD{88.2}	& \inDrC{94.5}	& \inUrC{94.6}	& \inUrD{93.5}	\\
&  Genus	& \inLrE{9.3}	& \inLrE{9.2}	& \inLrE{17.1}	& \inLrE{41.6}	& \inLrD{85.0}	& \inLrD{91.2}	& \inDrD{94.0}	& \inUrD{93.1}	\\
&  Species	& \inLrE{1.3}	& \inLrE{1.0}	& \inLrE{1.8}	& \inLrE{10.4}	& \inLrC{36.0}	& \inLrC{40.8}	& \inLrD{42.3}	& \inDrD{74.7}	\\
\cmidrule{1-10}
\multirow{7}{1em}{\rotatebox{90}{ClusterFit+}}
&  Kingdom	& \inLrC{77.0}	& \inDrE{96.4}	& \inUrE{96.1}	& \inUrE{95.8}	& \inUrE{95.7}	& \inUrE{95.7}	& \inUrE{95.4}	& \inUrE{97.0}	\\
&  Phylum	& \inLrC{73.8}	& \inLrD{94.2}	& \inDrE{95.0}	& \inUrE{94.6}	& \inUrE{94.3}	& \inUrE{94.4}	& \inUrE{93.8}	& \inUrE{95.5}	\\
&  Class	& \inLrC{63.3}	& \inLrC{88.7}	& \inLrD{90.1}	& \inDrE{91.3}	& \inUrE{90.1}	& \inUrE{90.9}	& \inUrE{90.6}	& \inUrE{93.5}	\\
&  Order	& \inLrC{17.9}	& \inLrC{65.5}	& \inLrC{67.9}	& \inLrC{70.9}	& \inDrE{76.8}	& \inUrE{79.0}	& \inUrE{78.1}	& \inUrE{83.2}	\\
&  Family	& \inLrC{12.4}	& \inLrC{59.5}	& \inLrC{62.0}	& \inLrC{65.4}	& \inLrE{71.7}	& \inDrE{75.6}	& \inUrE{75.3}	& \inUrE{80.4}	\\
&  Genus	& \inLrC{9.6}	& \inLrC{56.9}	& \inLrC{59.3}	& \inLrC{62.7}	& \inLrE{68.7}	& \inLrE{72.6}	& \inDrE{73.9}	& \inUrE{78.6}	\\
&  Species	& \inLrC{1.5}	& \inLrC{24.5}	& \inLrC{25.6}	& \inLrC{27.3}	& \inLrE{31.1}	& \inLrE{33.6}	& \inLrE{33.9}	& \inDrE{49.6}	\\
\cmidrule{1-10}
\multirow{7}{1em}{\rotatebox{90}{\ours FC}}
&  Kingdom	& \inLrB{93.1}	& \inDrB{98.9}	& \inUrB{99.0}	& \inUrA{99.2}	& \inUrB{99.2}	& \inUrA{99.4}	& \inUrA{99.4}	& \inUrA{99.2}	\\
&  Phylum	& \inLrB{90.9}	& \inLrB{98.2}	& \inDrA{98.9}	& \inUrA{99.1}	& \inUrB{99.1}	& \inUrA{99.3}	& \inUrA{99.2}	& \inUrA{99.0}	\\
&  Class	& \inLrB{82.6}	& \inLrB{94.9}	& \inLrB{96.4}	& \inDrA{98.2}	& \inUrD{98.3}	& \inUrA{98.7}	& \inUrA{98.7}	& \inUrA{98.3}	\\
&  Order	& \inLrB{52.5}	& \inLrB{80.0}	& \inLrB{83.5}	& \inLrB{89.5}	& \inDrB{95.8}	& \inUrA{96.0}	& \inUrB{95.9}	& \inUrA{95.3}	\\
&  Family	& \inLrB{45.3}	& \inLrB{74.6}	& \inLrB{78.9}	& \inLrB{86.0}	& \inLrA{93.4}	& \inDrA{95.2}	& \inUrA{95.4}	& \inUrA{94.7}	\\
&  Genus	& \inLrB{41.7}	& \inLrB{71.7}	& \inLrB{76.3}	& \inLrB{84.0}	& \inLrA{91.7}	& \inLrA{93.4}	& \inDrA{95.0}	& \inUrA{94.3}	\\
&  Species	& \inLrB{12.0}	& \inLrB{29.5}	& \inLrB{32.6}	& \inLrB{40.4}	& \inLrA{51.8}	& \inLrB{53.2}	& \inLrB{53.9}	& \inDrA{75.9}	\\
\cmidrule{1-10}
\multirow{7}{1em}{\rotatebox{90}{\ours}}
&  Kingdom	& \inLrA{96.9}	& \inDrA{99.2}	& \inUrA{99.1}	& \inUrA{99.2}	& \inUrB{99.2}	& \inUrD{99.0}	& \inUrD{99.0}	& \inUrB{99.0}	\\
&  Phylum	& \inLrA{96.4}	& \inLrA{98.8}	& \inDrA{98.9}	& \inUrC{99.0}	& \inUrD{99.0}	& \inUrD{98.9}	& \inUrC{98.9}	& \inUrC{98.7}	\\
&  Class	& \inLrA{93.0}	& \inLrA{97.0}	& \inLrA{97.1}	& \inDrA{98.2}	& \inUrC{98.4}	& \inUrC{98.3}	& \inUrC{98.1}	& \inUrC{97.8}	\\
&  Order	& \inLrA{81.3}	& \inLrA{89.0}	& \inLrA{89.3}	& \inLrA{91.2}	& \inDrA{95.9}	& \inUrC{95.3}	& \inUrC{95.3}	& \inUrC{94.5}	\\
&  Family	& \inLrA{76.5}	& \inLrA{85.2}	& \inLrA{85.2}	& \inLrA{87.8}	& \inLrB{93.1}	& \inDrC{94.5}	& \inUrD{94.5}	& \inUrC{93.8}	\\
&  Genus	& \inLrA{73.8}	& \inLrA{82.7}	& \inLrA{83.1}	& \inLrA{85.8}	& \inLrB{91.3}	& \inLrB{92.6}	& \inDrC{94.2}	& \inUrC{93.4}	\\
&  Species	& \inLrA{31.0}	& \inLrA{41.6}	& \inLrA{41.4}	& \inLrA{46.0}	& \inLrA{51.8}	& \inLrA{53.5}	& \inLrA{55.3}	& \inDrB{75.3}	\\

\bottomrule
\end{tabular}
\hspace{2cm}

\begin{tabular}{|ll|ccccccc|}
\toprule

&Test $\backslash$ Train &  King. & Phyl. & Class & Order & Fam. & Gen. & Spec. \\
\cmidrule{1-9} 
\multicolumn{9}{|c|}{iNaturalist-2018} \\
& \# classes: &  6 & 25 & 57 & 272 & 1118 & 4401 & 8142 \\
\cmidrule{1-9}

\multirow{7}{1em}{\rotatebox{90}{Baseline}}
&  Kingdom	& \inDrC{97.8}	& \inUrC{86.3}	& \inUrA{81.0}	& \inUrA{76.4}	& \inUrA{65.9}	& \inUrA{62.1}	& \inUrA{61.5}	\\
&  Phylum	& \inLrC{64.2}	& \inDrC{96.6}	& \inUrB{82.1}	& \inUrA{63.1}	& \inUrA{45.9}	& \inUrA{39.8}	& \inUrA{38.1}	\\
&  Class	& \inLrC{46.0}	& \inLrD{72.1}	& \inDrC{93.8}	& \inUrA{60.1}	& \inUrA{39.2}	& \inUrA{31.2}	& \inUrA{28.5}	\\
&  Order	& \inLrC{12.2}	& \inLrD{24.1}	& \inLrD{34.3}	& \inDrD{74.5}	& \inUrC{35.4}	& \inUrC{20.1}	& \inUrC{15.6}	\\
&  Family	& \inLrC{3.69}	& \inLrC{7.02}	& \inLrD{10.1}	& \inLrD{32.6}	& \inDrD{51.3}	& \inUrD{20.9}	& \inUrD{14.5}	\\
&  Genus	& \inLrD{1.30}	& \inLrD{3.06}	& \inLrE{4.47}	& \inLrD{16.6}	& \inLrD{30.4}	& \inDrD{33.3}	& \inUrD{24.0}	\\
&  Species	& \inLrD{1.18}	& \inLrD{2.63}	& \inLrE{3.63}	& \inLrC{12.8}	& \inLrD{25.7}	& \inLrD{31.4}	& \inDrD{27.9}	\\
\cmidrule{1-9}
\multirow{7}{1em}{\rotatebox{90}{SNCA+}}
&  Kingdom	& \inDrD{97.6}	& \inUrD{83.3}	& \inUrD{75.9}	& \inUrD{59.2}	& \inUrE{56.0}	& \inUrE{54.9}	& \inUrE{55.0}	\\
&  Phylum	& \inLrD{59.8}	& \inDrD{91.7}	& \inUrD{79.4}	& \inUrD{49.1}	& \inUrD{35.0}	& \inUrE{32.3}	& \inUrE{32.2}	\\
&  Class	& \inLrD{41.3}	& \inLrC{73.1}	& \inDrD{89.9}	& \inUrD{49.2}	& \inUrD{28.1}	& \inUrE{23.6}	& \inUrE{23.0}	\\
&  Order	& \inLrD{9.09}	& \inLrC{24.9}	& \inLrC{35.7}	& \inDrC{77.9}	& \inUrD{35.3}	& \inUrD{18.0}	& \inUrD{15.0}	\\
&  Family	& \inLrE{2.24}	& \inLrD{6.43}	& \inLrC{11.2}	& \inLrC{35.7}	& \inDrC{68.4}	& \inUrC{29.1}	& \inUrC{21.7}	\\
&  Genus	& \inLrE{0.39}	& \inLrE{2.47}	& \inLrC{5.03}	& \inLrC{18.1}	& \inLrC{36.6}	& \inDrC{60.5}	& \inUrC{46.0}	\\
&  Species	& \inLrE{0.19}	& \inLrE{1.86}	& \inLrD{3.80}	& \inLrC{12.8}	& \inLrC{26.4}	& \inLrC{46.0}	& \inDrC{54.9}	\\
\cmidrule{1-9}
\multirow{7}{1em}{\rotatebox{90}{ClusterFit+}}
&  Kingdom	& \inDrE{55.5}	& \inUrE{55.5}	& \inUrE{55.7}	& \inUrE{56.4}	& \inUrD{57.0}	& \inUrB{57.6}	& \inUrB{57.7}	\\
&  Phylum	& \inLrE{31.6}	& \inDrE{32.1}	& \inUrE{32.1}	& \inUrE{32.4}	& \inUrE{33.1}	& \inUrC{33.9}	& \inUrB{34.0}	\\
&  Class	& \inLrE{21.0}	& \inLrE{21.6}	& \inDrE{22.0}	& \inUrE{22.2}	& \inUrE{23.0}	& \inUrD{23.7}	& \inUrD{23.8}	\\
&  Order	& \inLrE{6.8}	& \inLrE{7.4}	& \inLrE{7.8}	& \inDrE{9.4}	& \inUrE{9.9}	& \inUrE{10.3}	& \inUrE{10.1}	\\
&  Family	& \inLrD{2.9}	& \inLrE{3.5}	& \inLrE{3.9}	& \inLrE{5.5}	& \inDrE{7.8}	& \inUrE{7.9}	& \inUrE{7.3}	\\
&  Genus	& \inLrC{3.6}	& \inLrC{4.3}	& \inLrD{4.8}	& \inLrE{7.1}	& \inLrE{10.8}	& \inDrE{13.3}	& \inUrE{12.0}	\\
&  Species	& \inLrC{4.7}	& \inLrC{5.4}	& \inLrC{5.9}	& \inLrE{8.6}	& \inLrE{12.5}	& \inLrE{15.3}	& \inDrE{14.5}	\\
\cmidrule{1-9}
\multirow{7}{1em}{\rotatebox{90}{\ours FC}}
&  Kingdom	& \inDrB{98.5}	& \inUrA{88.3}	& \inUrB{80.6}	& \inUrB{61.6}	& \inUrC{57.7}	& \inUrC{56.0}	& \inUrC{56.0}	\\
&  Phylum	& \inLrA{69.6}	& \inDrA{97.2}	& \inUrA{83.1}	& \inUrC{50.5}	& \inUrC{37.9}	& \inUrC{33.9}	& \inUrC{33.3}	\\
&  Class	& \inLrA{52.4}	& \inLrA{75.8}	& \inDrA{95.7}	& \inUrB{51.3}	& \inUrC{31.3}	& \inUrC{25.5}	& \inUrB{24.5}	\\
&  Order	& \inLrA{18.6}	& \inLrA{31.4}	& \inLrB{41.6}	& \inDrB{88.0}	& \inUrB{41.6}	& \inUrB{20.4}	& \inUrA{16.4}	\\
&  Family	& \inLrB{7.68}	& \inLrB{12.9}	& \inLrB{17.7}	& \inLrA{44.7}	& \inDrB{82.4}	& \inUrB{33.4}	& \inUrB{23.6}	\\
&  Genus	& \inLrB{4.97}	& \inLrB{8.82}	& \inLrB{11.9}	& \inLrB{27.3}	& \inLrB{45.0}	& \inDrB{75.5}	& \inUrB{52.2}	\\
&  Species	& \inLrB{4.95}	& \inLrB{8.25}	& \inLrB{10.7}	& \inLrB{21.4}	& \inLrB{33.6}	& \inLrB{53.8}	& \inDrB{68.1}	\\
\cmidrule{1-9}
\multirow{7}{1em}{\rotatebox{90}{\ours}}
&  Kingdom	& \inDrA{98.6}	& \inUrA{88.3}	& \inUrC{79.7}	& \inUrC{60.8}	& \inUrB{58.0}	& \inUrD{55.9}	& \inUrD{55.5}	\\
&  Phylum	& \inLrB{67.8}	& \inDrA{97.2}	& \inUrB{82.1}	& \inUrB{50.9}	& \inUrB{38.9}	& \inUrB{34.2}	& \inUrD{33.0}	\\
&  Class	& \inLrB{50.1}	& \inLrB{74.9}	& \inDrB{95.4}	& \inUrC{51.2}	& \inUrB{32.3}	& \inUrB{25.9}	& \inUrC{24.1}	\\
&  Order	& \inLrB{17.7}	& \inLrB{30.7}	& \inLrA{42.7}	& \inDrA{88.3}	& \inUrA{42.3}	& \inUrA{21.1}	& \inUrB{16.2}	\\
&  Family	& \inLrA{8.70}	& \inLrA{13.2}	& \inLrA{18.0}	& \inLrB{43.9}	& \inDrA{83.1}	& \inUrA{34.8}	& \inUrA{24.2}	\\
&  Genus	& \inLrA{6.78}	& \inLrA{9.72}	& \inLrA{13.5}	& \inLrA{29.0}	& \inLrA{46.9}	& \inDrA{77.2}	& \inUrA{53.9}	\\
&  Species	& \inLrA{6.45}	& \inLrA{9.02}	& \inLrA{12.1}	& \inLrA{23.6}	& \inLrA{35.6}	& \inLrA{55.4}	& \inDrA{70.0}	\\

\midrule

\multicolumn{9}{|c|}{iNaturalist-2019} \\
& \# classes: & 3 & 4 & 9 & 34 & 57 & 72 & 1010 \\
\cmidrule{1-9}

\multirow{7}{1em}{\rotatebox{90}{Baseline}}
&  Kingdom	& \inDrC{99.0}	& \inUrA{98.2}	& \inUrA{88.9}	& \inUrA{73.6}	& \inUrA{65.8}	& \inUrA{67.4}	& \inUrA{58.6}	\\
&  Phylum	& \inLrC{87.1}	& \inDrC{98.9}	& \inUrA{90.8}	& \inUrA{71.7}	& \inUrA{59.8}	& \inUrA{61.6}	& \inUrA{51.7}	\\
&  Class	& \inLrC{67.2}	& \inLrC{77.6}	& \inDrC{98.2}	& \inUrA{68.8}	& \inUrA{55.1}	& \inUrA{56.3}	& \inUrA{42.8}	\\
&  Order	& \inLrC{15.1}	& \inLrC{21.1}	& \inLrC{33.7}	& \inDrC{94.8}	& \inUrC{68.6}	& \inUrC{57.6}	& \inUrA{26.2}	\\
&  Family	& \inLrC{9.72}	& \inLrC{13.8}	& \inLrC{24.2}	& \inLrD{70.7}	& \inDrC{94.2}	& \inUrC{80.6}	& \inUrA{31.5}	\\
&  Genus	& \inLrC{7.77}	& \inLrC{11.0}	& \inLrC{21.3}	& \inLrD{59.6}	& \inLrD{81.4}	& \inDrC{93.9}	& \inUrA{34.8}	\\
&  Species	& \inLrD{1.09}	& \inLrD{1.55}	& \inLrC{3.60}	& \inLrD{10.8}	& \inLrD{14.8}	& \inLrD{16.6}	& \inDrD{57.0}	\\
\cmidrule{1-9}
\multirow{7}{1em}{\rotatebox{90}{SNCA+}}
&  Kingdom	& \inDrD{98.4}	& \inUrD{90.1}	& \inUrD{82.0}	& \inUrC{63.5}	& \inUrD{60.9}	& \inUrD{60.3}	& \inUrE{55.0}	\\
&  Phylum	& \inLrD{84.1}	& \inDrD{97.7}	& \inUrD{87.7}	& \inUrC{62.6}	& \inUrD{55.9}	& \inUrD{55.3}	& \inUrD{49.3}	\\
&  Class	& \inLrD{63.2}	& \inLrD{75.6}	& \inDrD{95.5}	& \inUrD{59.0}	& \inUrD{50.0}	& \inUrD{49.1}	& \inUrD{38.5}	\\
&  Order	& \inLrD{11.5}	& \inLrD{17.2}	& \inLrD{32.4}	& \inDrD{83.0}	& \inUrD{64.3}	& \inUrD{54.4}	& \inUrD{15.7}	\\
&  Family	& \inLrD{6.53}	& \inLrD{10.0}	& \inLrD{20.1}	& \inLrC{75.2}	& \inDrD{90.9}	& \inUrD{78.8}	& \inUrD{19.5}	\\
&  Genus	& \inLrD{5.08}	& \inLrD{7.61}	& \inLrD{18.1}	& \inLrC{71.5}	& \inLrC{84.6}	& \inDrD{92.8}	& \inUrD{22.0}	\\
&  Species	& \inLrE{0.40}	& \inLrE{0.65}	& \inLrE{2.11}	& \inLrC{15.4}	& \inLrC{17.1}	& \inLrC{18.6}	& \inDrC{72.3}	\\
\cmidrule{1-9}
\multirow{7}{1em}{\rotatebox{90}{ClusterFit+}}
&  Kingdom	& \inDrE{55.1}	& \inUrE{55.0}	& \inUrE{54.7}	& \inUrE{54.4}	& \inUrE{54.5}	& \inUrE{54.6}	& \inUrD{55.5}	\\
&  Phylum	& \inLrE{49.0}	& \inDrE{49.1}	& \inUrE{48.8}	& \inUrE{48.2}	& \inUrE{48.3}	& \inUrE{48.4}	& \inUrD{49.3}	\\
&  Class	& \inLrE{36.8}	& \inLrE{36.9}	& \inDrE{37.1}	& \inUrE{36.3}	& \inUrE{36.4}	& \inUrE{36.5}	& \inUrE{37.6}	\\
&  Order	& \inLrE{7.5}	& \inLrE{7.7}	& \inLrE{8.2}	& \inDrE{8.4}	& \inUrE{8.5}	& \inUrE{8.4}	& \inUrE{9.7}	\\
&  Family	& \inLrE{5.6}	& \inLrE{5.9}	& \inLrE{6.4}	& \inLrE{6.8}	& \inDrE{7.4}	& \inUrE{7.3}	& \inUrE{8.9}	\\
&  Genus	& \inLrE{4.9}	& \inLrE{5.2}	& \inLrE{5.8}	& \inLrE{6.1}	& \inLrE{6.8}	& \inDrE{6.9}	& \inUrE{8.6}	\\
&  Species	& \inLrC{2.4}	& \inLrC{2.5}	& \inLrD{2.9}	& \inLrE{3.5}	& \inLrE{4.1}	& \inLrE{4.2}	& \inDrE{10.1}	\\
\cmidrule{1-9}
\multirow{7}{1em}{\rotatebox{90}{\ours FC}}
&  Kingdom	& \inDrB{99.2}	& \inUrB{93.2}	& \inUrB{86.5}	& \inUrB{63.8}	& \inUrB{62.3}	& \inUrB{62.2}	& \inUrC{56.1}	\\
&  Phylum	& \inLrB{88.6}	& \inDrA{99.2}	& \inUrB{90.7}	& \inUrB{63.0}	& \inUrB{58.9}	& \inUrB{57.9}	& \inUrC{50.5}	\\
&  Class	& \inLrB{70.4}	& \inLrB{80.9}	& \inDrB{98.5}	& \inUrB{61.6}	& \inUrB{53.9}	& \inUrB{52.5}	& \inUrC{39.9}	\\
&  Order	& \inLrB{25.1}	& \inLrB{32.6}	& \inLrB{45.4}	& \inDrB{96.3}	& \inUrA{70.3}	& \inUrA{58.6}	& \inUrC{18.4}	\\
&  Family	& \inLrB{20.8}	& \inLrB{26.7}	& \inLrB{38.2}	& \inLrA{84.5}	& \inDrA{95.8}	& \inUrA{82.2}	& \inUrC{23.0}	\\
&  Genus	& \inLrB{18.9}	& \inLrB{24.7}	& \inLrB{34.0}	& \inLrA{78.3}	& \inLrA{88.4}	& \inDrA{95.7}	& \inUrC{25.7}	\\
&  Species	& \inLrB{6.83}	& \inLrB{9.63}	& \inLrB{14.7}	& \inLrB{28.4}	& \inLrB{29.7}	& \inLrB{31.5}	& \inDrA{78.4}	\\
\cmidrule{1-9}
\multirow{7}{1em}{\rotatebox{90}{\ours}}
&  Kingdom	& \inDrA{99.4}	& \inUrC{93.1}	& \inUrC{85.9}	& \inUrD{62.7}	& \inUrC{61.5}	& \inUrC{60.9}	& \inUrB{56.6}	\\
&  Phylum	& \inLrA{88.8}	& \inDrA{99.2}	& \inUrC{90.2}	& \inUrC{62.6}	& \inUrC{58.4}	& \inUrC{57.2}	& \inUrB{51.0}	\\
&  Class	& \inLrA{71.8}	& \inLrA{81.9}	& \inDrA{98.6}	& \inUrC{61.3}	& \inUrC{53.2}	& \inUrC{52.0}	& \inUrB{40.4}	\\
&  Order	& \inLrA{30.7}	& \inLrA{36.6}	& \inLrA{48.1}	& \inDrA{96.4}	& \inUrB{69.3}	& \inUrB{58.3}	& \inUrB{19.1}	\\
&  Family	& \inLrA{28.2}	& \inLrA{30.8}	& \inLrA{41.8}	& \inLrB{82.5}	& \inDrB{95.1}	& \inUrB{81.5}	& \inUrB{23.7}	\\
&  Genus	& \inLrA{28.0}	& \inLrA{29.8}	& \inLrA{40.5}	& \inLrB{76.2}	& \inLrB{87.5}	& \inDrB{94.8}	& \inUrB{26.3}	\\
&  Species	& \inLrA{18.7}	& \inLrA{18.9}	& \inLrA{21.8}	& \inLrA{32.5}	& \inLrA{33.3}	& \inLrA{34.7}	& \inDrB{77.5}	\\

\bottomrule
\end{tabular}
 
}}
\end{table*}

\subsection{Transfer Learning Tasks}
\label{app:tl}

This section details the experimental settings for the transfer learning and reports more results and comparisons.

\paragraph{Fine-tuning settings}
As described in Section~\ref{subsec:transfer_learning_task} 
We initialize the network trunk with ImageNet pre-trained weights and fine-tune only the classifier. 
For our method, the pre-trained network trunk $f_\theta$ remains fixed. 
The projector $P_\theta$ is discarded. 
For all methods we fine-tune during 240 epochs with a cosine learning rate schedule starting at 0.01, and batches of 512 images.

For fine-tuning results in Table~\ref{tab:transfer_learning_tasks_sota}  we additionally use Cutmix~\cite{Yun2019CutMix} and FixRes~\cite{Touvron2019FixRes} during fine-tuning and we fine-tune with more epochs (1000 for Flowers~\cite{Nilsback08} and Cars~\cite{Cars2013}, 300 for Food-101~\cite{bossard14Food101} and iNaturalist~\cite{Horn2018INaturalist,Horn2019INaturalist}). These choices improve the performance for all the methods.

\paragraph{Results.}
Table~\ref{tab:transfer_learning_task_diverse_arch} compares the performance obtained with \ours for different architectures. 
We report results with \ours topped with either a multi-layers perceptron (MLP) or a linear classifier (FC).
The accuracy increases for larger models. 
This shows that, although ResNet-50 serves as a running example architecture for \ours, the method applies without modifications to other architectures. 

Table~\ref{tab:transfer_learning_tasks} compares the performances= obtained with \ours and baselines with MLP and FC classifier.
For all settings, the flexibility of the MLP is useful to outperform the linear classifier (FC). 
The transfer learning results are better or as good for \ours variants. 
The gap with the baseline methods is higher for the iNaturalist variants. 
This is because the datasets are more challenging, as evidenced by the relatively low accuracies reported.

\section{Visualization}
\label{app:visu}

\paragraph{CIFAR.}

Figure~\ref{fig:sampleranking} shows for a giving test image in CIFAR-100 the 10 nearest neighbours in the training according the cosine similarity in the embedding space.
In Figure~\ref{fig:sampleranking} models  are trained on the 20 CIFAR-100 coarse classes.
The correct classes are indicated in green.
For example, in the first row, \ours correctly identifies a butterfly query in 9 out of 10 results, while the baseline method succeeds only 5 times. 
The second row is a relative failure case, because \ours confuses a van with a pickup truck. 
However, it correctly matches the colors of the vehicles. 

\paragraph{iNaturalist.}

Figure~\ref{fig:inat_ranking_evolution_appendix_1}, ~\ref{fig:inat_ranking_evolution_appendix_2} and Figure~\ref{fig:inat_ranking_evolution_appendix_3} present similar results for three examples for iNaturalist-2018, but with several levels of granularity for the training set, which allow one to vizualize the importance of the training granularity as well.  
Each granularity level is identified with a color. 
The frame color around the image indicates which at which granularity the match is correct: for example, light orange means it is correct at the order level and green means that the result is correct at the finest granularity (Species).

We can observe from the colors and the image content that the level at which \ours is correct is almost systematically better than the baseline\footnote{These examples are representative of typical comparisons, as we have not cherry-picked to show cases where our method is better. }. 
For example, the baseline model trained at the genus granularity in Figure~\ref{fig:inat_ranking_evolution_appendix_1} matches the deer query with a moose (rank 3). 

In Figure~\ref{fig:inat_ranking_evolution_appendix_2}, the butterfly is matched relatively easily with other butterflies by both classifiers, even when they are trained with coarse granularity. 
This is because butterflies have quite distinctive textures. 
However, \ours slightly outperforms the baseline for finer granularity levels. 

Figure~\ref{fig:inat_ranking_evolution_appendix_3} shows an orca query, which is quite distinctive with its black-and-white skin. 
The baseline method is unable to distinguish it from other marine mammals, even when trained at the finest granularity. 
\ours is able to distinguish these textures more accurately, so it gets perfect retrieval results even when trained at the genus granularity.

\begin{table*}
\caption{\label{tab:transfer_learning_task_diverse_arch}
 Transfer learning task with various architectures pretrained on ImageNet with \ours.
We report the Top-1 accuracy (\%) for the evaluation with a single center crop at resolution $224 \times 224$.
}
\centering
\newcommand{\reportarch}[2]{\rotatebox{90}{\begin{minipage}{2.1cm}#1\\#2\end{minipage}}}
\scalebox{0.75}{
\begin{tabular}{|l|cc|cc|cc|cc|cc|}
  \toprule
   & 
   \multicolumn{2}{c|}{\reportarch{ResNeXt50-32x4}{\cite{Xie2017AggregatedRT}}} & 
   \multicolumn{2}{c|}{\reportarch{ResNeXt50D-32x4}{\cite{Tong2018BagofTricks}}} & 
   \multicolumn{2}{c|}{\reportarch{ResNeXt50D-32x8}{\cite{Tong2018BagofTricks}}} & 
   \multicolumn{2}{c|}{\reportarch{RegNety-4GF}{\cite{Radosavovic2020RegNet}}} &  
   \multicolumn{2}{c|}{\reportarch{RegNety-8GF}{\cite{Radosavovic2020RegNet}}} \\
\midrule
\# Params & \multicolumn{2}{c|}{25M} & \multicolumn{2}{c|}{25M} & \multicolumn{2}{c|}{48M} & \multicolumn{2}{c|}{21M} & \multicolumn{2}{c|}{39M}\\
\midrule
Dataset & FC & MLP & FC & MLP  & FC & MLP  & FC & MLP  & FC & MLP   \\

\midrule
	Flowers-102~\cite{Nilsback08}               & 95.5 & \textbf{98.3}  & 95.9 & \textbf{98.6} & 96.3 & \textbf{98.7} & 98.1 & \textbf{98.6}  & \textbf{99.0} & 98.8\\
	Stanford Cars~\cite{Cars2013}               & 91.6 & \textbf{92.9}  & 88.7 & \textbf{93.3}  & 90.9 & \textbf{93.8}   & \textbf{93.3} & 92.7  & \textbf{94.0} & 93.4\\ 
	Food101~\cite{bossard14Food101}             & 89.6 & \textbf{89.9} & 90.2 & \textbf{90.3}  & 90.9 & \textbf{91.1}  & 91.2 & 91.3 & 92.1 & \textbf{92.4}\\
	iNaturalist 2018~\cite{Horn2018INaturalist}&  67.7 & \textbf{71.2}  & 68.9 & \textbf{72.4} & 71.4 & \textbf{74.4}  & 73.8 & \textbf{74.2} & 76.4 & \textbf{76.8}\\
	iNaturalist 2019~\cite{Horn2019INaturalist} & 75.3 & \textbf{76.3} & 75.8 & \textbf{77.6}  & 77.8 & \textbf{78.7}   & \textbf{78.1} & 77.9 & 79.8 & \textbf{80.0}\\

  \bottomrule
\end{tabular}}
\end{table*}

    \begin{table*}
    \caption{\label{tab:transfer_learning_tasks}
     Transfer learning with ResNet-50 pretrained on ImageNet. 
     Comparison between different pre-training  methods and two different classifiers trained on the target domain (a linear FC or an MLP).
    We report the top-1 accuracy (\%) with a single center crop evaluation at resolution $224 \times 224$.
    }
    \centering
    \scalebox{0.75}{
    \begin{tabular}{|@{\ }@{\ }l|cccccc|cc|cc|}
      \toprule
       \multirow{2}{3em}{Dataset} & \multicolumn{2}{c}{\rotatebox{90}{Baseline}} & \multicolumn{2}{c}{\rotatebox{90}{SNCA+}} & \multicolumn{2}{c}{\rotatebox{90}{ClusterFit+}}  & \multicolumn{2}{|c|}{\rotatebox{90}{\ours}} &  \multicolumn{2}{|c|}{\rotatebox{90}{\ours FC}} \\
   
    & FC & MLP  & FC & MLP  & FC & MLP  & FC & MLP & FC & MLP \\
    \midrule
    	Flowers-102~\cite{Nilsback08}       & 96.2 & 95.7  & 94.3 & \textbf{98.2} & 96.2 & 96.1     & 97.6  & \textbf{98.2} & 94.3 & 97.6 \\ 
    	Stanford Cars~\cite{Cars2013}              & 90.0 & 89.8  & 91.6  & \textbf{92.5}   & 89.4 & 89.3   & 91.4 & \textbf{92.5}  & 91.4 & \textbf{92.7} \\ 
    	Food101~\cite{bossard14Food101}            & 88.2 & 88.9   & 88.7 & 88.8   & 88.5  & 88.9 & 88.9 & \textbf{89.5} & 88.5 & 88.7\\ 
    	iNaturalist 2018~\cite{Horn2018INaturalist}& 65.0 & 68.4     & 64.7  & 69.2 & 64.2 & 67.5  & 65.6 & \textbf{69.8} & 65.2 & 68.5\\ 
    	iNaturalist 2019~\cite{Horn2019INaturalist}& 72.8 & 73.7   & 73.1 & 74.5   & 71.8 & 73.8   & 74.1 & \textbf{75.9} & 73.9 & 74.6  \\ 
    
      \bottomrule
    \end{tabular}}
    \end{table*}
    
\newcommand{\imCifour}[1]{\includegraphics[width=0.08\linewidth]{figs/cifar_100_images/train_images_our/#1}}
\newcommand{\imCifbas}[1]{\includegraphics[width=0.08\linewidth]{figs/cifar_100_images/train_images_baseline/#1}}
\newcommand{\imCiftest}[1]{\includegraphics[width=0.08\linewidth]{figs/cifar_100_images/test_images/#1}}
\fboxrule=1pt%
\fboxsep=0.1mm

\def \mysp {\hspace{2pt}}

\begin{figure*}
\centering
\scalebox{0.75}{
\small
\begin{tabular}{@{}cc@{\mysp}c@{\mysp}c@{\mysp}c@{\mysp}c@{\mysp}c@{\mysp}c@{\mysp}c@{\mysp}c@{\mysp}c@{\mysp}c}
\makebox{Query$\downarrow$}  & & \rotatebox{90}{Rank 1} & \rotatebox{90}{Rank 2} & \rotatebox{90}{Rank 3} & \rotatebox{90}{Rank 4} & \rotatebox{90}{Rank 5} & \rotatebox{90}{Rank 6} & \rotatebox{90}{Rank 7} & \rotatebox{90}{Rank 8} & \rotatebox{90}{Rank 9} & \rotatebox{90}{Rank 10} \\
\multirow{2}{*}{\imCiftest{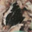}} & \rotatebox{90}{\footnotesize Baseline} &
\fcolorbox{orange}{orange}{\imCifbas{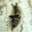}}&
\fcolorbox{green}{green}{\imCifbas{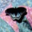}}&
\fcolorbox{green}{green}{\imCifbas{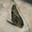}}&
\fcolorbox{green}{green}{\imCifbas{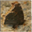}}&
\fcolorbox{orange}{orange}{\imCifbas{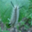}}&
\fcolorbox{orange}{orange}{\imCifbas{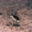}}&
\fcolorbox{orange}{orange}{\imCifbas{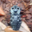}}&
\fcolorbox{green}{green}{\imCifbas{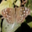}}&
\fcolorbox{orange}{orange}{\imCifbas{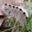}}&
\fcolorbox{orange}{orange}{\imCifbas{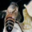}}\\
& \rotatebox{90}{\footnotesize \ours} &
\fcolorbox{green}{green}{\imCifour{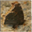}}&
\fcolorbox{green}{green}{\imCifour{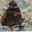}}&
\fcolorbox{green}{green}{\imCifour{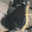}}&
\fcolorbox{green}{green}{\imCifour{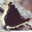}}&
\fcolorbox{green}{green}{\imCifour{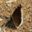}}&
\fcolorbox{green}{green}{\imCifour{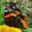}}&
\fcolorbox{orange}{orange}{\imCifour{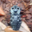}}&
\fcolorbox{green}{green}{\imCifour{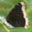}}&
\fcolorbox{green}{green}{\imCifour{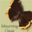}}&
\fcolorbox{green}{green}{\imCifour{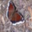}}\\[0.12cm]
\multirow{2}{*}{\imCiftest{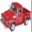}} & \rotatebox{90}{\footnotesize Baseline} &
\fcolorbox{green}{green}{\imCifbas{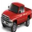}}&
\fcolorbox{green}{green}{\imCifbas{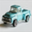}}&
\fcolorbox{green}{green}{\imCifbas{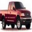}}&
\fcolorbox{green}{green}{\imCifbas{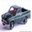}}&
\fcolorbox{green}{green}{\imCifbas{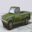}}&
\fcolorbox{green}{green}{\imCifbas{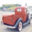}}&
\fcolorbox{green}{green}{\imCifbas{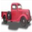}}&
\fcolorbox{green}{green}{\imCifbas{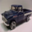}}&
\fcolorbox{green}{green}{\imCifbas{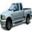}}&
\fcolorbox{green}{green}{\imCifbas{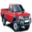}}\\
& \rotatebox{90}{\footnotesize \ours} &
\fcolorbox{green}{green}{\imCifour{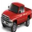}}&
\fcolorbox{green}{green}{\imCifour{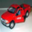}}&
\fcolorbox{green}{green}{\imCifour{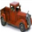}}&
\fcolorbox{green}{green}{\imCifour{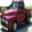}}&
\fcolorbox{green}{green}{\imCifour{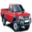}}&
\fcolorbox{green}{green}{\imCifour{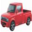}}&
\fcolorbox{orange}{orange}{\imCifour{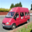}}&
\fcolorbox{green}{green}{\imCifour{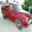}}&
\fcolorbox{green}{green}{\imCifour{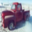}}&
\fcolorbox{green}{green}{\imCifour{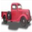}}\\[0.12cm]
\multirow{2}{*}{\imCiftest{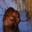}} & \rotatebox{90}{\footnotesize Baseline} &
\fcolorbox{green}{green}{\imCifbas{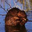}}&
\fcolorbox{orange}{orange}{\imCifbas{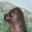}}&
\fcolorbox{green}{green}{\imCifbas{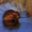}}&
\fcolorbox{orange}{orange}{\imCifbas{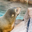}}&
\fcolorbox{green}{green}{\imCifbas{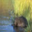}}&
\fcolorbox{green}{green}{\imCifbas{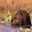}}&
\fcolorbox{green}{green}{\imCifbas{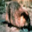}}&
\fcolorbox{orange}{orange}{\imCifbas{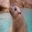}}&
\fcolorbox{green}{green}{\imCifbas{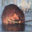}}&
\fcolorbox{green}{green}{\imCifbas{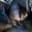}}\\
& \rotatebox{90}{\footnotesize \ours} &
\fcolorbox{green}{green}{\imCifour{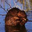}}&
\fcolorbox{green}{green}{\imCifour{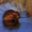}}&
\fcolorbox{green}{green}{\imCifour{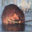}}&
\fcolorbox{green}{green}{\imCifour{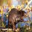}}&
\fcolorbox{green}{green}{\imCifour{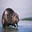}}&
\fcolorbox{green}{green}{\imCifour{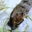}}&
\fcolorbox{green}{green}{\imCifour{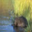}}&
\fcolorbox{orange}{orange}{\imCifour{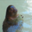}}&
\fcolorbox{green}{green}{\imCifour{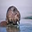}}&
\fcolorbox{orange}{orange}{\imCifour{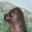}}
\end{tabular}}
\caption{CIFAR-100: For given test images (\textit{top}), we present the ranked list of train images that are most similar with embeddings obtained with a baseline method (top) and our method (bottom) train with coarse labels.
Images in \textcolor{green}{green} indicate that the image belongs to the correct fine class; \textcolor{orange}{orange} indicates the correct coarse class  but incorrect fine class. 
In this example, all results are correct w.r.t. coarse granularity. 
\label{fig:sampleranking}}
\end{figure*}

\FloatBarrier

\fboxrule=3pt
\fboxsep=0.1mm

\def \mysp {\hspace{2pt}}

\begin{figure*}[t]
    \centering
       \scalebox{0.8}{
    \begin{tabular}{|c|c|c@{\mysp}c@{\mysp}c@{\mysp}c@{\mysp}c@{\mysp}c@{\mysp}c@{\mysp}c@{\mysp}c@{\mysp}c@{\mysp}c|}
    \toprule
    \multirow{2}{2em}{Train levels} &  \multirow{2}{3em}{Method} &  \multirow{2}{4em}{Query Image} & \multicolumn{10}{c}{Neighbours in train}\\
    & & & Rank 1 & Rank 2 & Rank 3 & Rank 4 & Rank 5 & Rank 6 & Rank 7 & Rank 8 & Rank 9 & Rank 10\\
    \midrule
    \multirow{2}{2em}{\rotatebox{90}{\textcolor{kingdom}{kingdom}}} & \rotatebox{90}{Baseline} &
    \href{https://www.inaturalist.org/photos/10702054}{\iminatevol{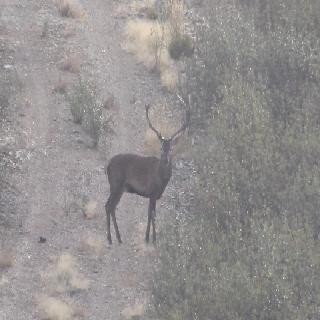}} &
    \fcolorbox{phylum}{phylum}{\iminatevol{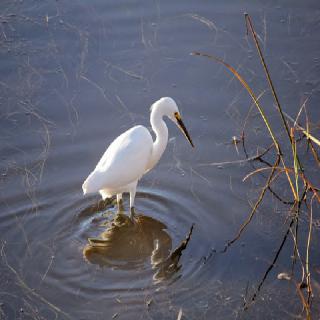}} &
    \fcolorbox{phylum}{phylum}{\iminatevol{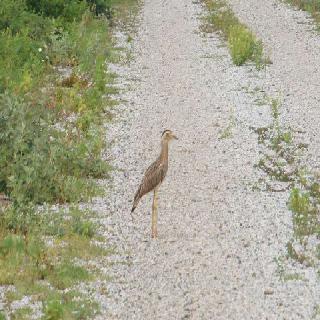}} &
    \fcolorbox{class}{class}{\iminatevol{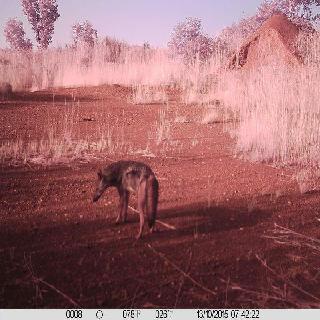}} &
    \fcolorbox{order}{order}{\iminatevol{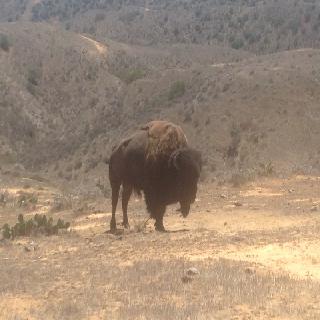}} &
    \fcolorbox{phylum}{phylum}{\iminatevol{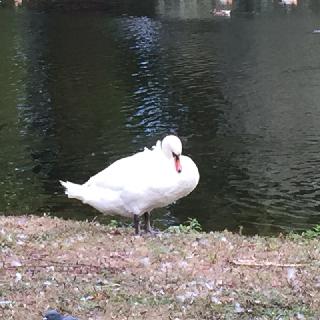}} &
    \fcolorbox{phylum}{phylum}{\iminatevol{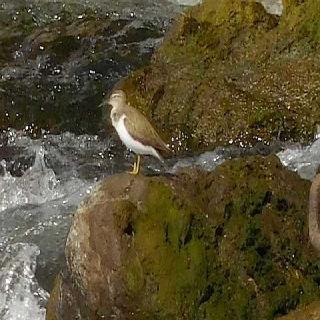}} &
    \fcolorbox{family}{family}{\iminatevol{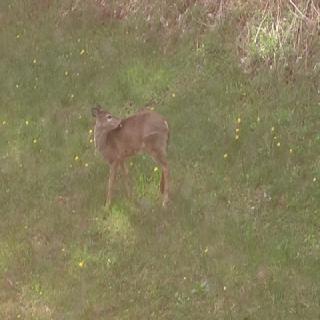}} &
    \fcolorbox{phylum}{phylum}{\iminatevol{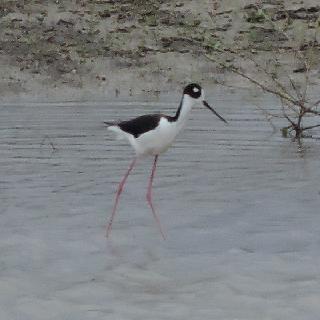}} &
    \fcolorbox{order}{order}{\iminatevol{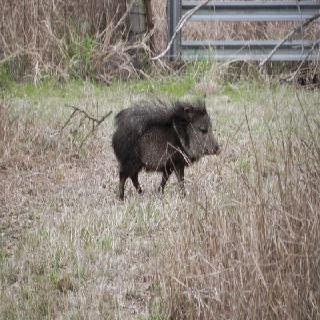}} &
    \fcolorbox{phylum}{phylum}{\iminatevol{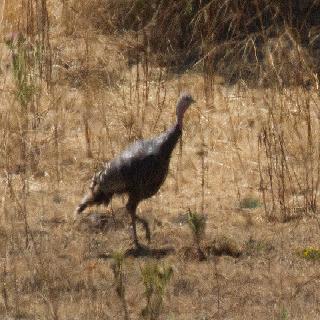}} \\
    & \rotatebox{90}{\ours} & &
    \fcolorbox{class}{class}{\iminatevol{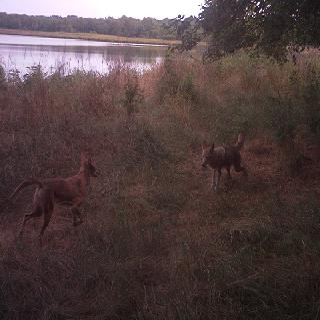}} &
    \fcolorbox{class}{class}{\iminatevol{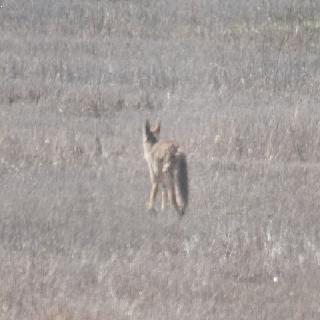}} &
    \fcolorbox{order}{order}{\iminatevol{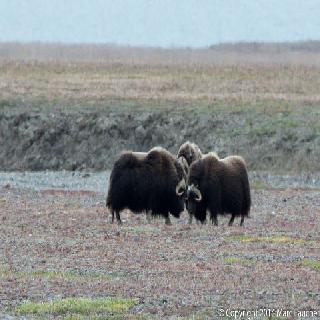}} &
    \fcolorbox{family}{family}{\iminatevol{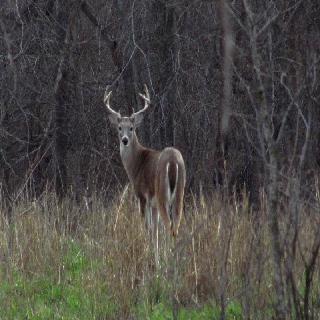}} &
    \fcolorbox{family}{family}{\iminatevol{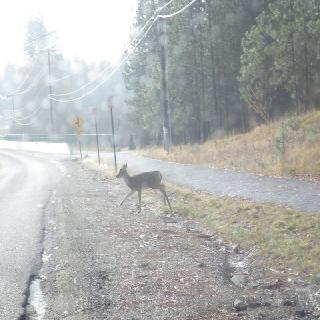}} &
    \fcolorbox{class}{class}{\iminatevol{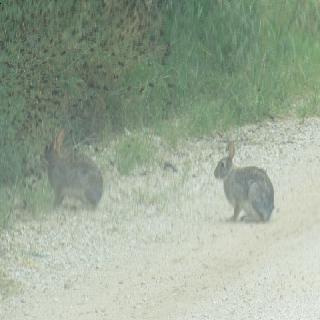}} &
    \href{https://www.inaturalist.org/observations/5052654}{\fcolorbox{species}{species}{\iminatevol{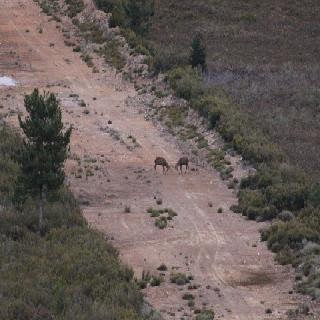}}} &
    \fcolorbox{class}{class}{\iminatevol{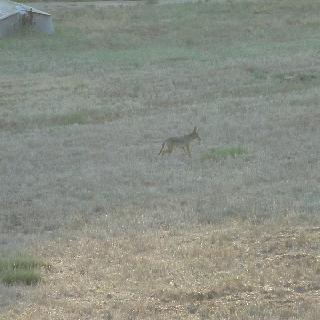}} &
    \fcolorbox{class}{class}{\iminatevol{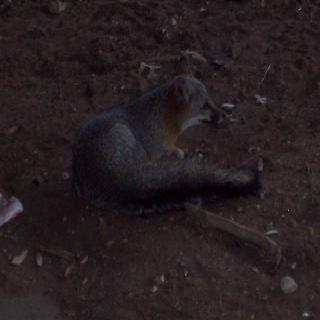}} &
    \fcolorbox{family}{family}{\iminatevol{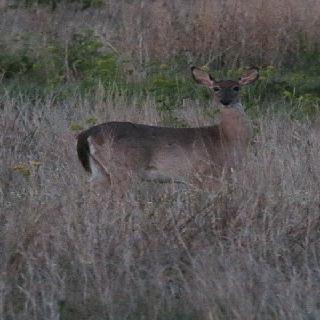}} \\
    \midrule
    \multirow{2}{2em}{\rotatebox{90}{\textcolor{phylum}{phylum}}} & \rotatebox{90}{Baseline} &
    \href{https://www.inaturalist.org/photos/10702054}{\iminatevol{baseline/15191/species/query_classe_4066.jpg}} &
    \fcolorbox{class}{class}{\iminatevol{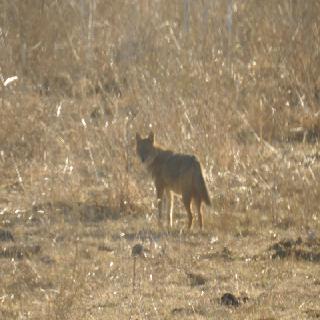}} &
    \fcolorbox{family}{family}{\iminatevol{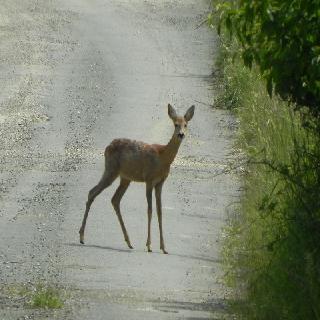}} &
    \fcolorbox{class}{class}{\iminatevol{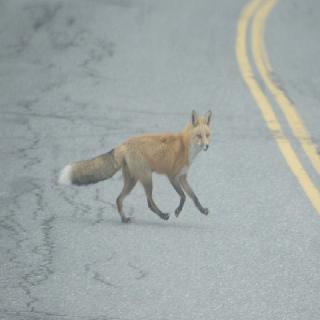}} &
    \fcolorbox{family}{family}{\iminatevol{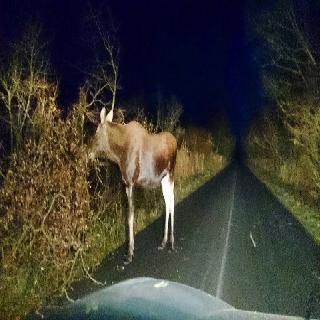}} &
    \fcolorbox{class}{class}{\iminatevol{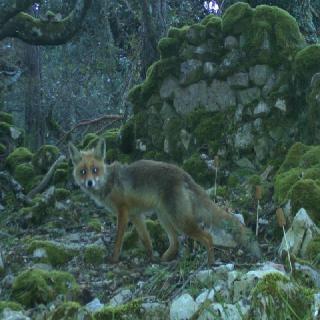}} &
    \fcolorbox{family}{family}{\iminatevol{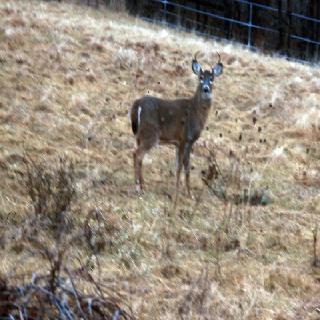}} &
    \fcolorbox{family}{family}{\iminatevol{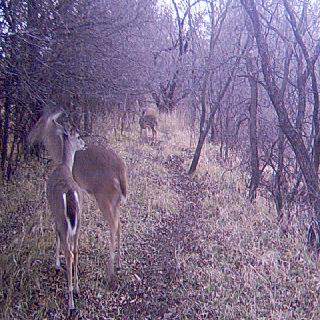}} &
    \fcolorbox{phylum}{phylum}{\iminatevol{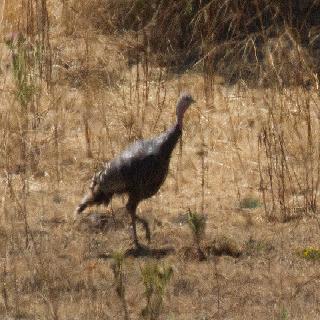}} &
    \fcolorbox{family}{family}{\iminatevol{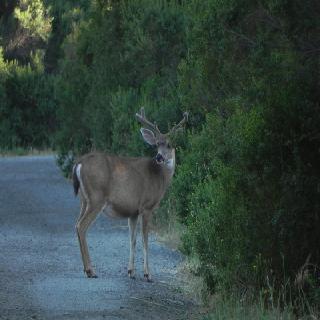}} &
    \fcolorbox{class}{class}{\iminatevol{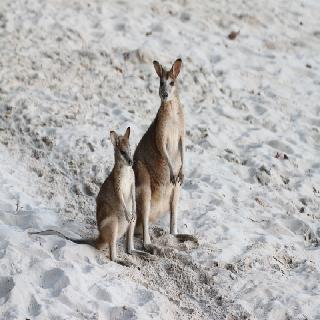}} \\
    & \rotatebox{90}{\ours} & &
    \fcolorbox{phylum}{phylum}{\iminatevol{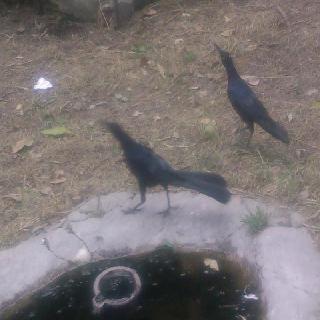}} &
    \fcolorbox{order}{order}{\iminatevol{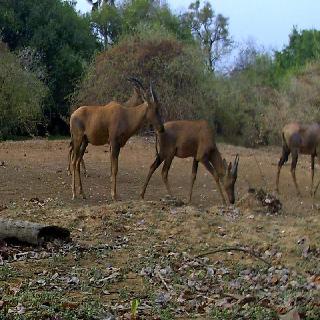}} &
    \fcolorbox{order}{order}{\iminatevol{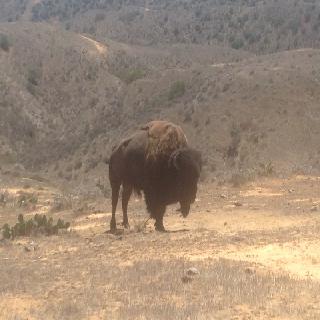}} &
    \fcolorbox{order}{order}{\iminatevol{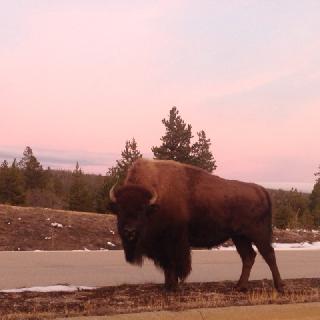}} &
    \fcolorbox{order}{order}{\iminatevol{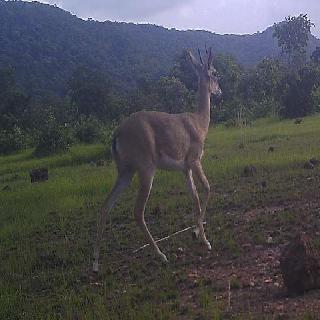}} &
    \fcolorbox{order}{order}{\iminatevol{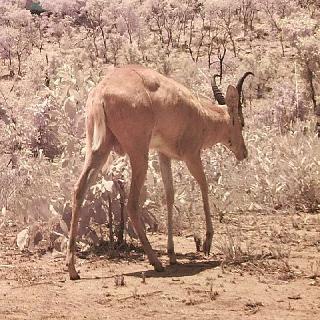}} &
    \fcolorbox{class}{class}{\iminatevol{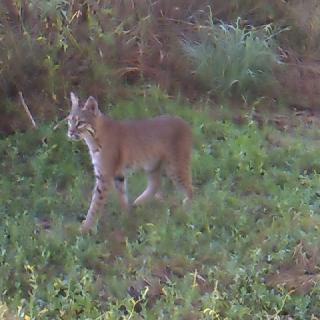}} &
    \fcolorbox{class}{class}{\iminatevol{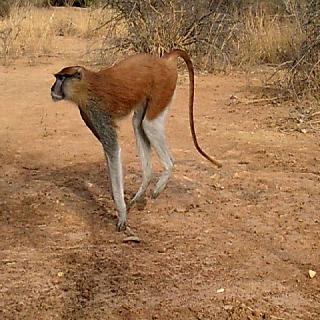}} &
    \fcolorbox{class}{class}{\iminatevol{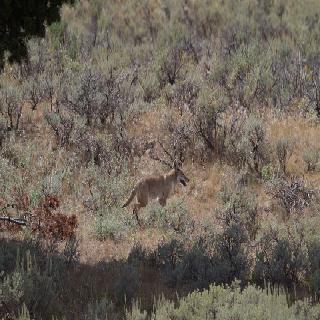}} &
    \fcolorbox{order}{order}{\iminatevol{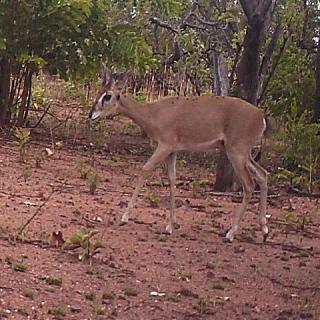}} \\
    \midrule
    \multirow{2}{2em}{\rotatebox{90}{\textcolor{class}{class}}} & \rotatebox{90}{Baseline} &
    \href{https://www.inaturalist.org/photos/10702054}{\iminatevol{baseline/15191/species/query_classe_4066.jpg}} &
    \fcolorbox{family}{family}{\iminatevol{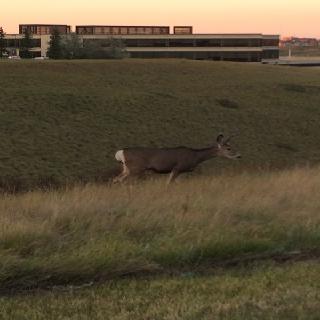}} &
    \fcolorbox{family}{family}{\iminatevol{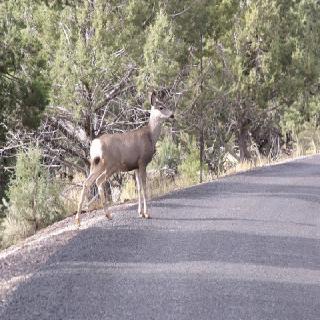}} &
    \fcolorbox{family}{family}{\iminatevol{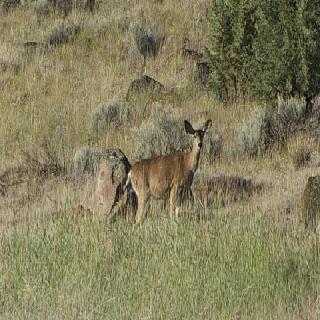}} &
    \fcolorbox{order}{order}{\iminatevol{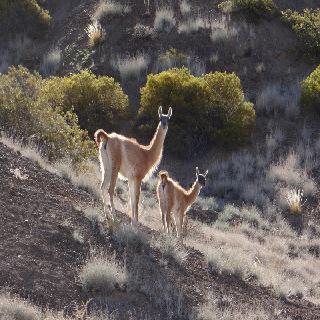}} &
    \fcolorbox{family}{family}{\iminatevol{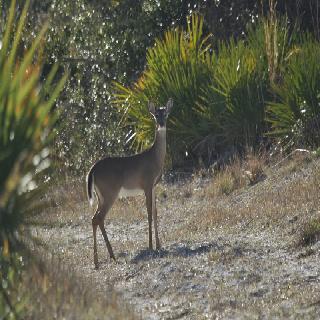}} &
    \fcolorbox{order}{order}{\iminatevol{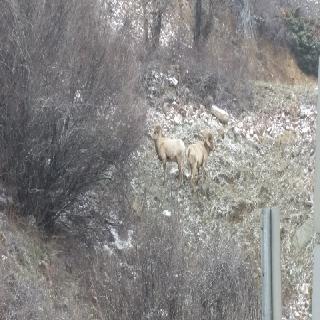}} &
    \fcolorbox{class}{class}{\iminatevol{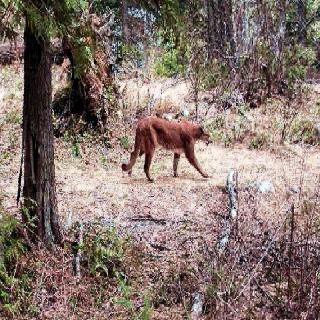}} &
    \fcolorbox{order}{order}{\iminatevol{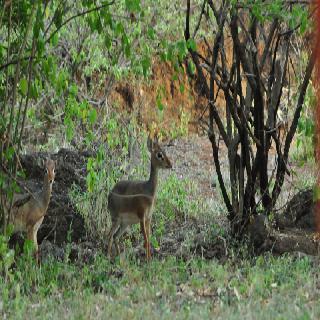}} &
    \fcolorbox{order}{order}{\iminatevol{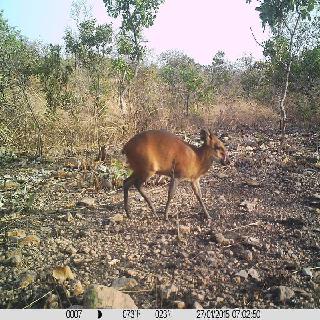}} &
    \fcolorbox{family}{family}{\iminatevol{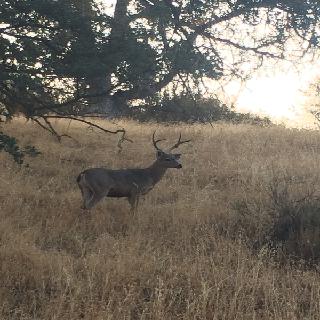}} \\
    & \rotatebox{90}{\ours} & &
    \fcolorbox{class}{class}{\iminatevol{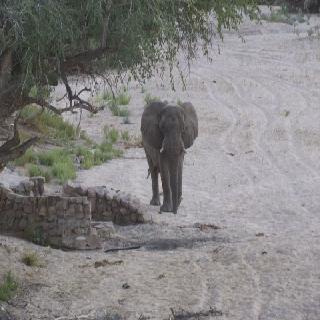}} &
    \fcolorbox{family}{family}{\iminatevol{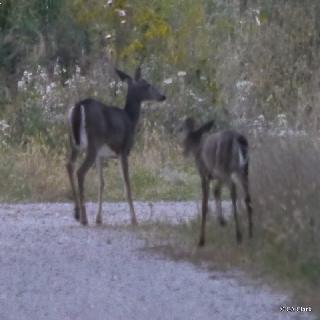}} &
    \fcolorbox{class}{class}{\iminatevol{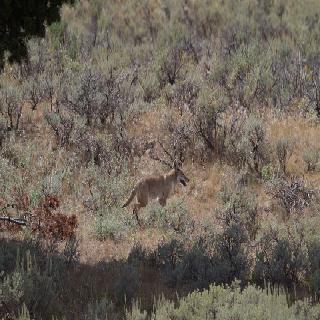}} &
    \fcolorbox{family}{family}{\iminatevol{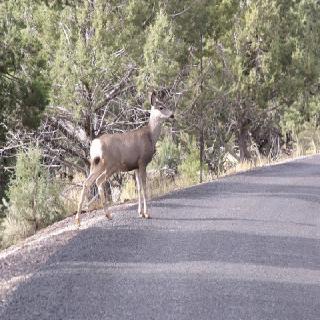}} &
    \fcolorbox{order}{order}{\iminatevol{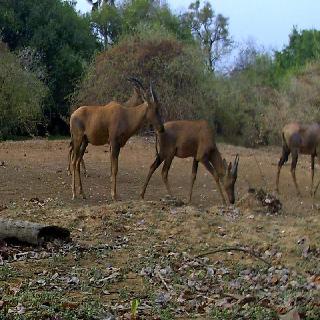}} &
    \fcolorbox{family}{family}{\iminatevol{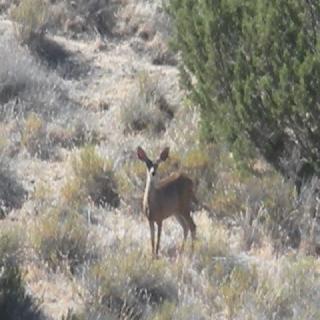}} &
    \fcolorbox{family}{family}{\iminatevol{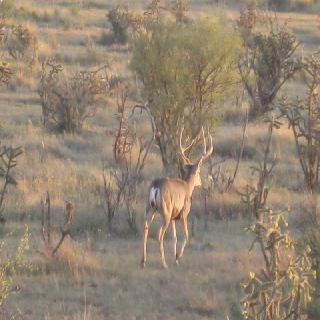}} &
    \fcolorbox{family}{family}{\iminatevol{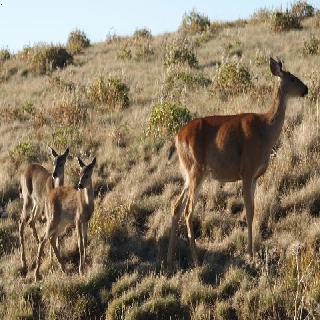}} &
    \fcolorbox{family}{family}{\iminatevol{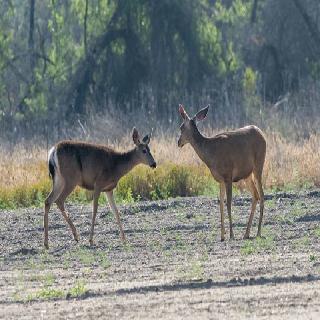}} &
    \fcolorbox{order}{order}{\iminatevol{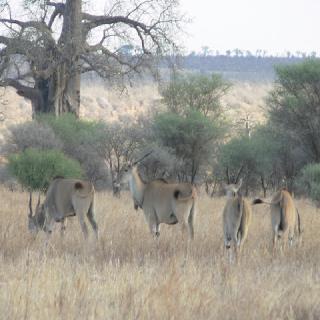}} \\
    \midrule
    \multirow{2}{2em}{\rotatebox{90}{\textcolor{order}{order}}} & \rotatebox{90}{Baseline} &
    \href{https://www.inaturalist.org/photos/10702054}{\iminatevol{baseline/15191/species/query_classe_4066.jpg}} &
    \fcolorbox{family}{family}{\iminatevol{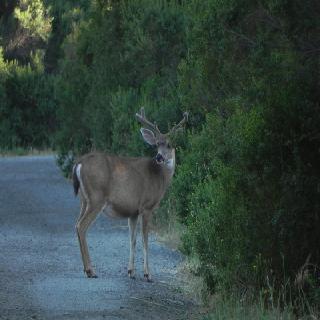}} &
    \fcolorbox{family}{family}{\iminatevol{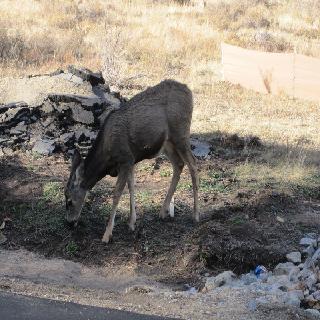}} &
    \fcolorbox{family}{family}{\iminatevol{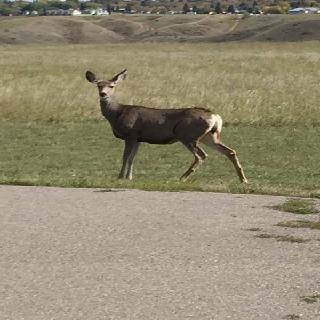}} &
    \fcolorbox{order}{order}{\iminatevol{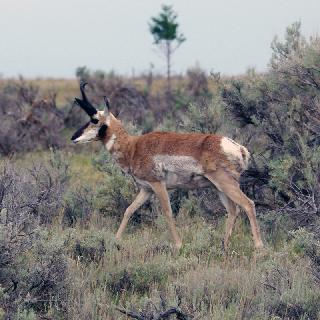}} &
    \fcolorbox{family}{family}{\iminatevol{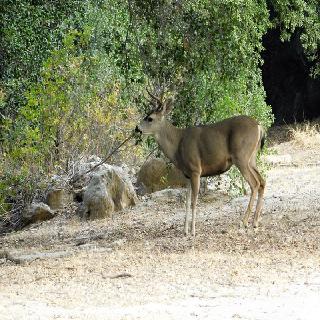}} &
    \fcolorbox{family}{family}{\iminatevol{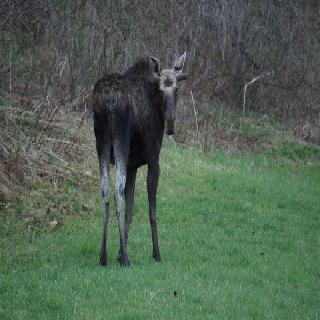}} &
    \fcolorbox{family}{family}{\iminatevol{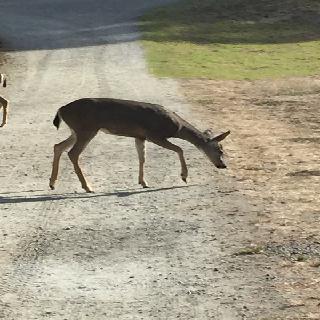}} &
    \fcolorbox{family}{family}{\iminatevol{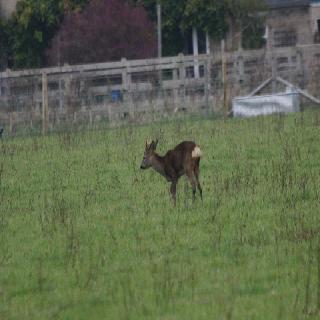}} &
    \fcolorbox{family}{family}{\iminatevol{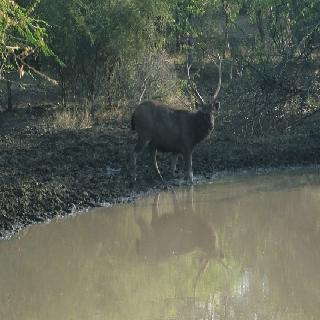}} &
    \fcolorbox{order}{order}{\iminatevol{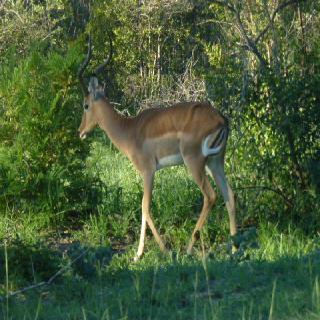}} \\
    & \rotatebox{90}{\ours} & &
    \fcolorbox{family}{family}{\iminatevol{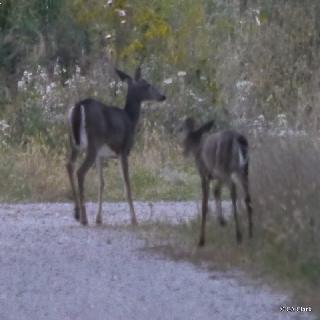}} &
    \fcolorbox{family}{family}{\iminatevol{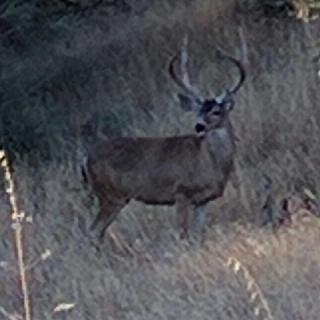}} &
    \fcolorbox{family}{family}{\iminatevol{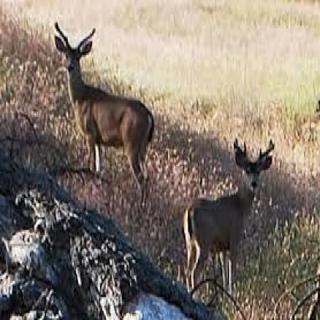}} &
    \fcolorbox{family}{family}{\iminatevol{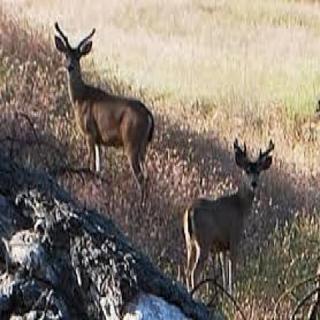}} &
    \fcolorbox{family}{family}{\iminatevol{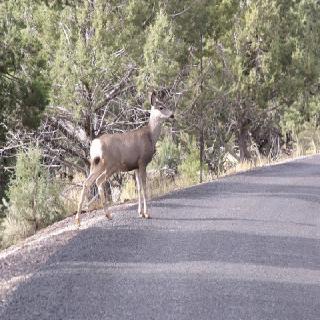}} &
    \fcolorbox{family}{family}{\iminatevol{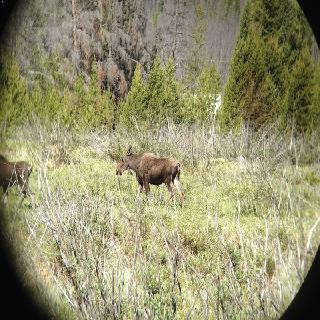}} &
    \fcolorbox{family}{family}{\iminatevol{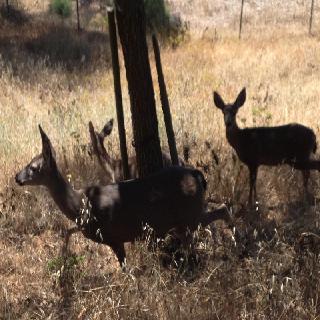}} &
    \fcolorbox{order}{order}{\iminatevol{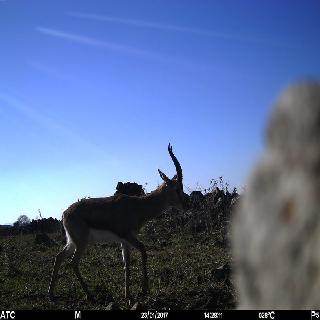}} &
    \fcolorbox{family}{family}{\iminatevol{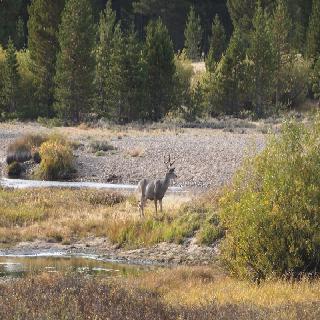}} &
    \fcolorbox{family}{family}{\iminatevol{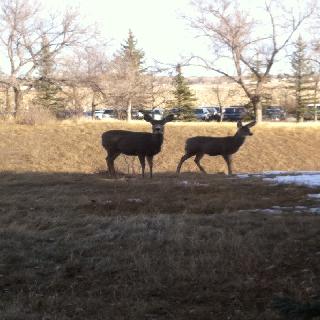}} \\
    \midrule
    \multirow{2}{2em}{\rotatebox{90}{\textcolor{family}{family}}} & \rotatebox{90}{Baseline} &
    \href{https://www.inaturalist.org/photos/10702054}{\iminatevol{baseline/15191/species/query_classe_4066.jpg}} &
    \fcolorbox{family}{family}{\iminatevol{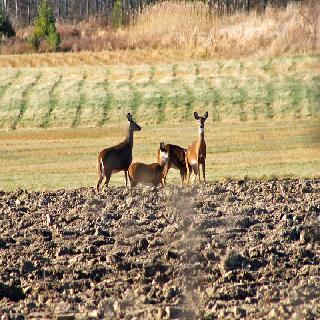}} &
    \fcolorbox{family}{family}{\iminatevol{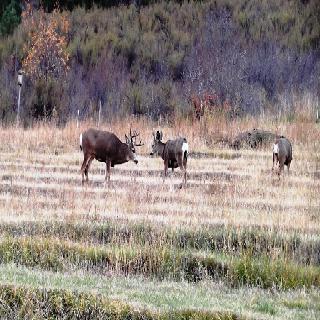}} &
    \fcolorbox{family}{family}{\iminatevol{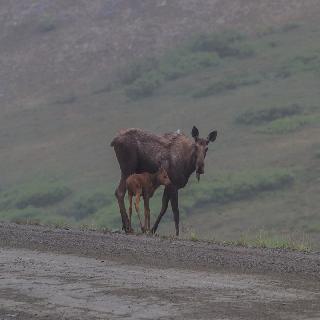}} &
    \fcolorbox{family}{family}{\iminatevol{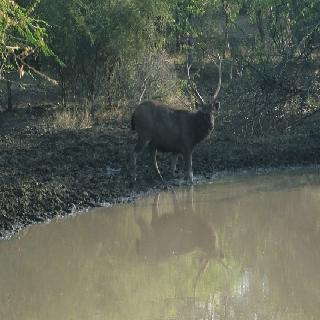}} &
    \fcolorbox{family}{family}{\iminatevol{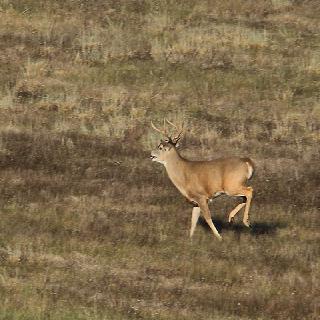}} &
    \fcolorbox{family}{family}{\iminatevol{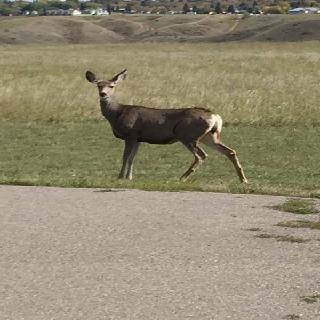}} &
    \fcolorbox{family}{family}{\iminatevol{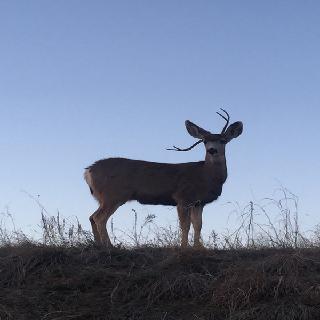}} &
    \fcolorbox{family}{family}{\iminatevol{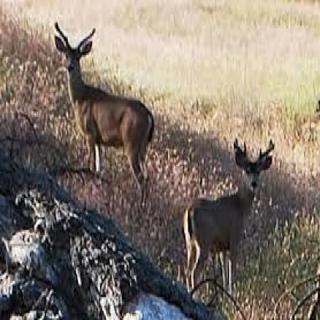}} &
    \fcolorbox{family}{family}{\iminatevol{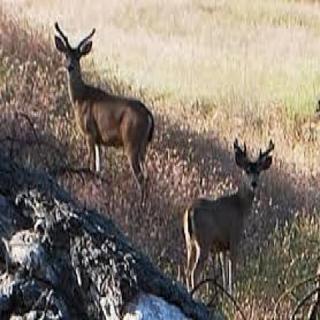}} &
    \fcolorbox{family}{family}{\iminatevol{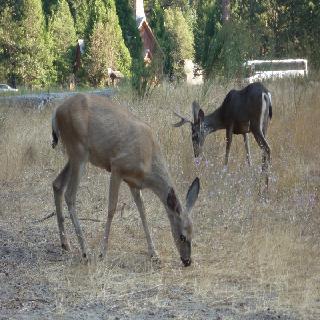}} \\
    & \rotatebox{90}{\ours} & &
    \fcolorbox{family}{family}{\iminatevol{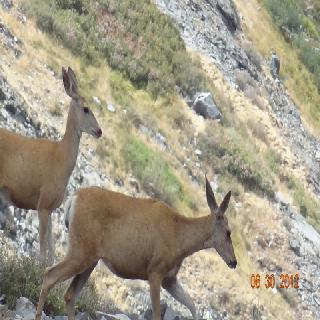}} &
    \fcolorbox{family}{family}{\iminatevol{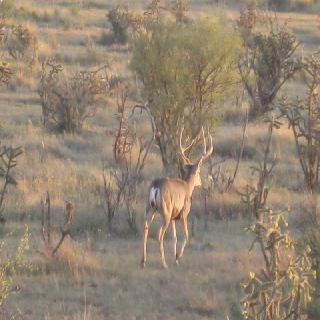}} &
    \fcolorbox{family}{family}{\iminatevol{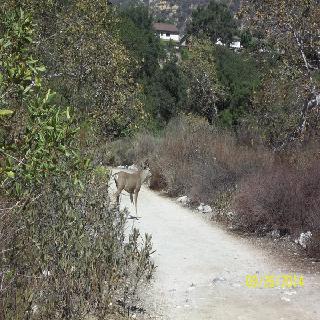}} &
    \fcolorbox{family}{family}{\iminatevol{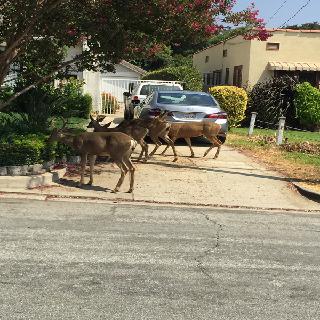}} &
    \fcolorbox{family}{family}{\iminatevol{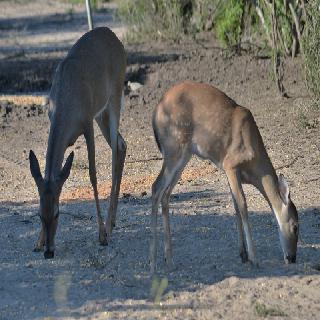}} &
    \fcolorbox{family}{family}{\iminatevol{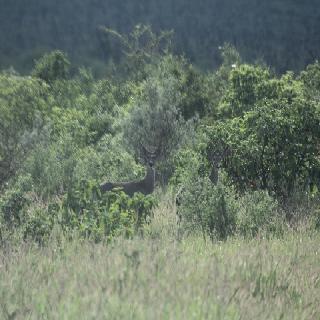}} &
    \fcolorbox{family}{family}{\iminatevol{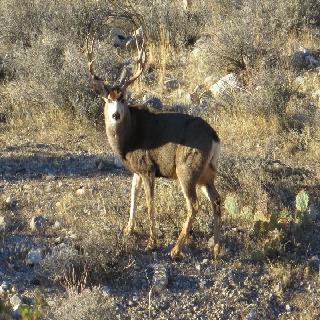}} &
    \fcolorbox{family}{family}{\iminatevol{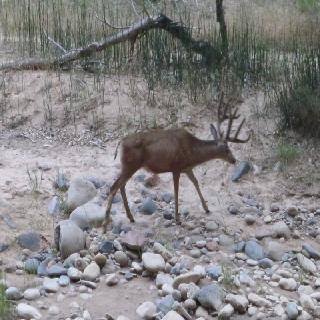}} &
    \fcolorbox{family}{family}{\iminatevol{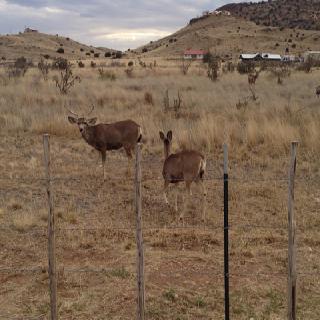}} &
    \fcolorbox{family}{family}{\iminatevol{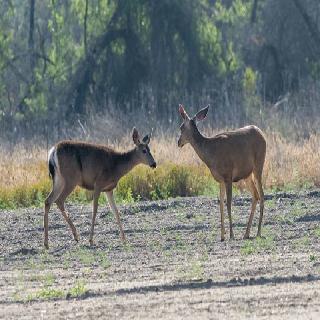}} \\
    \midrule
    \multirow{2}{2em}{\rotatebox{90}{\textcolor{genus}{genus}}} & \rotatebox{90}{Baseline} &
    \href{https://www.inaturalist.org/photos/10702054}{\iminatevol{baseline/15191/species/query_classe_4066.jpg}} &
    \href{https://www.inaturalist.org/photos/5686114}{\fcolorbox{family}{family}{\iminatevol{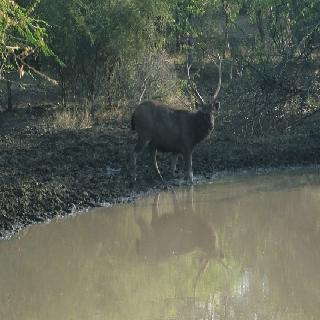}}} &
    \fcolorbox{family}{family}{\iminatevol{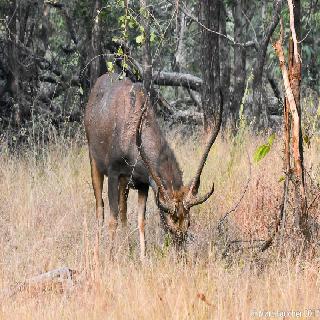}} &
    \fcolorbox{family}{family}{\iminatevol{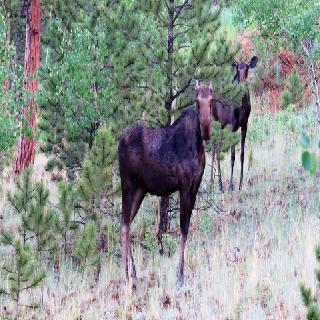}} &
    \fcolorbox{family}{family}{\iminatevol{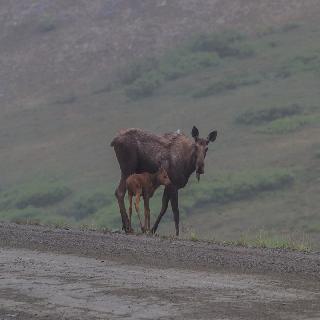}} &
    \href{https://www.inaturalist.org/photos/9628100}{\fcolorbox{genus}{genus}{\iminatevol{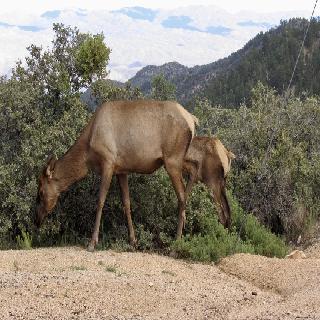}}} &
    \fcolorbox{family}{family}{\iminatevol{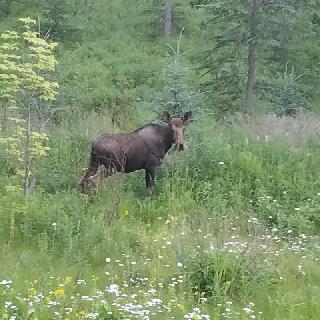}} &
    \fcolorbox{order}{order}{\iminatevol{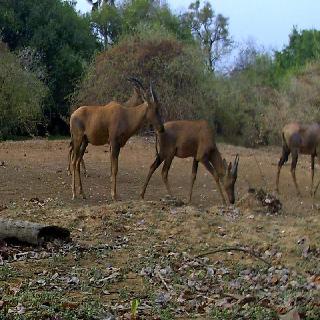}} &
    \fcolorbox{order}{order}{\iminatevol{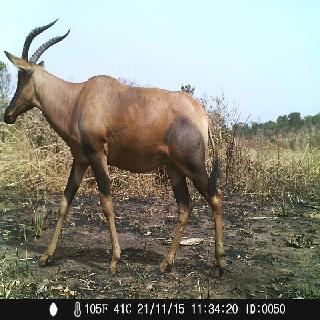}} &
    \fcolorbox{order}{order}{\iminatevol{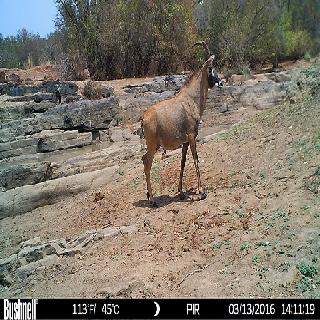}} &
    \href{https://www.inaturalist.org/photos/10900364}{\fcolorbox{species}{species}{\iminatevol{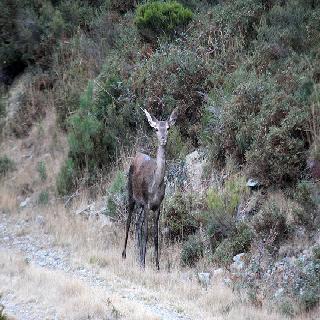}}} \\
    & \rotatebox{90}{\ours} & &
    \fcolorbox{species}{species}{\href{https://www.inaturalist.org/photos/6131201}{\iminatevol{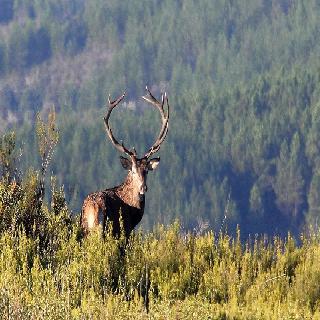}}} &
    \href{https://www.inaturalist.org/photos/10900364}{\fcolorbox{species}{species}{\iminatevol{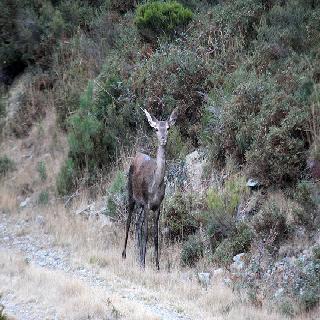}}} &
    \href{https://www.inaturalist.org/observations/5046475}{\fcolorbox{species}{species}{\iminatevol{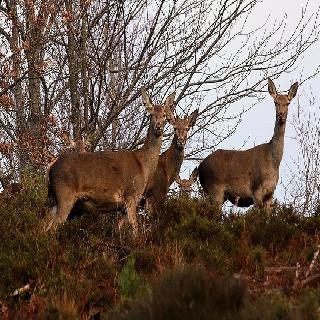}}} &
    \href{https://www.inaturalist.org/photos/10900448}{\fcolorbox{species}{species}{\iminatevol{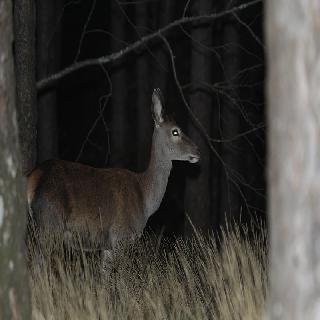}}} &
    \href{https://www.inaturalist.org/observations/5052681}{\fcolorbox{species}{species}{\iminatevol{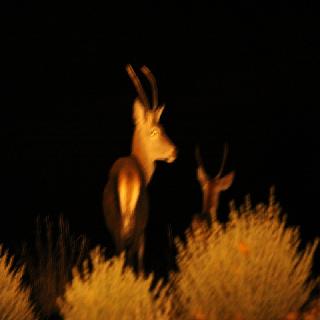}}} &
    \href{https://www.inaturalist.org/photos/5825715}{\fcolorbox{genus}{genus}{\iminatevol{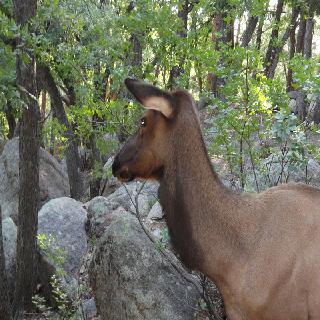}}} &
    \href{https://www.inaturalist.org/observations/5035204}{\fcolorbox{species}{species}{\iminatevol{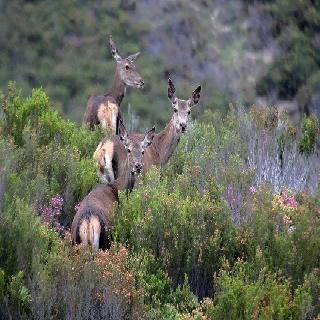}}} &
    \href{https://www.inaturalist.org/photos/6131197}{\fcolorbox{species}{species}{\iminatevol{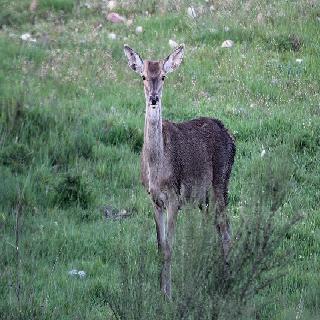}}} &
    \href{https://www.inaturalist.org/photos/5699232}{\fcolorbox{genus}{genus}{\iminatevol{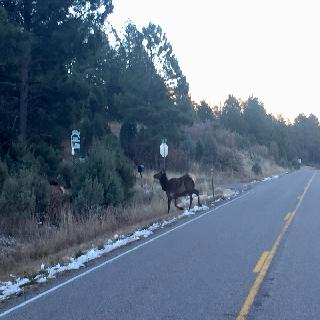}}} &
    \href{https://www.inaturalist.org/observations/5052258}{\fcolorbox{species}{species}{\iminatevol{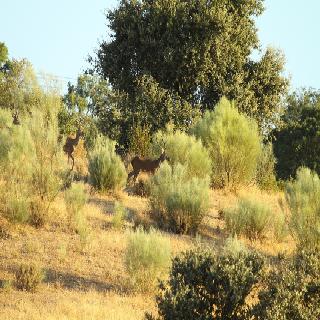}}} \\
    \midrule
    \multirow{2}{2em}{\rotatebox{90}{\textcolor{species}{species}}} & \rotatebox{90}{Baseline} &
    \href{https://www.inaturalist.org/photos/10702054}{\iminatevol{baseline/15191/species/query_classe_4066.jpg}} &
    \fcolorbox{family}{family}{\iminatevol{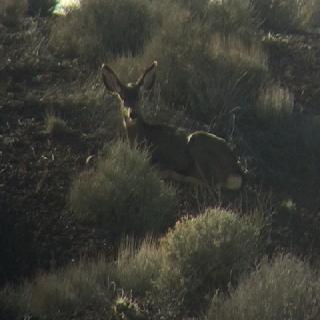}} &
    \fcolorbox{order}{order}{\iminatevol{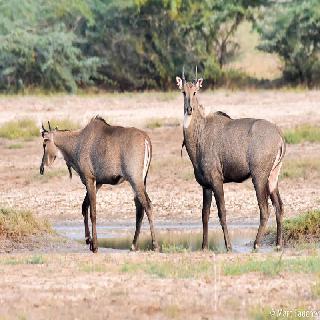}} &
    \fcolorbox{order}{order}{\iminatevol{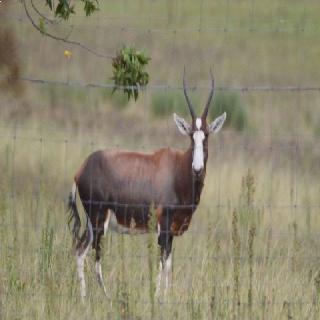}} &
    \fcolorbox{family}{family}{\iminatevol{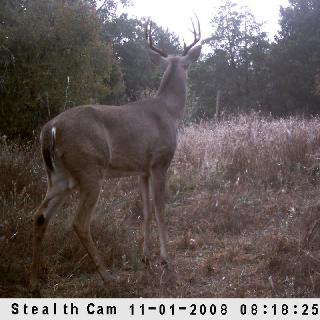}} &
    \fcolorbox{order}{order}{\iminatevol{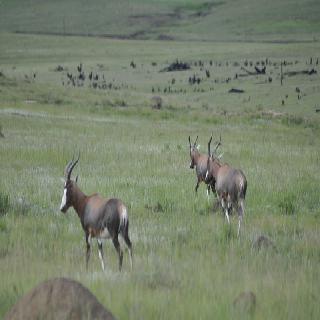}} &
    \fcolorbox{family}{family}{\iminatevol{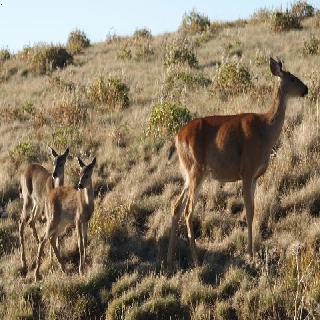}} &
    \fcolorbox{order}{order}{\iminatevol{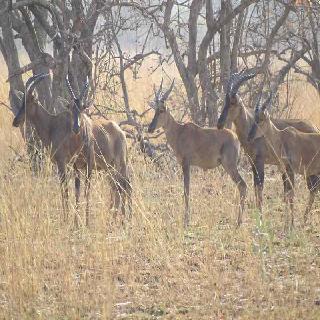}} &
    \fcolorbox{family}{family}{\iminatevol{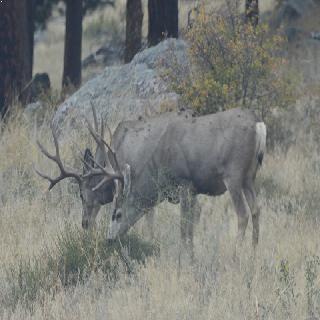}} &
    \href{https://www.inaturalist.org/photos/10900364}{\fcolorbox{species}{species}{\iminatevol{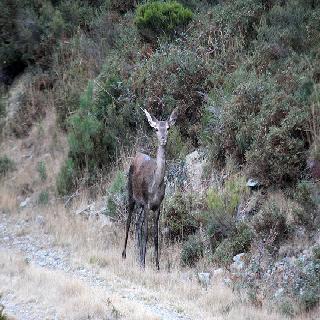}}} &
    \fcolorbox{family}{family}{\iminatevol{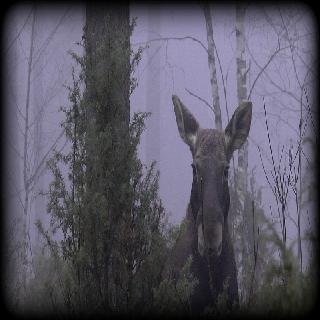}} \\
    & \rotatebox{90}{\ours} & &
    \fcolorbox{species}{species}{\href{https://www.inaturalist.org/photos/6131201}{\iminatevol{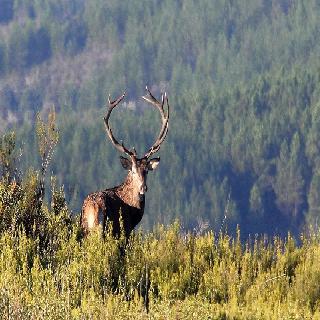}}} &
    \href{https://www.inaturalist.org/observations/5035204}{\fcolorbox{species}{species}{\iminatevol{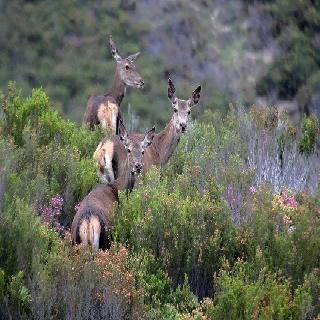}}} &
    \href{https://www.inaturalist.org/photos/10900364}{\fcolorbox{species}{species}{\iminatevol{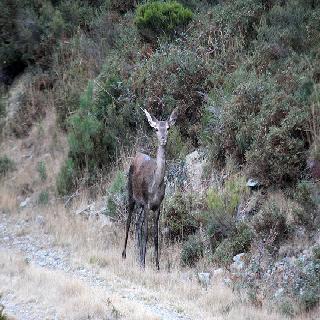}}} &
    \href{https://www.inaturalist.org/photos/6131197}{\fcolorbox{species}{species}{\iminatevol{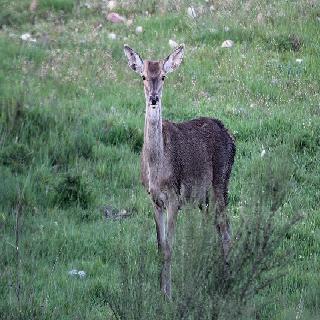}}} &
    \href{https://www.inaturalist.org/observations/5052681}{\fcolorbox{species}{species}{\iminatevol{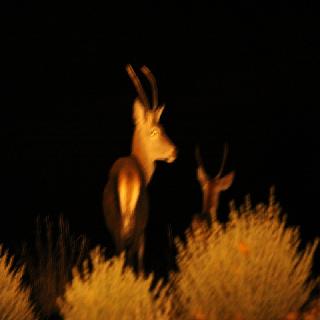}}} &
    \href{https://www.inaturalist.org/photos/10900448}{\fcolorbox{species}{species}{\iminatevol{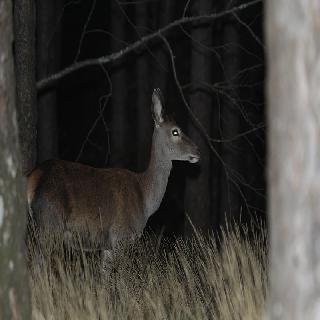}}} &
    \href{https://www.inaturalist.org/observations/5046475}{\fcolorbox{species}{species}{\iminatevol{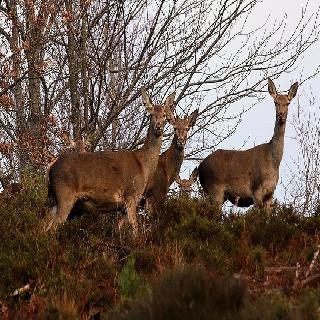}}} &
    \href{https://www.inaturalist.org/observations/5052258}{\fcolorbox{species}{species}{\iminatevol{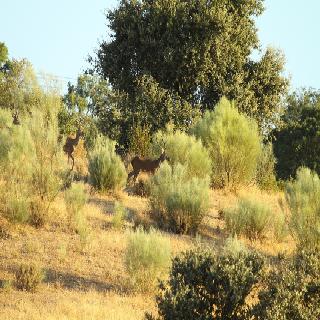}}} &
    \href{https://www.inaturalist.org/observations/5052654}{\fcolorbox{species}{species}{\iminatevol{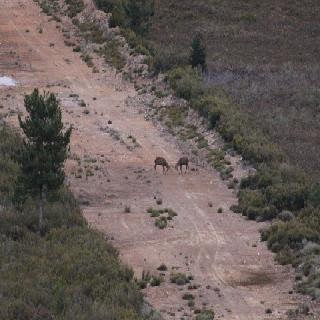}}} &
    \fcolorbox{family}{family}{\iminatevol{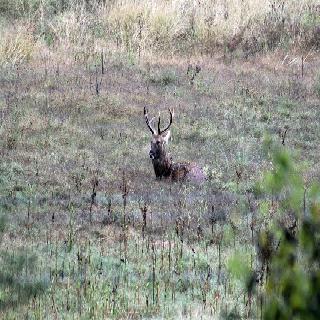}} \\

    \bottomrule
    \end{tabular}}
    \caption{We compare  \ours and Baseline for different training granularity. We rank the 10 closest images in the iNaturalist-2018 train set for a query image in the test set.
    The ranking is obtained with a cosine similarity on the features space of each of the two approaches. See Table~\ref{tab:copyright_appendix_1} for authors and image copyrights. \label{fig:inat_ranking_evolution_appendix_1}}
    
\end{figure*}

\fboxrule=3pt
\fboxsep=0.1mm

\def \mysp {\hspace{2pt}}

\begin{figure*}[t]
    \centering
    
\scalebox{0.8}{
\begin{tabular}{|c|c|c@{\mysp}c@{\mysp}c@{\mysp}c@{\mysp}c@{\mysp}c@{\mysp}c@{\mysp}c@{\mysp}c@{\mysp}c@{\mysp}c|}
\toprule
\multirow{2}{2em}{Train levels} &  \multirow{2}{3em}{Method} &  \multirow{2}{4em}{Query Image} & \multicolumn{10}{c}{Neighbours in train}\\
& & & Rank 1 & Rank 2 & Rank 3 & Rank 4 & Rank 5 & Rank 6 & Rank 7 & Rank 8 & Rank 9 & Rank 10\\
\midrule
\multirow{2}{2em}{\rotatebox{90}{\textcolor{kingdom}{kingdom}}} & \rotatebox{90}{Baseline} &
\href{https://www.inaturalist.org/photos/2609440}{\iminatevol{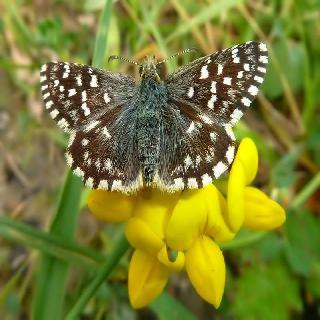}} &
\fcolorbox{genus}{genus}{\iminatevol{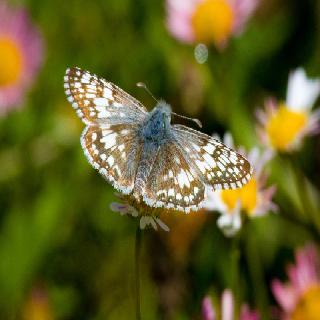}} &
\fcolorbox{order}{order}{\iminatevol{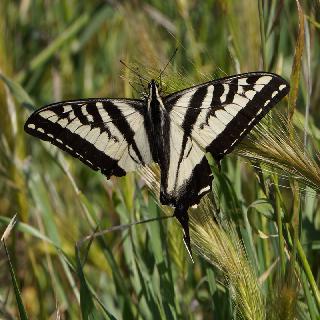}} &
\fcolorbox{order}{order}{\iminatevol{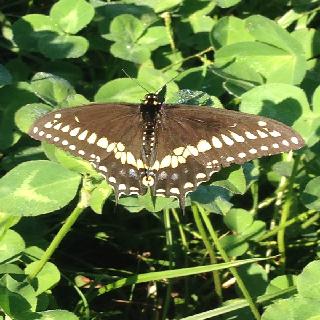}} &
\fcolorbox{family}{family}{\iminatevol{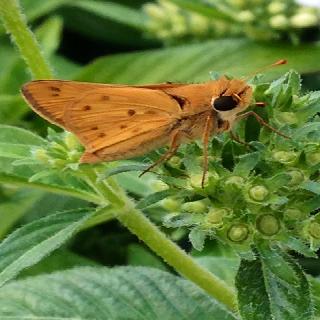}} &
\fcolorbox{order}{order}{\iminatevol{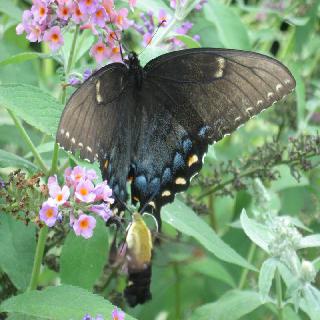}} &
\fcolorbox{order}{order}{\iminatevol{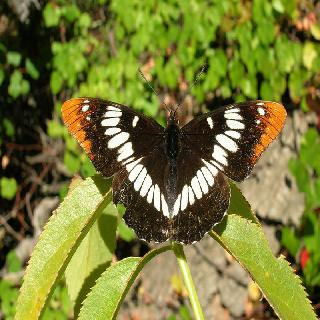}} &
\fcolorbox{order}{order}{\iminatevol{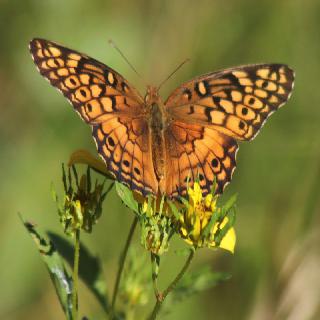}} &
\fcolorbox{order}{order}{\iminatevol{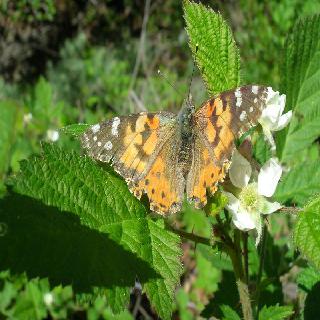}} &
\fcolorbox{order}{order}{\iminatevol{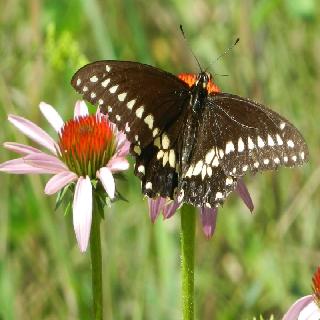}} &
\fcolorbox{order}{order}{\iminatevol{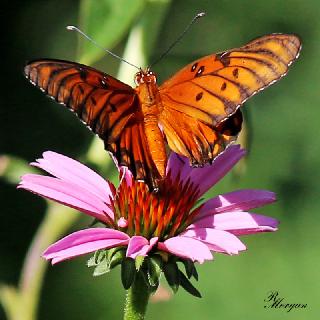}} \\
& \rotatebox{90}{\ours} & &
\href{https://www.inaturalist.org/observations/2939424}{\fcolorbox{species}{species}{\iminatevol{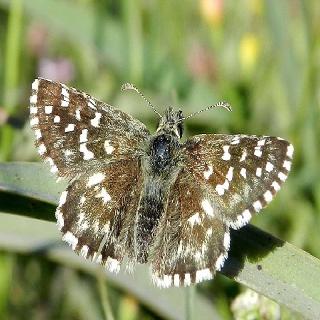}}}&
\href{https://www.inaturalist.org/observations/1953410}{\fcolorbox{genus}{genus}{\iminatevol{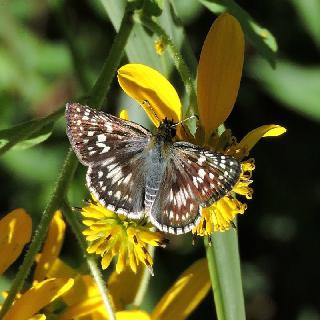}}} &
\href{https://www.inaturalist.org/observations/5939102}{\fcolorbox{species}{species}{\iminatevol{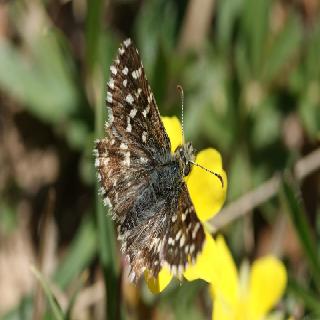}}} &
\href{https://www.inaturalist.org/photos/3434401}{\fcolorbox{genus}{genus}{\iminatevol{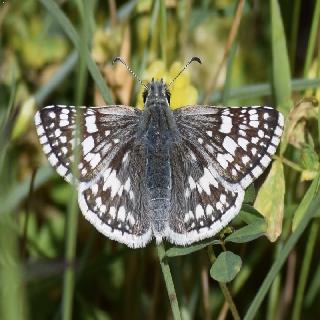}}} &
\href{https://www.inaturalist.org/photos/1843534}{\fcolorbox{genus}{genus}{\iminatevol{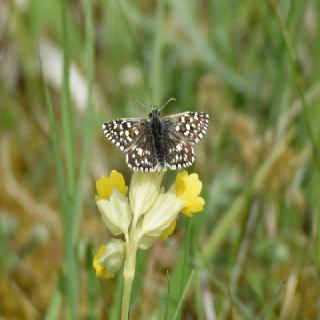}}} &
\href{https://www.inaturalist.org/photos/3615109}{\fcolorbox{genus}{genus}{\iminatevol{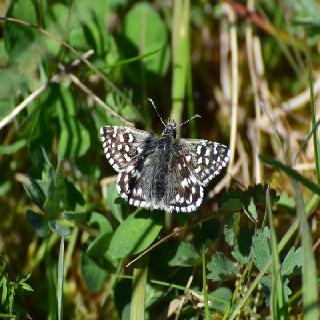}}} &
\fcolorbox{genus}{genus}{\iminatevol{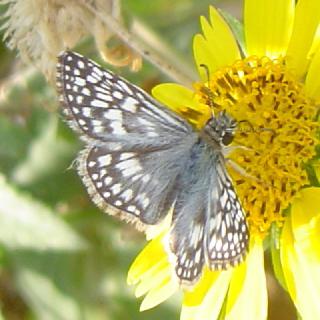}} &
\fcolorbox{species}{species}{\iminatevol{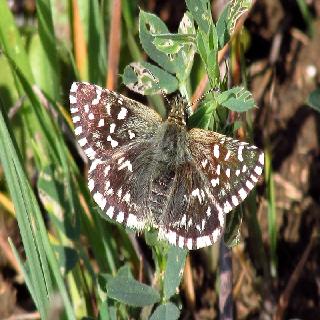}} &
\fcolorbox{genus}{genus}{\iminatevol{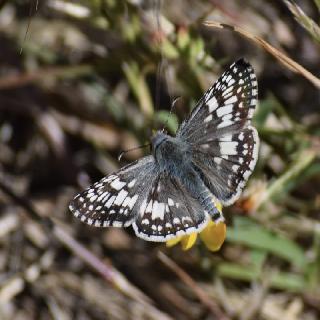}} &
\fcolorbox{species}{species}{\iminatevol{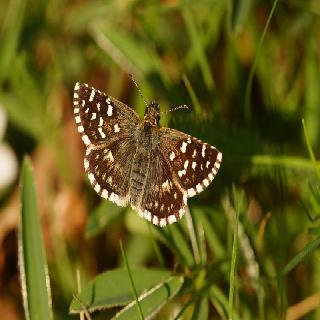}} \\
\midrule
\multirow{2}{2em}{\rotatebox{90}{\textcolor{phylum}{phylum}}} & \rotatebox{90}{Baseline} &
\href{https://www.inaturalist.org/photos/2609440}{\iminatevol{baseline/419/kingdom/query_classe_1092.jpg}} &
\fcolorbox{order}{order}{\iminatevol{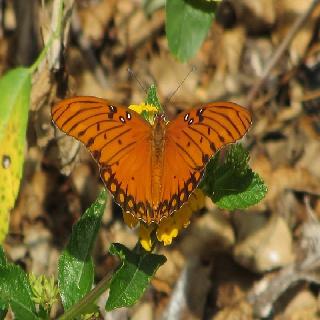}} &
\fcolorbox{order}{order}{\iminatevol{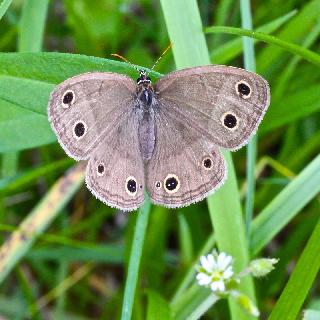}} &
\fcolorbox{order}{order}{\iminatevol{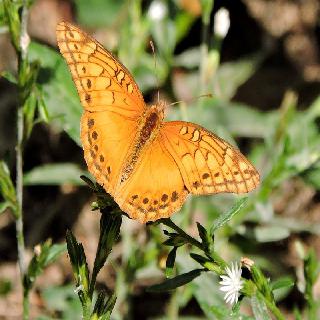}} &
\fcolorbox{order}{order}{\iminatevol{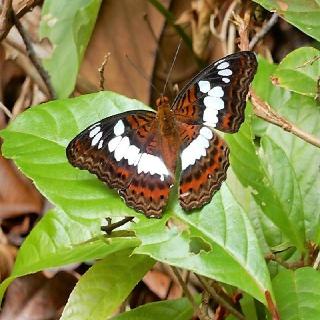}} &
\fcolorbox{order}{order}{\iminatevol{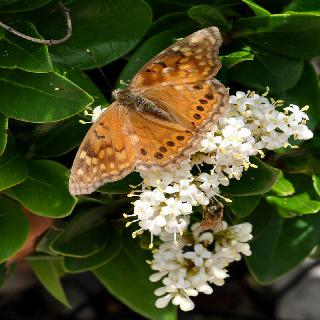}} &
\fcolorbox{order}{order}{\iminatevol{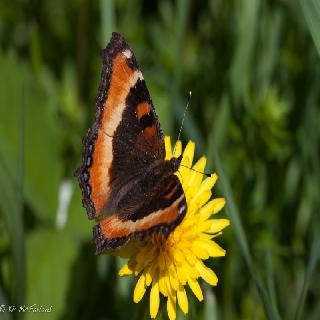}} &
\fcolorbox{order}{order}{\iminatevol{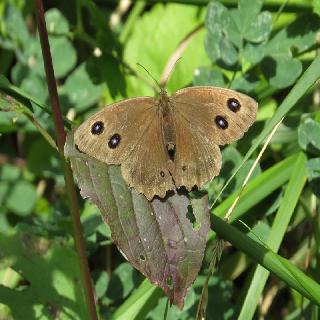}} &
\fcolorbox{family}{family}{\iminatevol{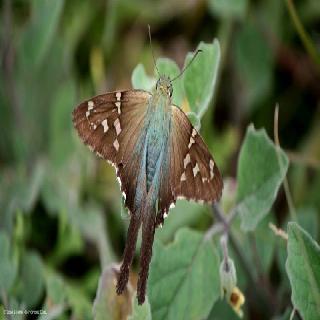}} &
\fcolorbox{order}{order}{\iminatevol{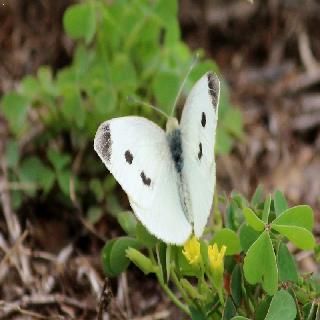}} &
\fcolorbox{order}{order}{\iminatevol{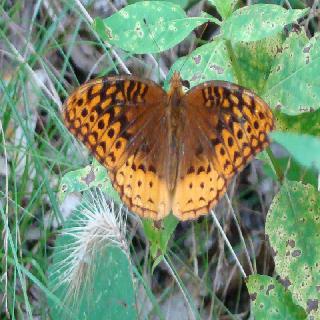}} \\
& \rotatebox{90}{\ours} & &
\fcolorbox{species}{species}{\iminatevol{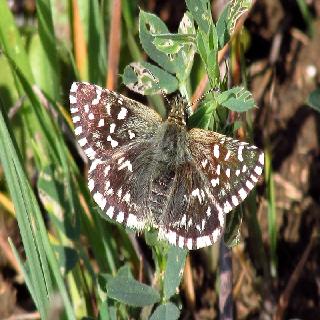}} &
\href{https://www.inaturalist.org/observations/2939424}{\fcolorbox{species}{species}{\iminatevol{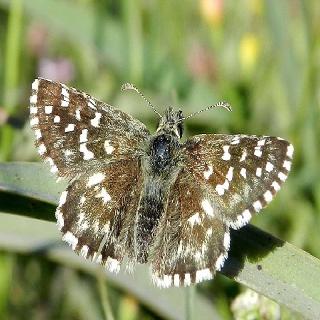}}} &
\href{https://www.inaturalist.org/observations/1953410}{\fcolorbox{genus}{genus}{\iminatevol{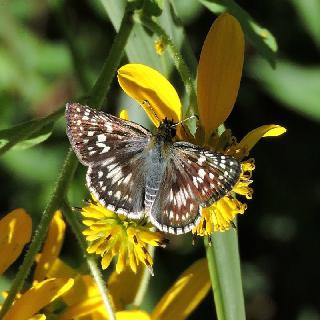}}} &
\fcolorbox{genus}{genus}{\iminatevol{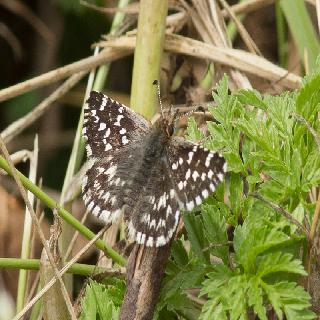}} &
\fcolorbox{species}{species}{\iminatevol{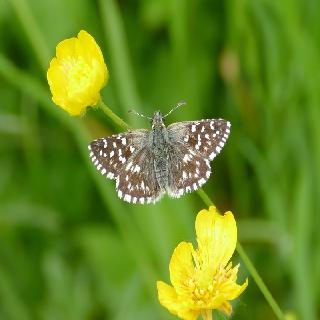}} &
\href{https://www.inaturalist.org/observations/5939102}{\fcolorbox{species}{species}{\iminatevol{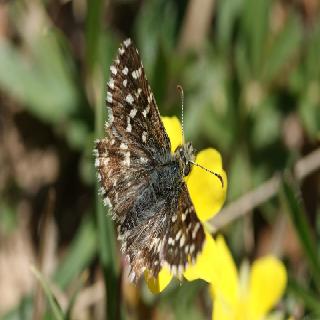}}} &
\fcolorbox{species}{species}{\iminatevol{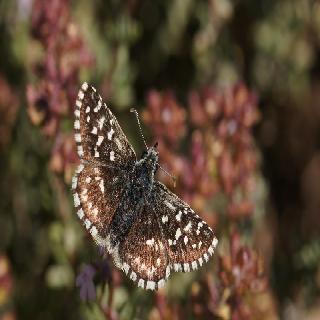}} &
\fcolorbox{genus}{genus}{\iminatevol{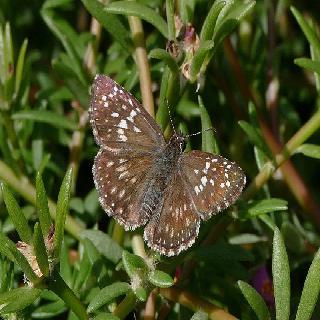}} &
\fcolorbox{genus}{genus}{\iminatevol{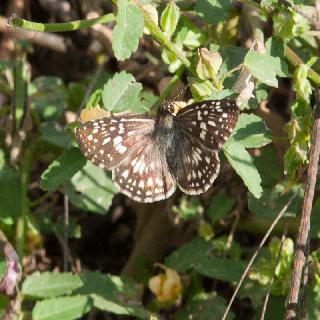}} &
\fcolorbox{species}{species}{\iminatevol{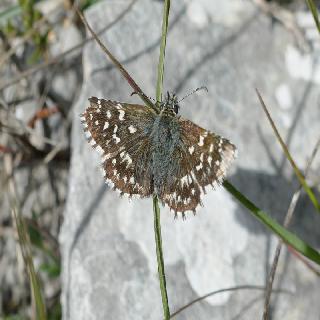}} \\
\midrule
\multirow{2}{2em}{\rotatebox{90}{\textcolor{class}{class}}} & \rotatebox{90}{Baseline} &
\href{https://www.inaturalist.org/photos/2609440}{\iminatevol{baseline/419/kingdom/query_classe_1092.jpg}}&
\href{https://www.inaturalist.org/photos/3434401}{\fcolorbox{genus}{genus}{\iminatevol{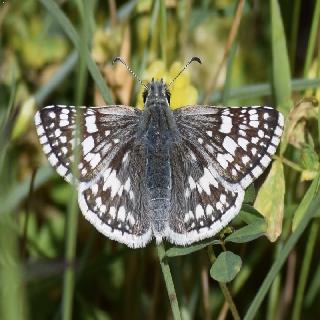}}} &
\fcolorbox{order}{order}{\iminatevol{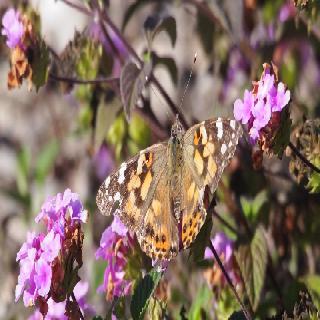}} &
\fcolorbox{genus}{genus}{\iminatevol{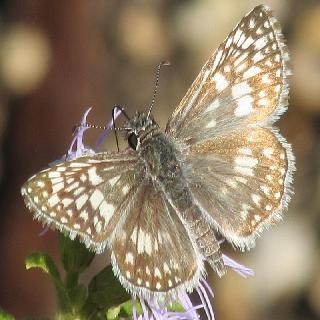}} &
\fcolorbox{genus}{genus}{\iminatevol{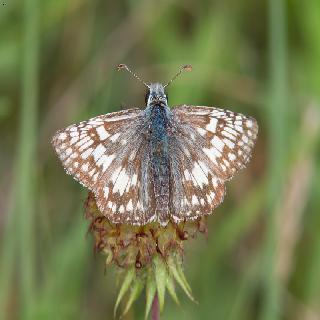}} &
\fcolorbox{order}{order}{\iminatevol{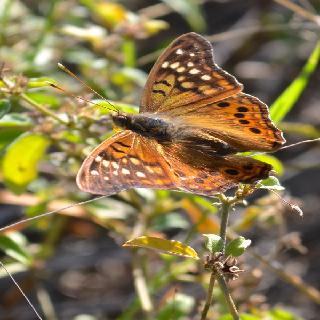}} &
\fcolorbox{genus}{genus}{\iminatevol{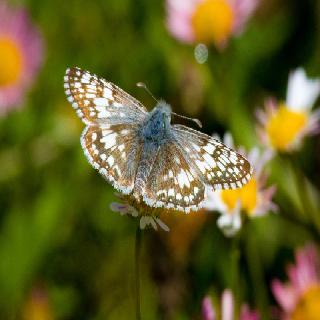}} &
\fcolorbox{order}{order}{\iminatevol{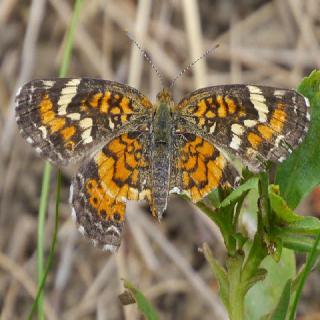}} &
\fcolorbox{species}{species}{\iminatevol{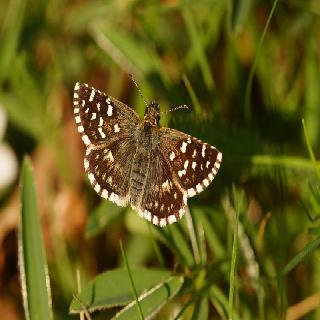}} &
\fcolorbox{order}{order}{\iminatevol{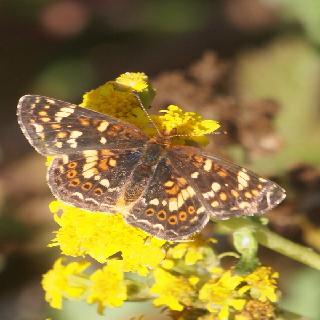}} &
\fcolorbox{order}{order}{\iminatevol{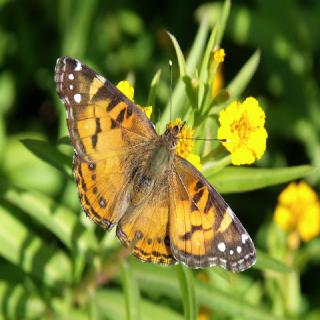}} \\
& \rotatebox{90}{\ours} & &
\fcolorbox{species}{species}{\iminatevol{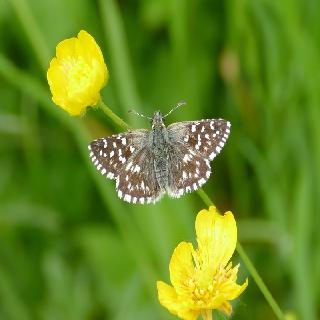}} &
\href{https://www.inaturalist.org/observations/2939424}{\fcolorbox{species}{species}{\iminatevol{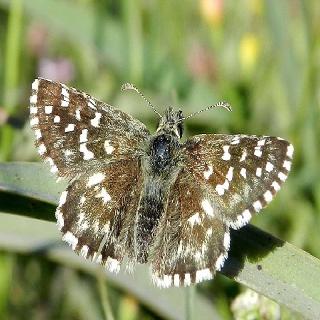}}}&
\href{https://www.inaturalist.org/photos/3434401}{\fcolorbox{genus}{genus}{\iminatevol{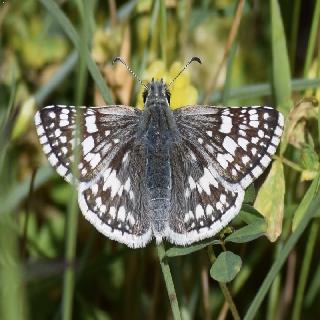}}} &
\fcolorbox{species}{species}{\iminatevol{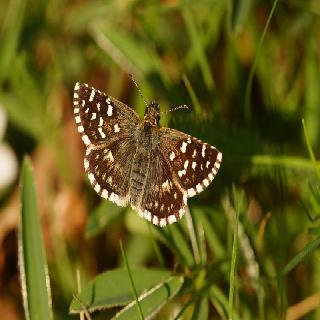}} &
\fcolorbox{genus}{genus}{\iminatevol{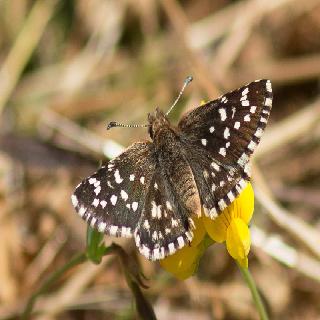}} &
\fcolorbox{species}{species}{\iminatevol{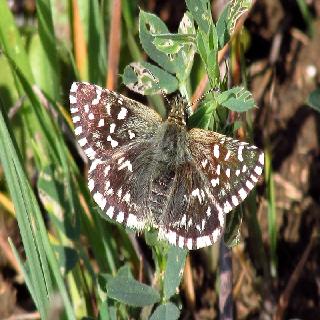}} &
\fcolorbox{genus}{genus}{\iminatevol{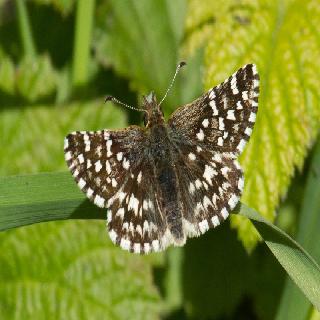}} &
\fcolorbox{genus}{genus}{\iminatevol{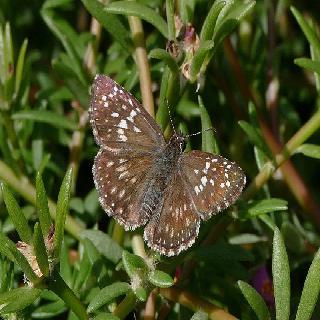}} &
\fcolorbox{genus}{genus}{\iminatevol{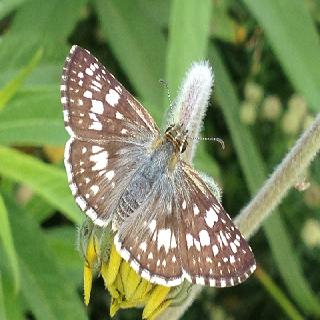}} &
\fcolorbox{genus}{genus}{\iminatevol{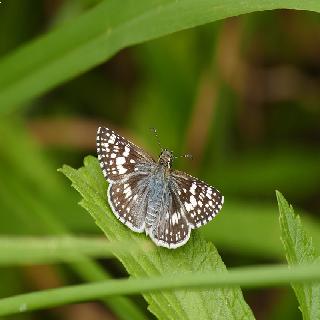}} \\
\midrule
\multirow{2}{2em}{\rotatebox{90}{\textcolor{order}{order}}} & \rotatebox{90}{Baseline} &
\href{https://www.inaturalist.org/photos/2609440}{\iminatevol{baseline/419/kingdom/query_classe_1092.jpg}} &
\fcolorbox{family}{family}{\iminatevol{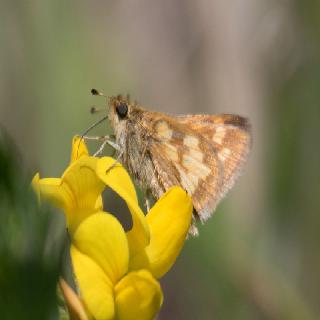}} &
\fcolorbox{family}{family}{\iminatevol{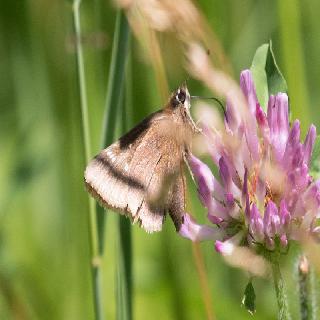}} &
\fcolorbox{order}{order}{\iminatevol{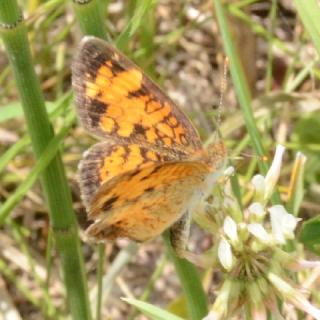}} &
\fcolorbox{genus}{genus}{\iminatevol{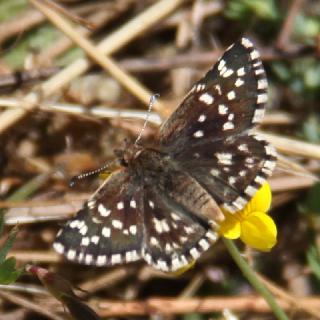}} &
\fcolorbox{family}{family}{\iminatevol{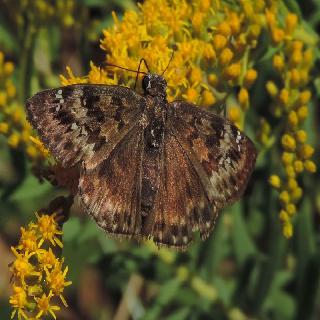}} &
\fcolorbox{species}{species}{\iminatevol{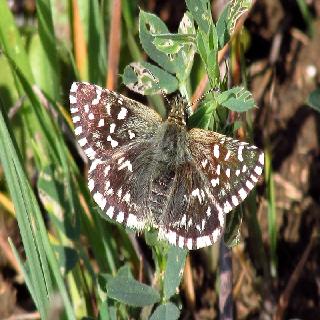}} &
\fcolorbox{family}{family}{\iminatevol{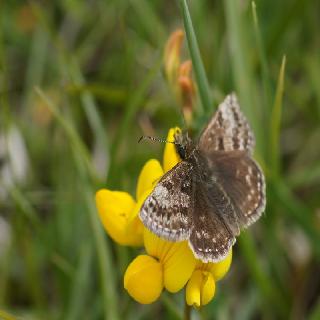}} &
\fcolorbox{genus}{genus}{\iminatevol{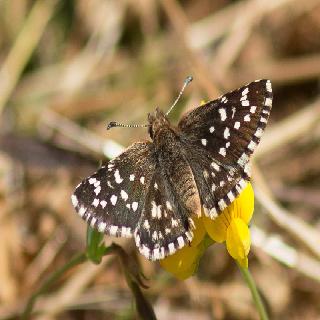}} &
\fcolorbox{family}{family}{\iminatevol{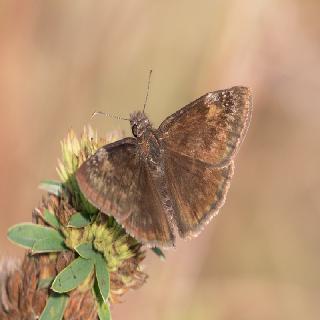}} &
\fcolorbox{family}{family}{\iminatevol{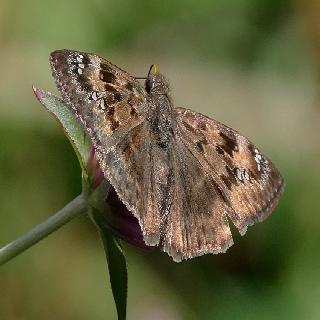}} \\
& \rotatebox{90}{\ours} & &
\href{https://www.inaturalist.org/photos/3434401}{\fcolorbox{genus}{genus}{\iminatevol{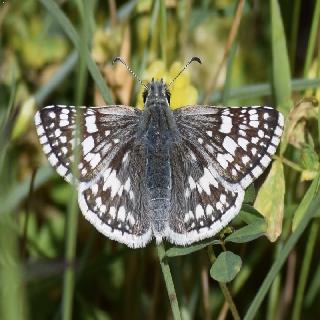}}} &
\fcolorbox{species}{species}{\iminatevol{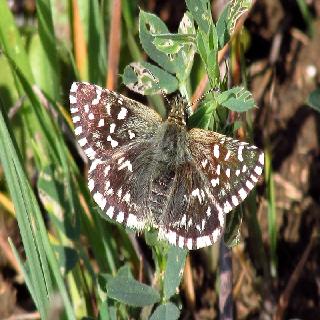}} &
\fcolorbox{genus}{genus}{\iminatevol{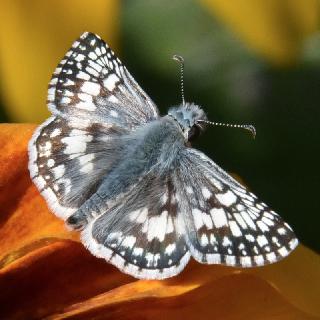}} &
\href{https://www.inaturalist.org/observations/5939102}{\fcolorbox{species}{species}{\iminatevol{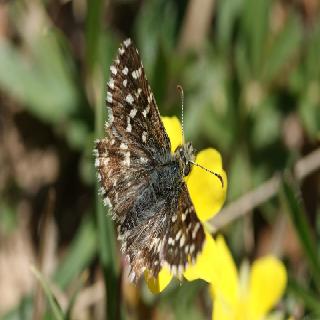}}} &
\fcolorbox{genus}{genus}{\iminatevol{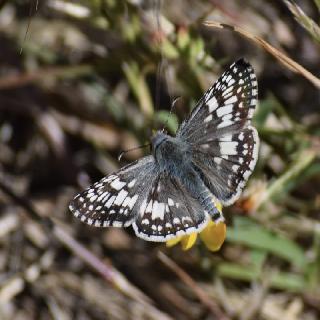}} &
\fcolorbox{genus}{genus}{\iminatevol{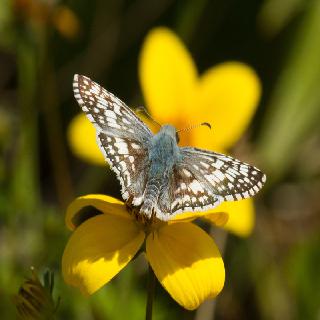}} &
\fcolorbox{genus}{genus}{\iminatevol{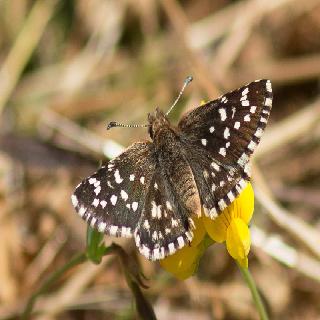}} &
\fcolorbox{genus}{genus}{\iminatevol{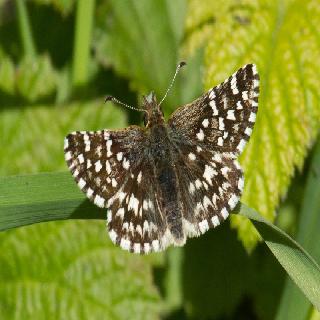}} &
\fcolorbox{family}{family}{\iminatevol{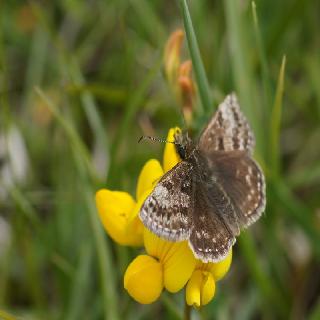}} &
\href{https://www.inaturalist.org/observations/1953410}{\fcolorbox{genus}{genus}{\iminatevol{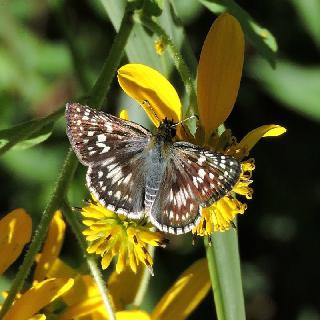}}} \\
\midrule
\multirow{2}{2em}{\rotatebox{90}{\textcolor{family}{family}}} & \rotatebox{90}{Baseline} &
\href{https://www.inaturalist.org/photos/2609440}{\iminatevol{baseline/419/kingdom/query_classe_1092.jpg}} &
\fcolorbox{genus}{genus}{\iminatevol{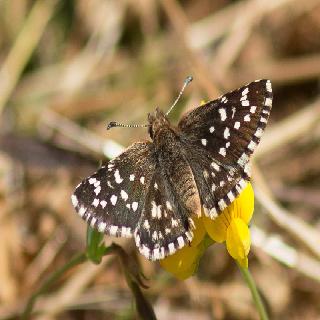}} &
\fcolorbox{genus}{genus}{\iminatevol{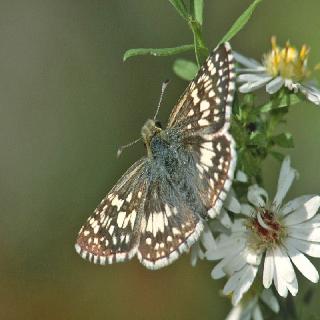}} &
\fcolorbox{species}{species}{\iminatevol{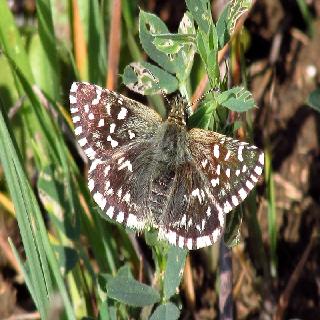}} &
\fcolorbox{genus}{genus}{\iminatevol{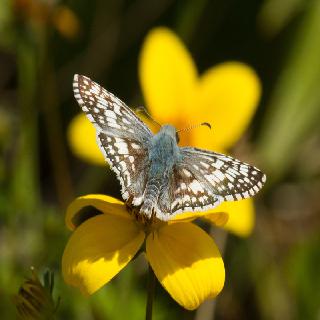}} &
\fcolorbox{genus}{genus}{\iminatevol{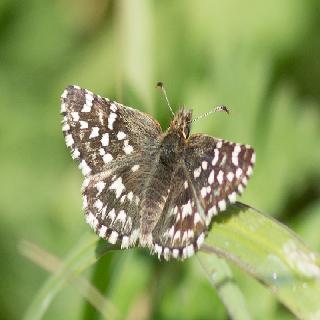}} &
\fcolorbox{genus}{genus}{\iminatevol{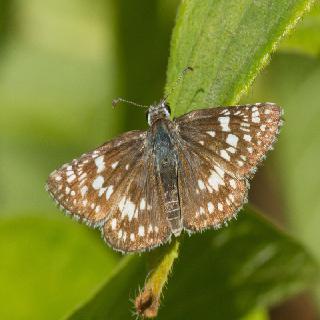}} &
\fcolorbox{species}{species}{\iminatevol{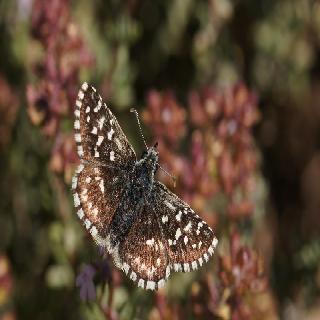}} &
\fcolorbox{genus}{genus}{\iminatevol{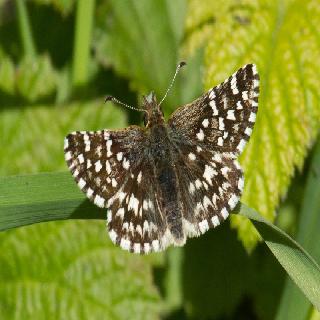}} &
\fcolorbox{species}{species}{\iminatevol{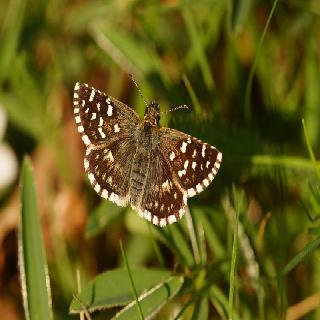}} &
\fcolorbox{genus}{genus}{\iminatevol{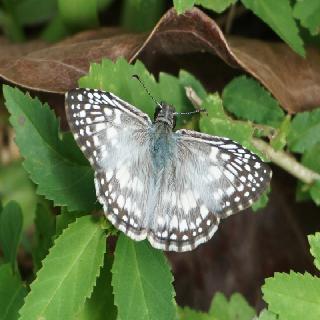}} \\
& \rotatebox{90}{\ours} & &
\fcolorbox{family}{family}{\iminatevol{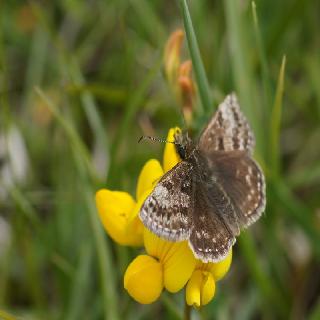}} &
\fcolorbox{genus}{genus}{\iminatevol{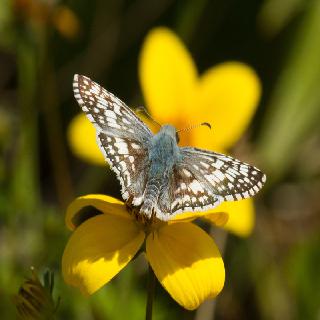}} &
\fcolorbox{genus}{genus}{\iminatevol{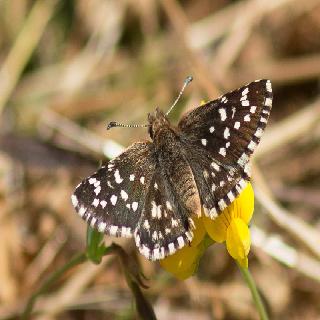}} &
\href{https://www.inaturalist.org/observations/5939102}{\fcolorbox{species}{species}{\iminatevol{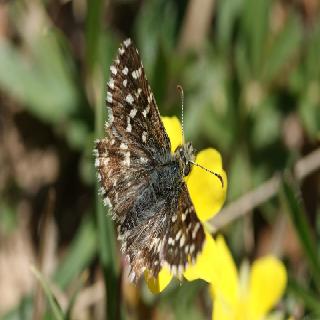}}} &
\fcolorbox{genus}{genus}{\iminatevol{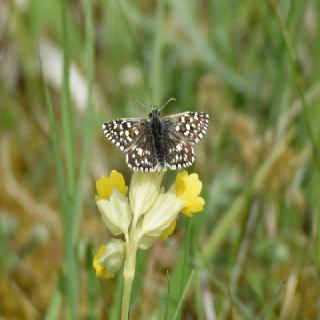}} &
\fcolorbox{genus}{genus}{\iminatevol{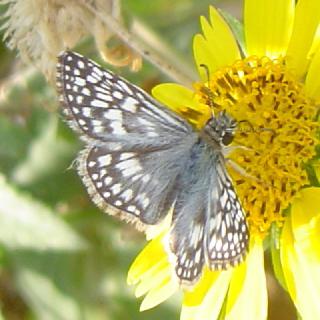}} &
\fcolorbox{species}{species}{\iminatevol{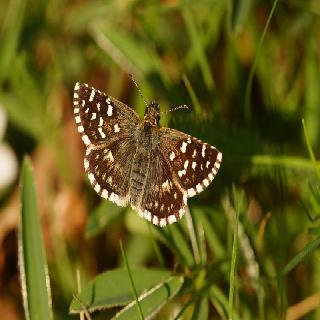}} &
\fcolorbox{species}{species}{\iminatevol{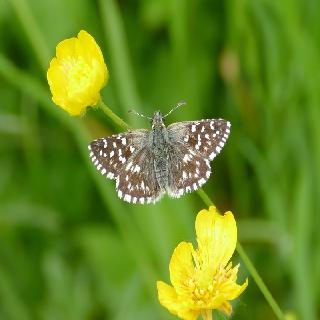}} &
\fcolorbox{genus}{genus}{\iminatevol{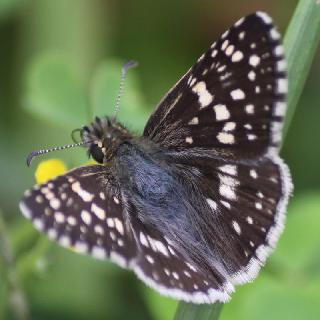}} &
\fcolorbox{genus}{genus}{\iminatevol{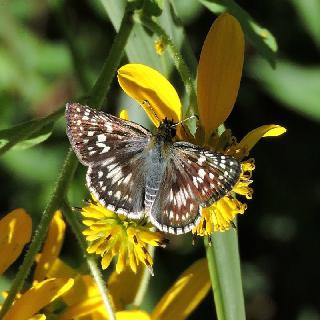}} \\
\midrule
\multirow{2}{2em}{\rotatebox{90}{\textcolor{genus}{genus}}} & \rotatebox{90}{Baseline} &
\href{https://www.inaturalist.org/photos/2609440}{\iminatevol{baseline/419/kingdom/query_classe_1092.jpg}} &
\fcolorbox{species}{species}{\iminatevol{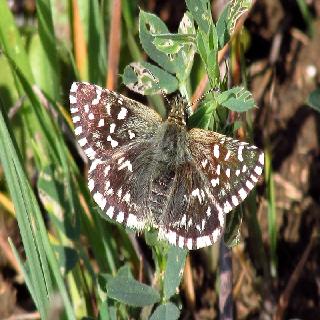}} &
\fcolorbox{genus}{genus}{\iminatevol{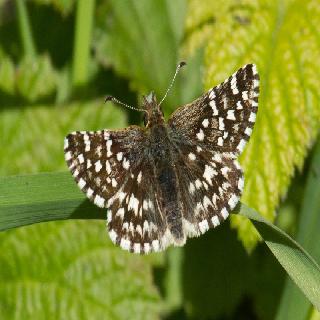}} &
\fcolorbox{genus}{genus}{\iminatevol{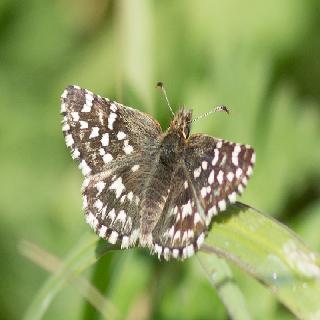}} &
\fcolorbox{species}{species}{\iminatevol{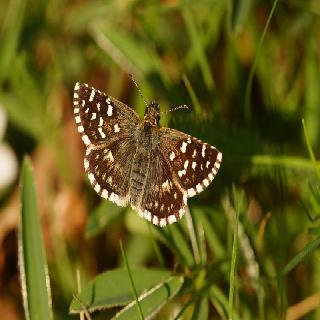}} &
\fcolorbox{genus}{genus}{\iminatevol{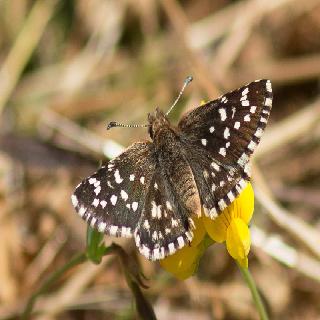}} &
\fcolorbox{species}{species}{\iminatevol{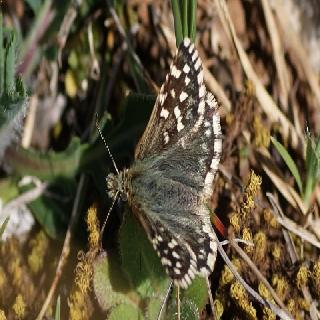}} &
\fcolorbox{species}{species}{\iminatevol{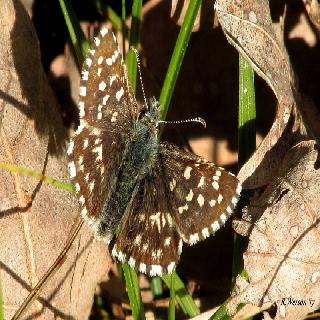}} &
\fcolorbox{species}{species}{\iminatevol{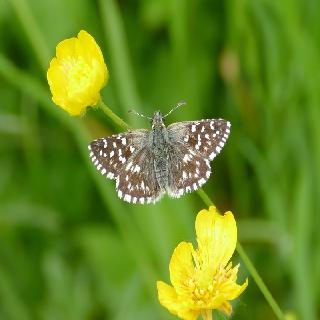}} &
\fcolorbox{genus}{genus}{\iminatevol{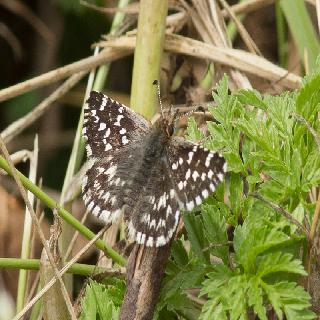}} &
\href{https://www.inaturalist.org/photos/3615109}{\fcolorbox{genus}{genus}{\iminatevol{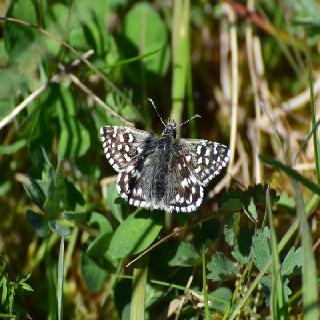}}} \\
& \rotatebox{90}{\ours} & &
\fcolorbox{species}{species}{\iminatevol{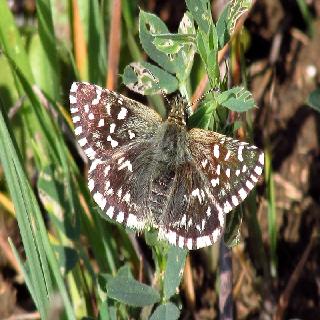}} &
\fcolorbox{species}{species}{\iminatevol{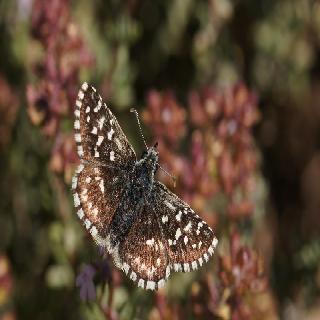}} &
\fcolorbox{species}{species}{\iminatevol{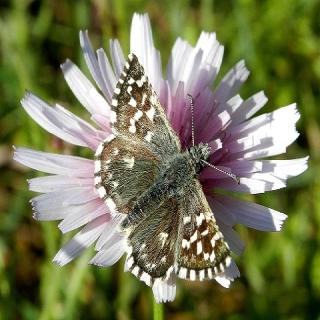}} &
\fcolorbox{species}{species}{\iminatevol{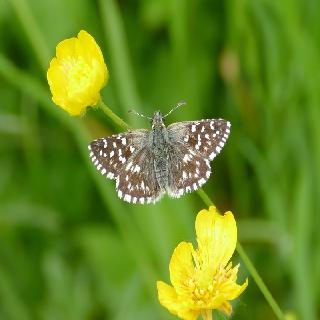}} &
\fcolorbox{genus}{genus}{\iminatevol{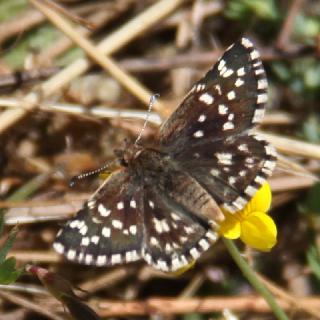}} &
\fcolorbox{species}{species}{\iminatevol{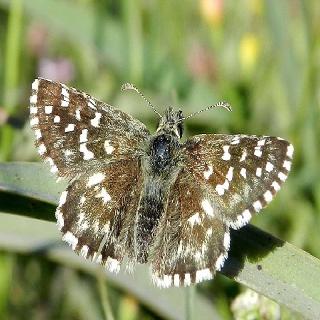}} &
\href{https://www.inaturalist.org/photos/3615109}{\fcolorbox{genus}{genus}{\iminatevol{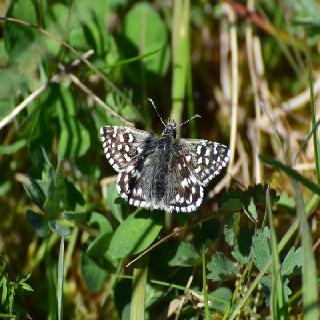}}}&
\fcolorbox{species}{species}{\iminatevol{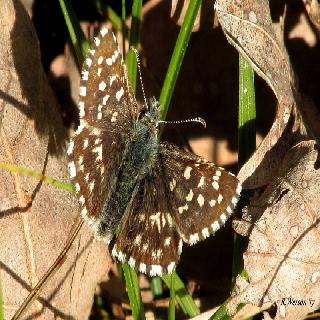}} &
\fcolorbox{genus}{genus}{\iminatevol{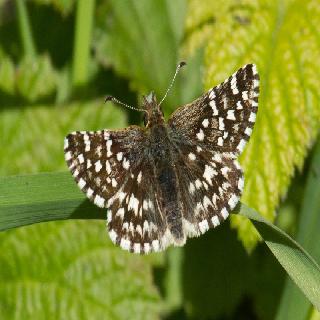}} &
\fcolorbox{species}{species}{\iminatevol{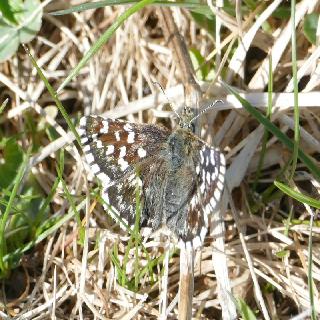}} \\
\midrule
\multirow{2}{2em}{\rotatebox{90}{\textcolor{species}{species}}} & \rotatebox{90}{Baseline} &
\href{https://www.inaturalist.org/photos/2609440}{\iminatevol{baseline/419/kingdom/query_classe_1092.jpg}} &
\fcolorbox{species}{species}{\iminatevol{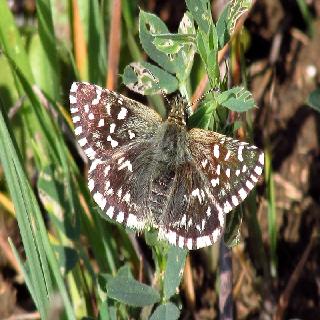}} &
\fcolorbox{species}{species}{\iminatevol{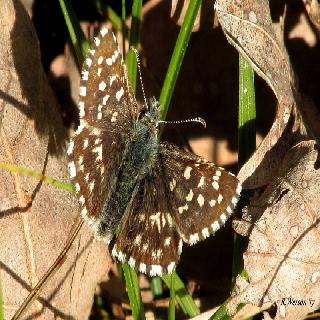}} &
\fcolorbox{species}{species}{\iminatevol{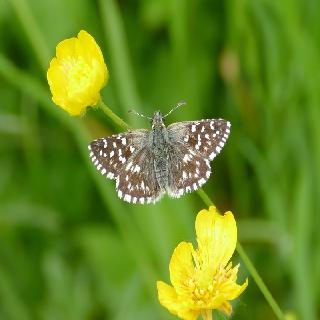}} &
\fcolorbox{species}{species}{\iminatevol{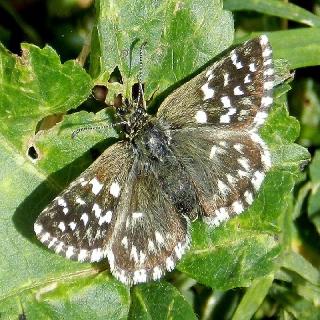}} &
\fcolorbox{species}{species}{\iminatevol{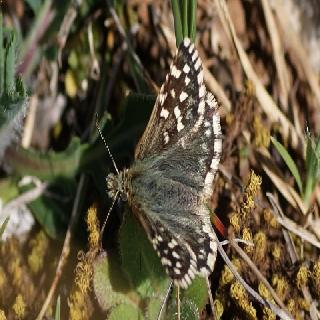}} &
\fcolorbox{species}{species}{\iminatevol{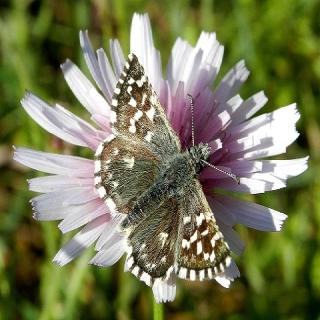}} &
\fcolorbox{species}{species}{\iminatevol{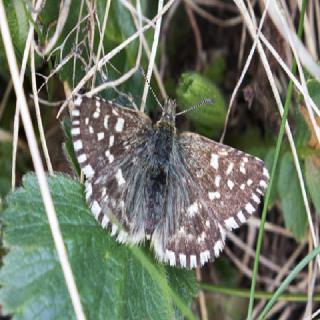}} &
\fcolorbox{genus}{genus}{\iminatevol{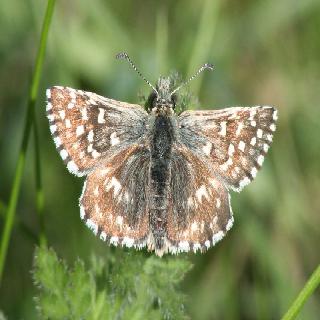}} &
\href{https://www.inaturalist.org/observations/5939102}{\fcolorbox{species}{species}{\iminatevol{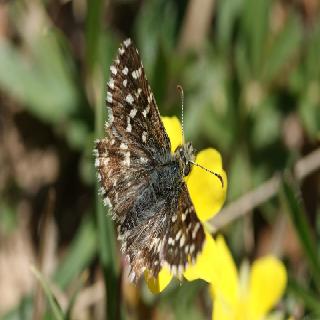}}} &
\href{https://www.inaturalist.org/observations/2939424}{\fcolorbox{species}{species}{\iminatevol{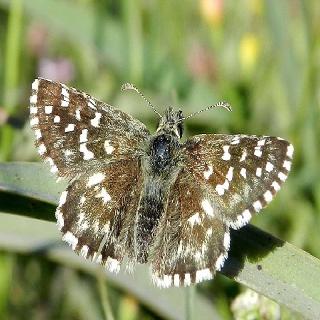}}} \\
& \rotatebox{90}{\ours} & &
\fcolorbox{species}{species}{\iminatevol{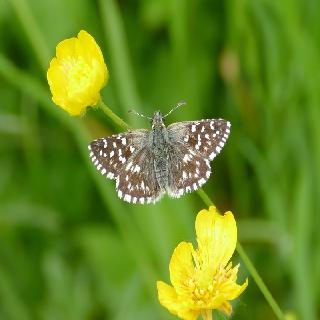}} &
\href{https://www.inaturalist.org/observations/2939424}{\fcolorbox{species}{species}{\iminatevol{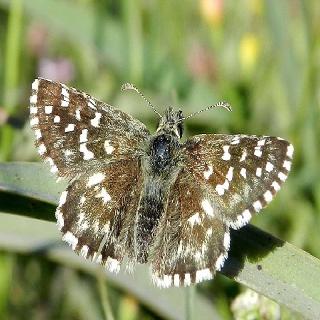}}}&
\fcolorbox{species}{species}{\iminatevol{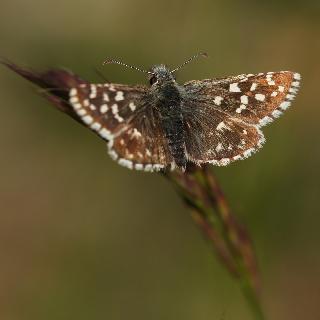}} &
\fcolorbox{species}{species}{\iminatevol{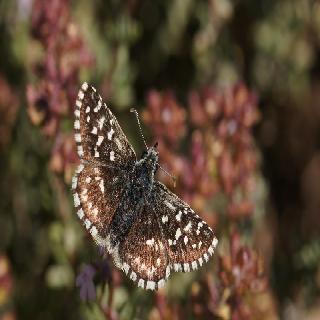}} &
\fcolorbox{species}{species}{\iminatevol{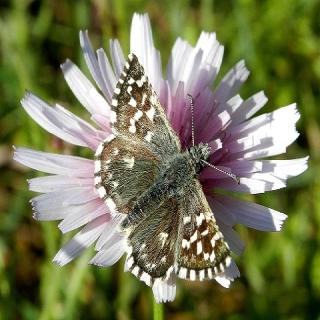}} &
\fcolorbox{species}{species}{\iminatevol{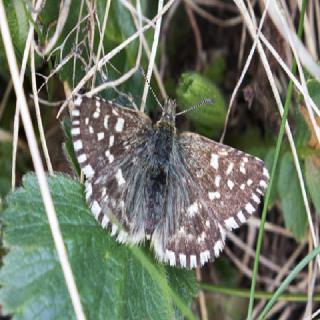}} &
\fcolorbox{species}{species}{\iminatevol{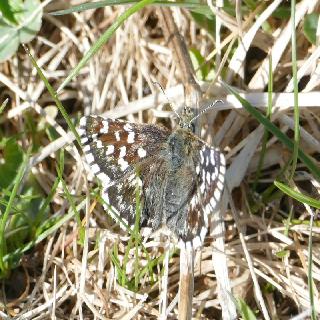}} &
\fcolorbox{species}{species}{\iminatevol{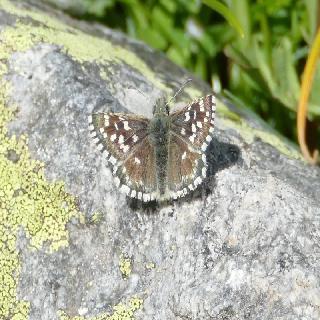}} &
\fcolorbox{species}{species}{\iminatevol{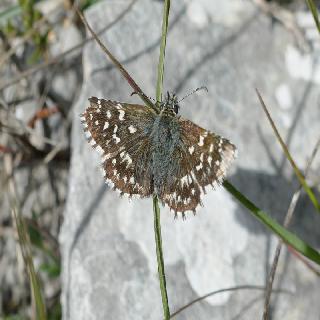}} &
\fcolorbox{species}{species}{\iminatevol{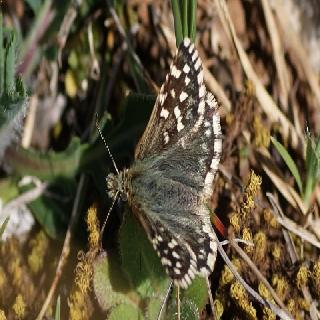}} \\
\bottomrule
\end{tabular}}

    \caption{We compare  \ours and Baseline for different training granularity. We rank the 10 closest images in the iNaturalist-2018 train set for a query image in the test set.
    The ranking is obtained with a cosine similarity on the features space of each of the two approaches. See Table~\ref{tab:copyright_appendix_2} for authors and image copyrights.  \label{fig:inat_ranking_evolution_appendix_2}}
    
\end{figure*}

\fboxrule=3pt
\fboxsep=0.1mm

\def \mysp {\hspace{2pt}}

\begin{figure*}[t]
    \centering
    \scalebox{0.8}{
\begin{tabular}{|c|c|c@{\mysp}c@{\mysp}c@{\mysp}c@{\mysp}c@{\mysp}c@{\mysp}c@{\mysp}c@{\mysp}c@{\mysp}c@{\mysp}c|}
\toprule
\multirow{2}{2em}{Train levels} &  \multirow{2}{3em}{Method} &  \multirow{2}{4em}{Query Image} & \multicolumn{10}{c}{Neighbours in train}\\
& & & Rank 1 & Rank 2 & Rank 3 & Rank 4 & Rank 5 & Rank 6 & Rank 7 & Rank 8 & Rank 9 & Rank 10\\
\midrule
\multirow{2}{2em}{\rotatebox{90}{\textcolor{kingdom}{kingdom}}} & \rotatebox{90}{Baseline} &
\iminatevol{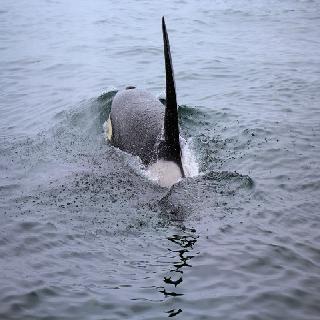} &
\fcolorbox{phylum}{phylum}{\iminatevol{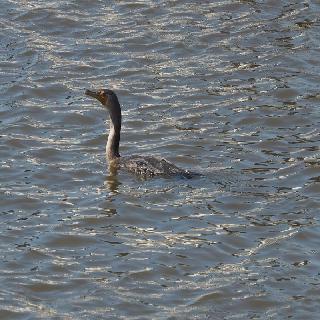}} &
\fcolorbox{phylum}{phylum}{\iminatevol{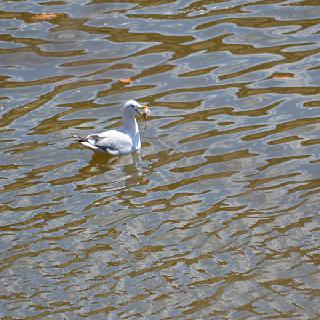}} &
\fcolorbox{phylum}{phylum}{\iminatevol{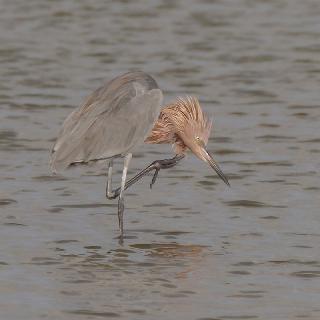}} &
\fcolorbox{phylum}{phylum}{\iminatevol{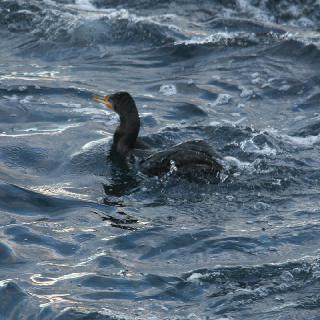}} &
\fcolorbox{phylum}{phylum}{\iminatevol{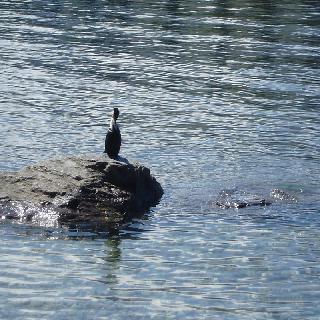}} &
\fcolorbox{order}{order}{\iminatevol{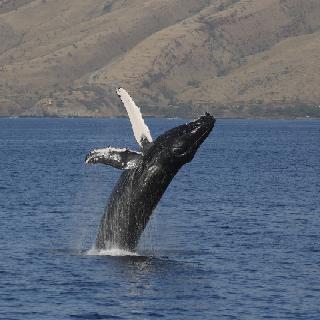}} &
\fcolorbox{order}{order}{\iminatevol{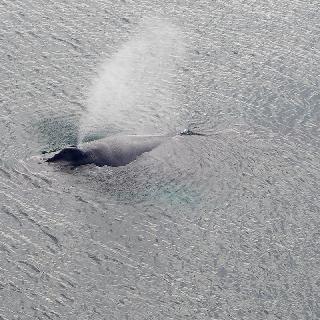}} &
\fcolorbox{phylum}{phylum}{\iminatevol{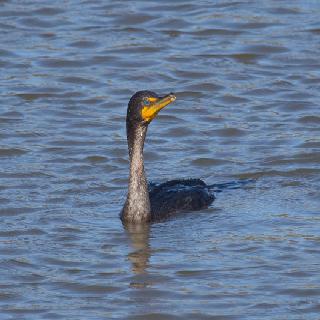}} &
\fcolorbox{phylum}{phylum}{\iminatevol{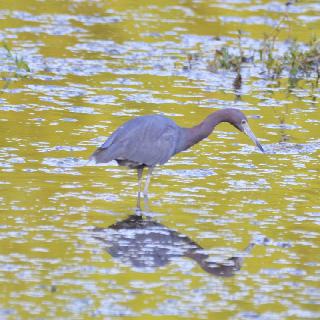}} &
\fcolorbox{phylum}{phylum}{\iminatevol{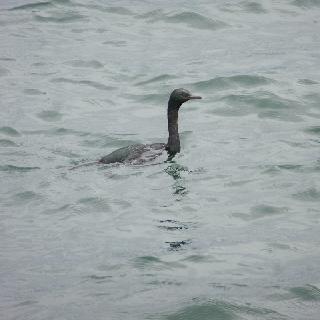}} \\
& \rotatebox{90}{\ours} & &
\fcolorbox{order}{order}{\iminatevol{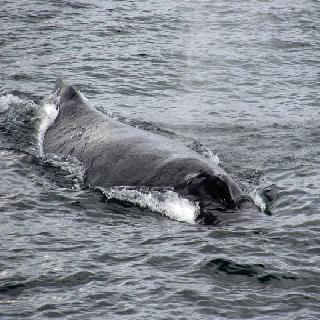}} &
\fcolorbox{order}{order}{\iminatevol{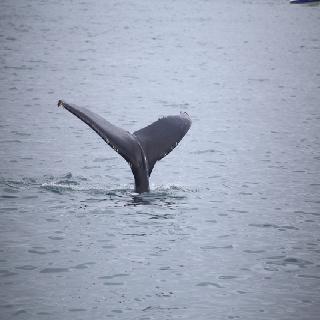}} &
\fcolorbox{order}{order}{\iminatevol{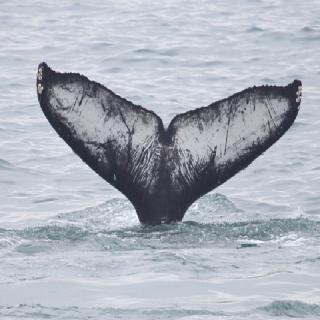}} &
\fcolorbox{order}{order}{\iminatevol{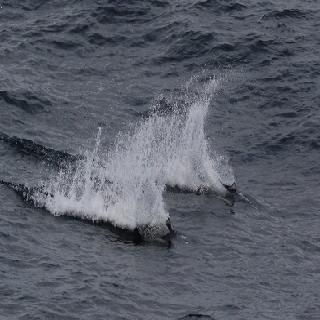}} &
\fcolorbox{phylum}{phylum}{\iminatevol{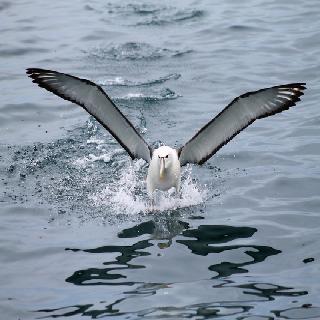}} &
\fcolorbox{family}{family}{\iminatevol{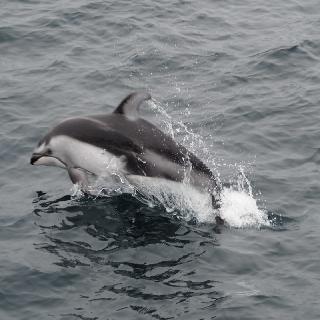}} &
\fcolorbox{order}{order}{\iminatevol{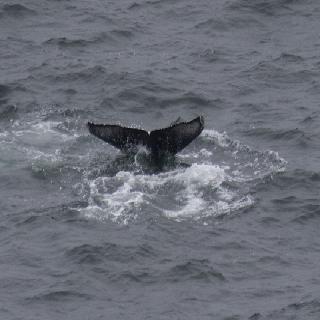}} &
\fcolorbox{order}{order}{\iminatevol{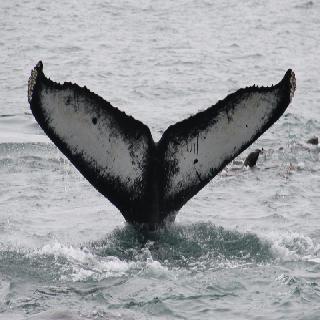}} &
\fcolorbox{order}{order}{\iminatevol{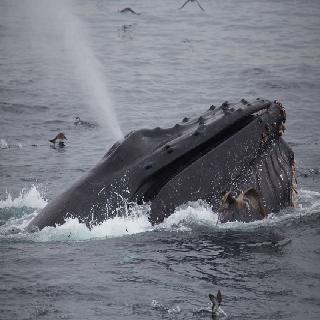}} &
\fcolorbox{order}{order}{\iminatevol{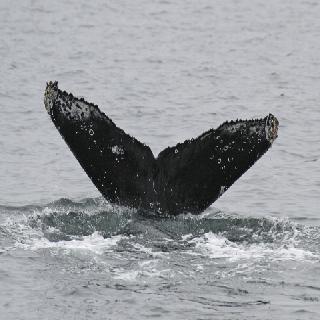}} \\
\midrule
\multirow{2}{2em}{\rotatebox{90}{\textcolor{phylum}{phylum}}} & \rotatebox{90}{Baseline} &
\iminatevol{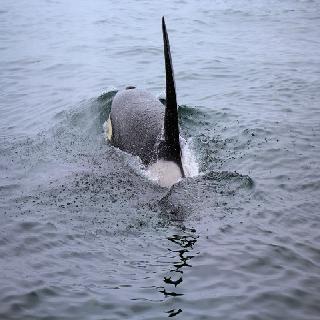} &
\fcolorbox{phylum}{phylum}{\iminatevol{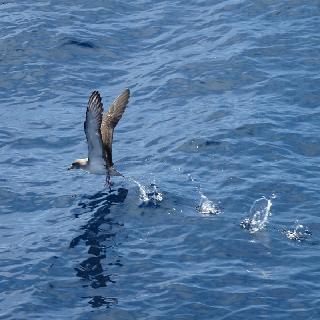}} &
\fcolorbox{phylum}{phylum}{\iminatevol{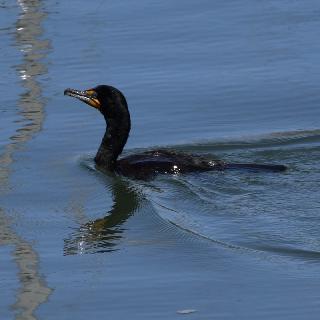}} &
\fcolorbox{phylum}{phylum}{\iminatevol{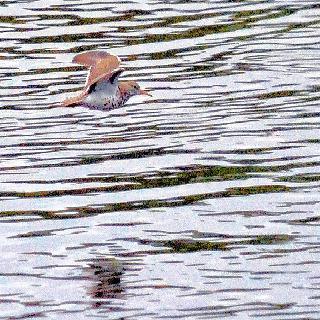}} &
\fcolorbox{phylum}{phylum}{\iminatevol{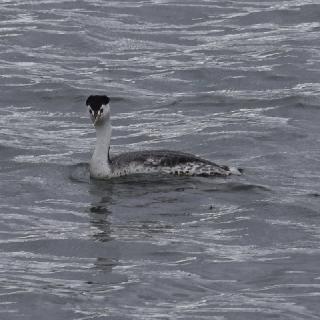}} &
\fcolorbox{phylum}{phylum}{\iminatevol{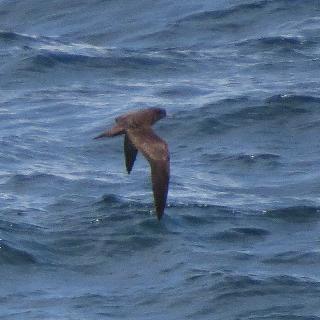}} &
\fcolorbox{phylum}{phylum}{\iminatevol{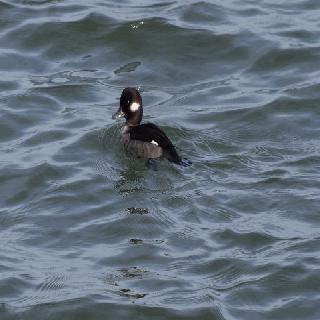}} &
\fcolorbox{phylum}{phylum}{\iminatevol{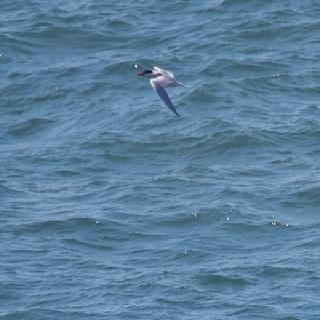}} &
\fcolorbox{phylum}{phylum}{\iminatevol{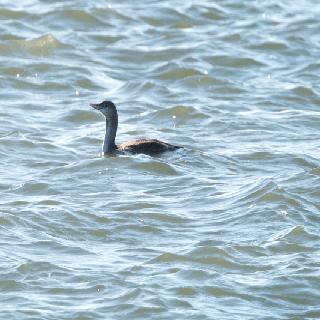}} &
\fcolorbox{family}{family}{\iminatevol{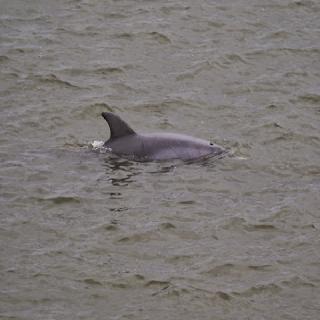}} &
\fcolorbox{phylum}{phylum}{\iminatevol{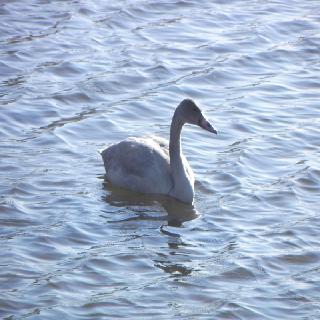}} \\
& \rotatebox{90}{\ours} & &
\fcolorbox{order}{order}{\iminatevol{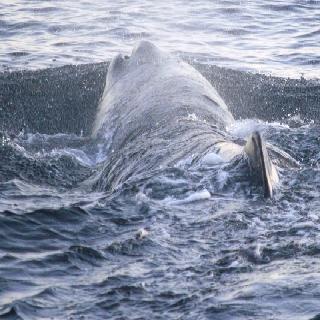}} &
\fcolorbox{class}{class}{\iminatevol{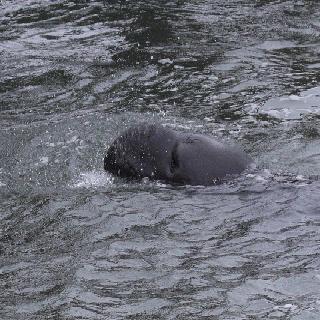}} &
\fcolorbox{order}{order}{\iminatevol{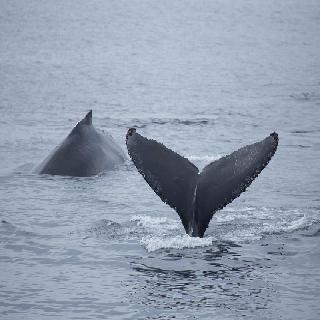}} &
\fcolorbox{order}{order}{\iminatevol{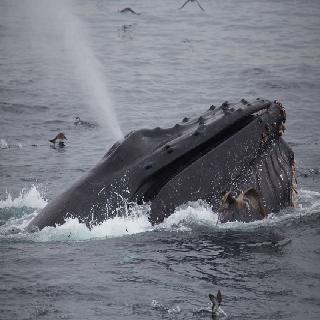}} &
\fcolorbox{family}{family}{\iminatevol{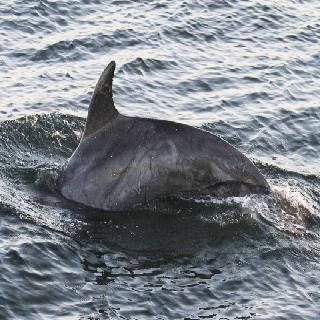}} &
\fcolorbox{order}{order}{\iminatevol{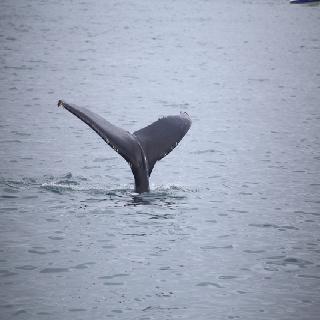}} &
\fcolorbox{order}{order}{\iminatevol{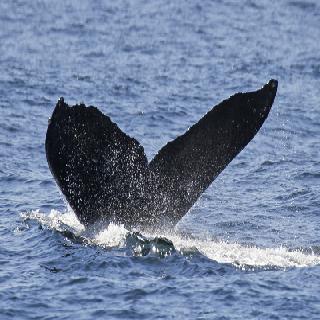}} &
\fcolorbox{order}{order}{\iminatevol{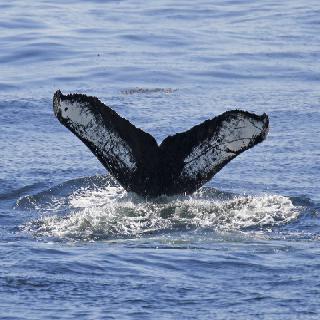}} &
\fcolorbox{family}{family}{\iminatevol{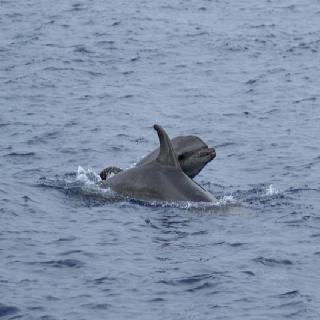}} &
\fcolorbox{order}{order}{\iminatevol{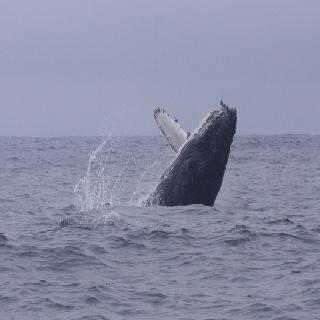}} \\
\midrule
\multirow{2}{2em}{\rotatebox{90}{\textcolor{class}{class}}} & \rotatebox{90}{Baseline} &
\iminatevol{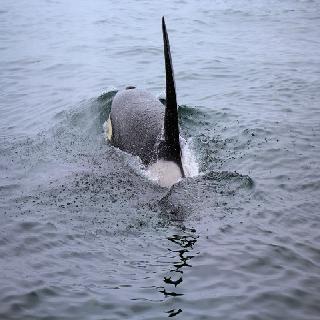} &
\fcolorbox{family}{family}{\iminatevol{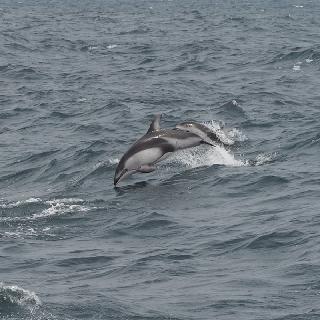}} &
\fcolorbox{class}{class}{\iminatevol{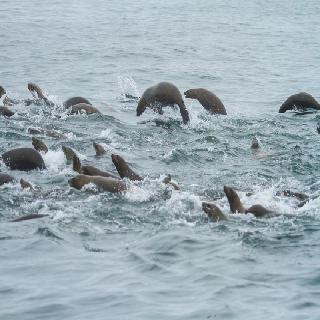}} &
\fcolorbox{order}{order}{\iminatevol{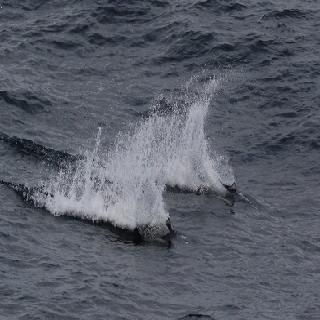}} &
\fcolorbox{order}{order}{\iminatevol{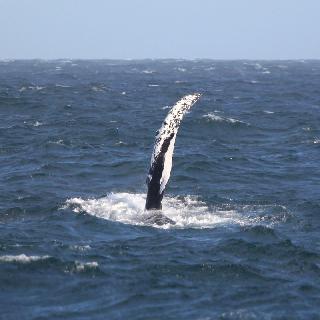}} &
\fcolorbox{family}{family}{\iminatevol{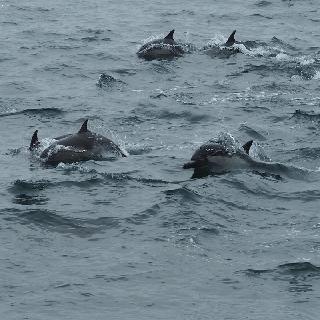}} &
\fcolorbox{family}{family}{\iminatevol{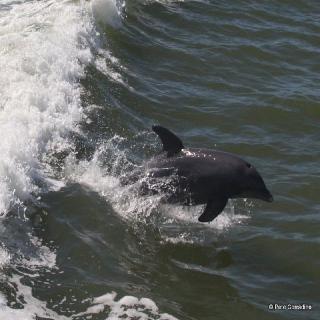}} &
\fcolorbox{class}{class}{\iminatevol{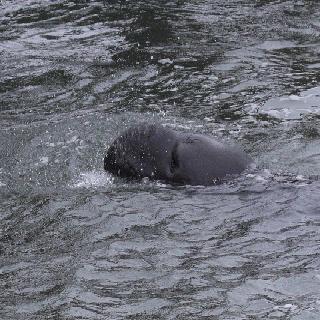}} &
\fcolorbox{order}{order}{\iminatevol{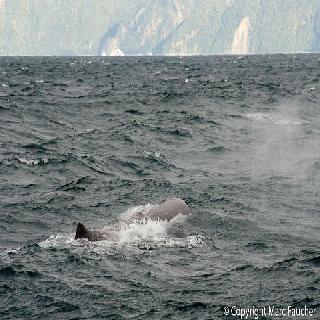}} &
\fcolorbox{order}{order}{\iminatevol{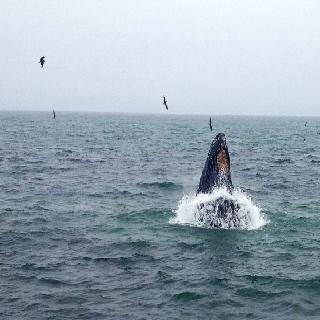}} &
\fcolorbox{class}{class}{\iminatevol{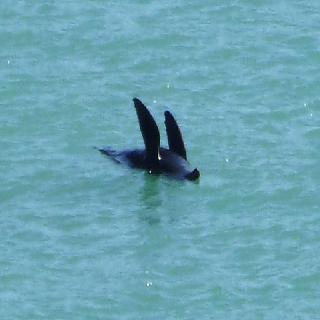}} \\
& \rotatebox{90}{\ours} & &
\fcolorbox{family}{family}{\iminatevol{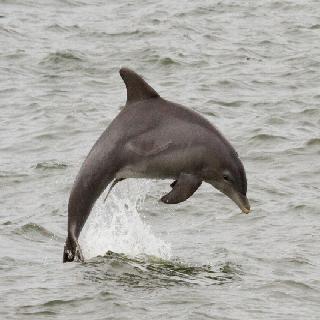}} &
\fcolorbox{order}{order}{\iminatevol{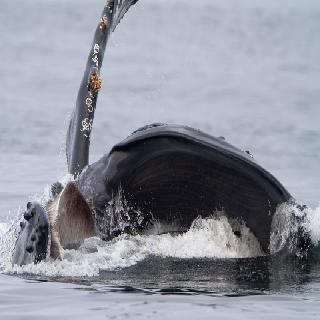}} &
\fcolorbox{order}{order}{\iminatevol{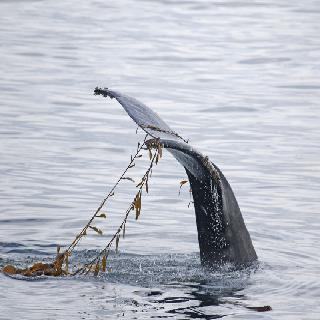}} &
\fcolorbox{order}{order}{\iminatevol{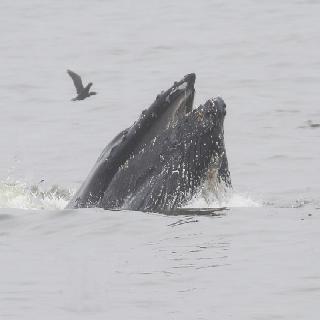}} &
\fcolorbox{order}{order}{\iminatevol{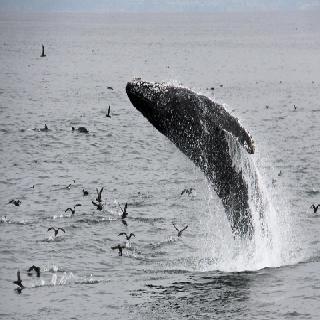}} &
\fcolorbox{order}{order}{\iminatevol{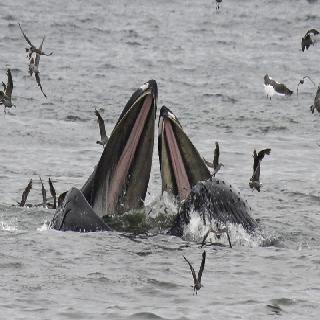}} &
\fcolorbox{order}{order}{\iminatevol{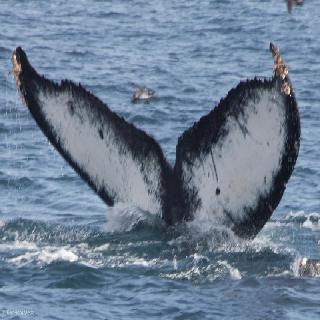}} &
\fcolorbox{order}{order}{\iminatevol{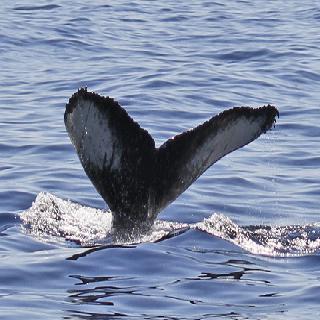}} &
\fcolorbox{order}{order}{\iminatevol{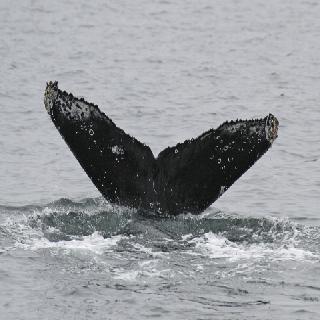}} &
\fcolorbox{class}{class}{\iminatevol{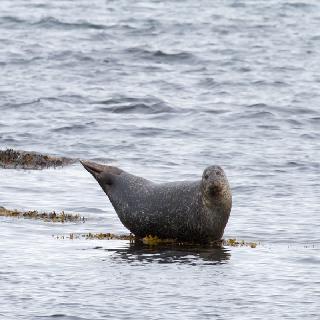}} \\
\midrule
\multirow{2}{2em}{\rotatebox{90}{\textcolor{order}{order}}} & \rotatebox{90}{Baseline} &
\iminatevol{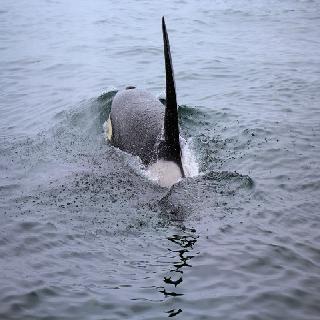} &
\fcolorbox{family}{family}{\iminatevol{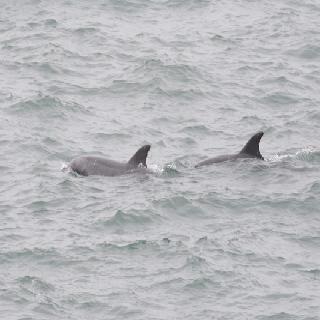}} &
\fcolorbox{family}{family}{\iminatevol{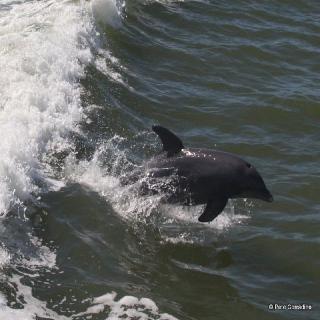}} &
\fcolorbox{order}{order}{\iminatevol{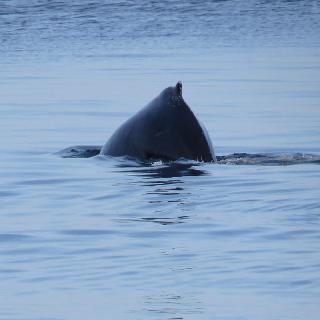}} &
\fcolorbox{species}{species}{\iminatevol{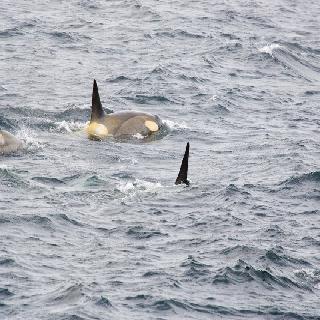}} &
\fcolorbox{order}{order}{\iminatevol{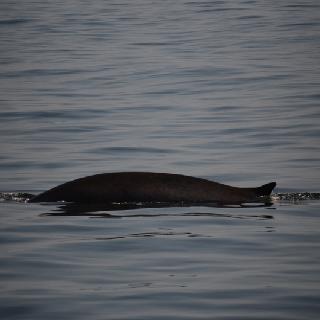}} &
\fcolorbox{family}{family}{\iminatevol{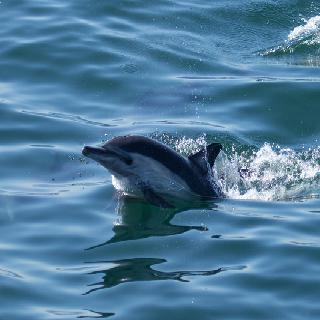}} &
\fcolorbox{order}{order}{\iminatevol{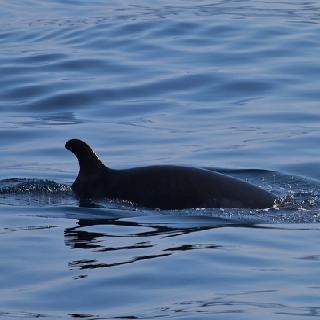}} &
\fcolorbox{family}{family}{\iminatevol{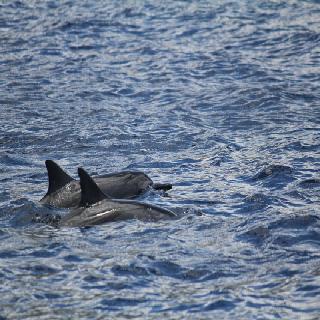}} &
\fcolorbox{class}{class}{\iminatevol{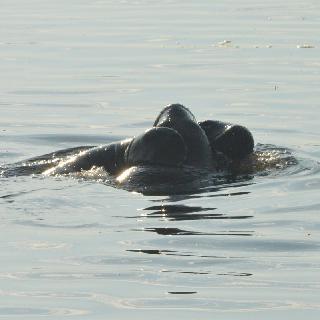}} &
\fcolorbox{family}{family}{\iminatevol{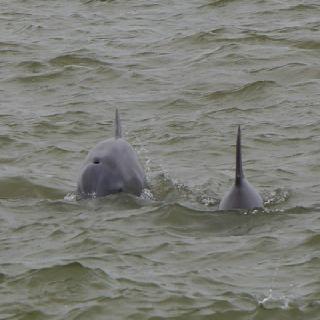}} \\
& \rotatebox{90}{\ours} & &
\fcolorbox{order}{order}{\iminatevol{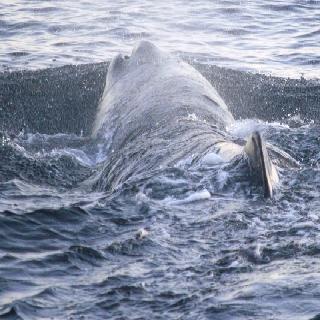}} &
\fcolorbox{family}{family}{\iminatevol{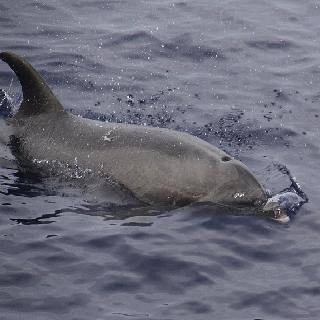}} &
\fcolorbox{family}{family}{\iminatevol{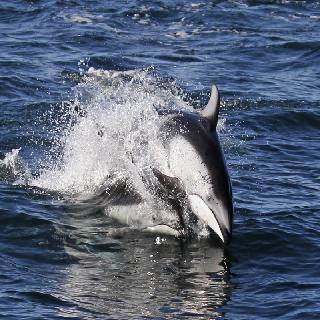}} &
\fcolorbox{order}{order}{\iminatevol{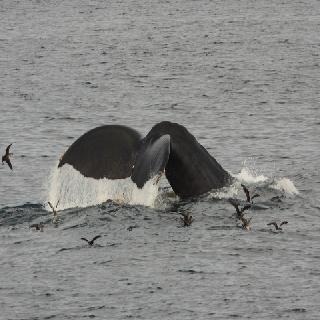}} &
\fcolorbox{order}{order}{\iminatevol{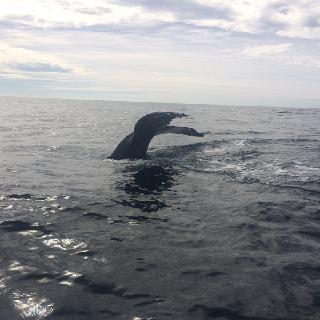}} &
\fcolorbox{family}{family}{\iminatevol{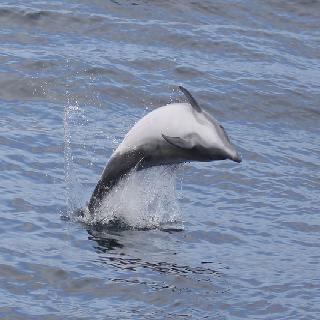}} &
\fcolorbox{family}{family}{\iminatevol{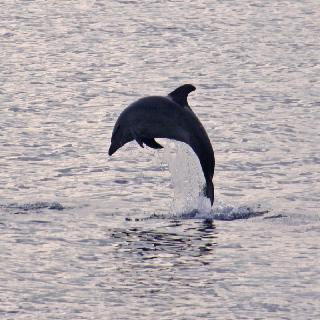}} &
\fcolorbox{order}{order}{\iminatevol{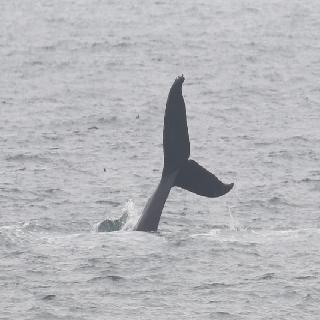}} &
\fcolorbox{order}{order}{\iminatevol{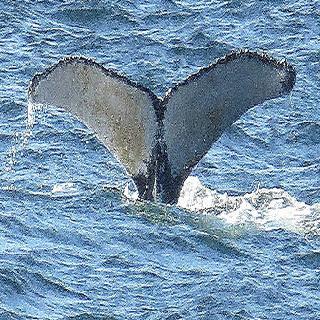}} &
\fcolorbox{order}{order}{\iminatevol{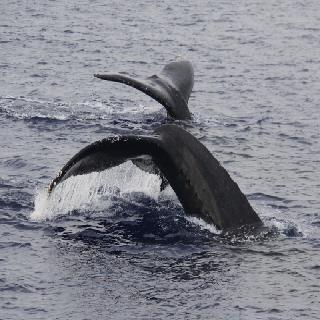}} \\
\midrule
\multirow{2}{2em}{\rotatebox{90}{\textcolor{family}{family}}} & \rotatebox{90}{Baseline} &
\iminatevol{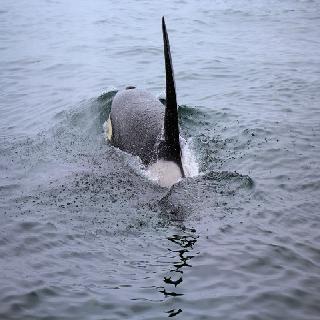} &
\fcolorbox{order}{order}{\iminatevol{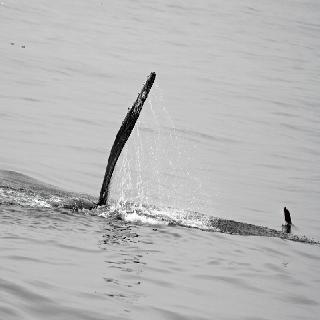}} &
\fcolorbox{class}{class}{\iminatevol{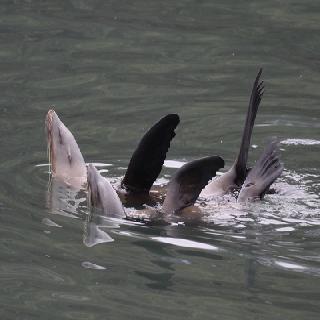}} &
\fcolorbox{family}{family}{\iminatevol{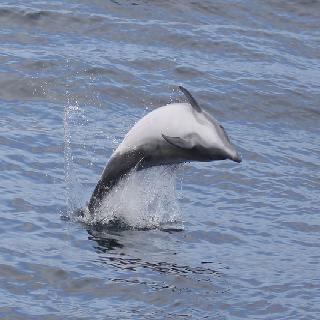}} &
\fcolorbox{species}{species}{\iminatevol{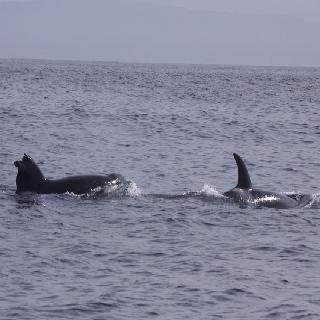}} &
\fcolorbox{species}{species}{\iminatevol{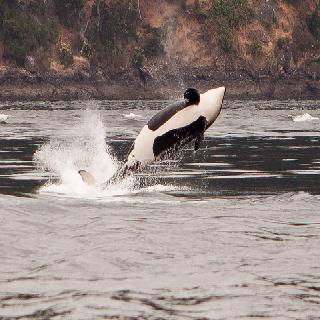}} &
\fcolorbox{family}{family}{\iminatevol{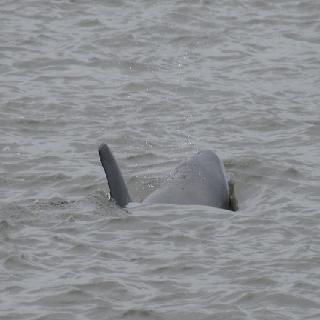}} &
\fcolorbox{order}{order}{\iminatevol{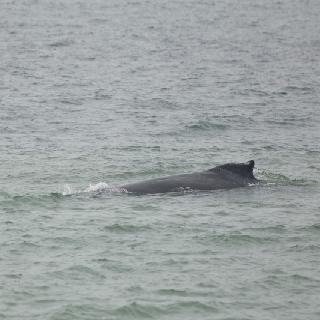}} &
\fcolorbox{species}{species}{\iminatevol{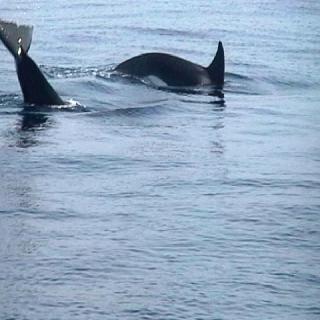}} &
\fcolorbox{family}{family}{\iminatevol{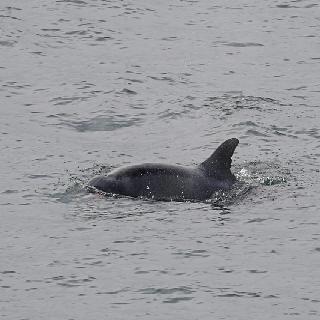}} &
\fcolorbox{order}{order}{\iminatevol{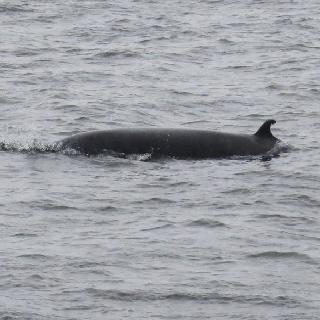}} \\
& \rotatebox{90}{\ours} & &
\fcolorbox{family}{family}{\iminatevol{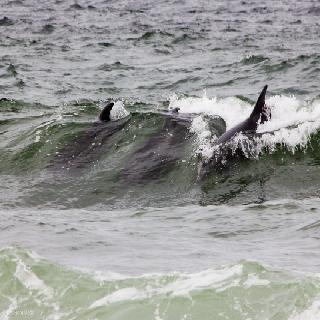}} &
\fcolorbox{family}{family}{\iminatevol{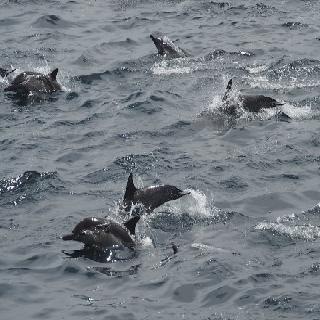}} &
\fcolorbox{family}{family}{\iminatevol{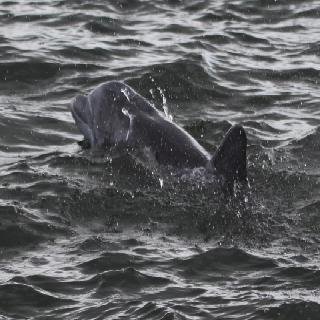}} &
\fcolorbox{species}{species}{\iminatevol{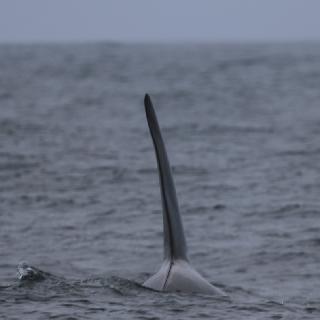}} &
\fcolorbox{family}{family}{\iminatevol{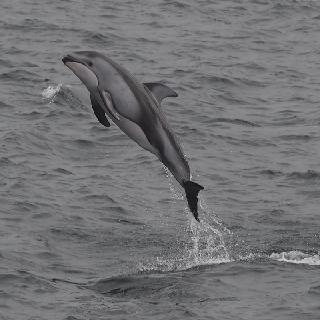}} &
\fcolorbox{species}{species}{\iminatevol{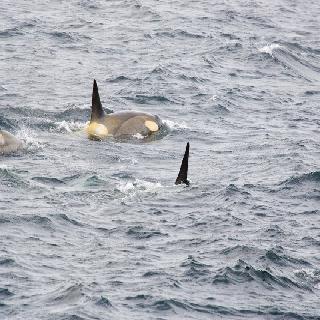}} &
\fcolorbox{family}{family}{\iminatevol{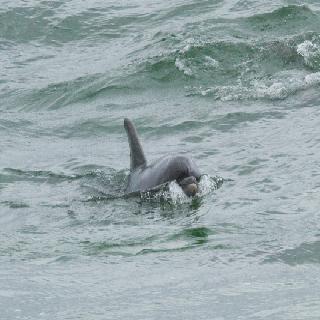}} &
\fcolorbox{family}{family}{\iminatevol{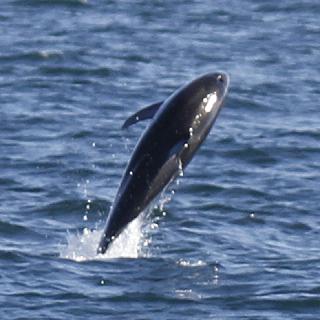}} &
\fcolorbox{family}{family}{\iminatevol{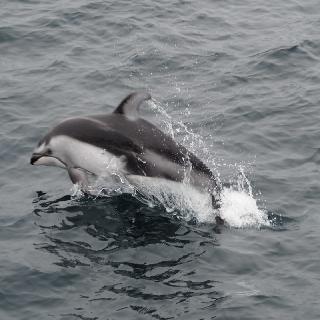}} &
\fcolorbox{family}{family}{\iminatevol{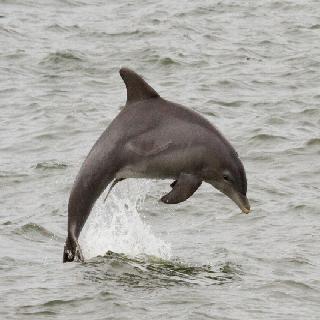}} \\
\midrule
\multirow{2}{2em}{\rotatebox{90}{\textcolor{genus}{genus}}} & \rotatebox{90}{Baseline} &
\iminatevol{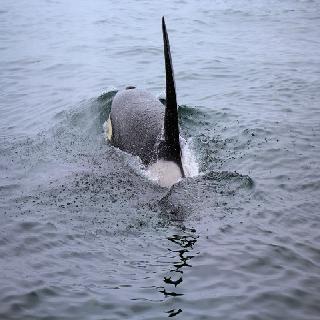} &
\fcolorbox{family}{family}{\iminatevol{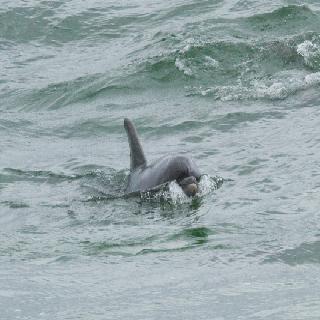}} &
\fcolorbox{family}{family}{\iminatevol{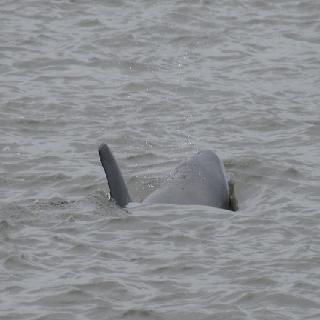}} &
\fcolorbox{order}{order}{\iminatevol{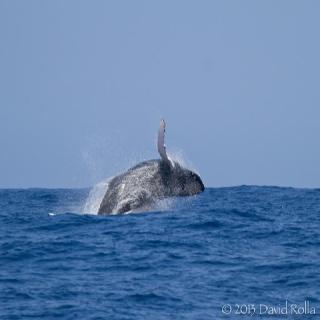}} &
\fcolorbox{family}{family}{\iminatevol{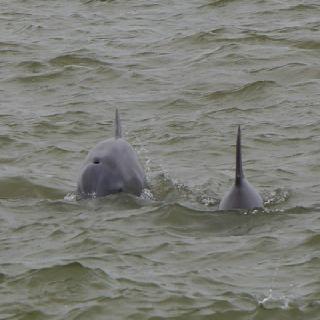}} &
\fcolorbox{order}{order}{\iminatevol{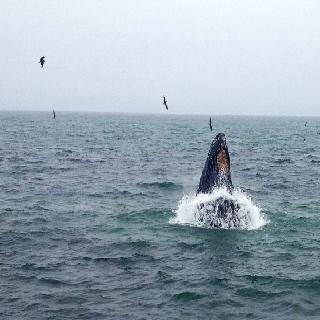}} &
\fcolorbox{order}{order}{\iminatevol{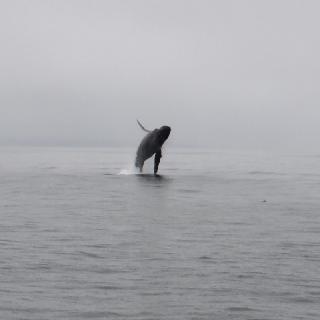}} &
\fcolorbox{class}{class}{\iminatevol{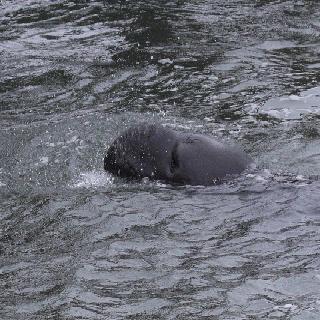}} &
\fcolorbox{order}{order}{\iminatevol{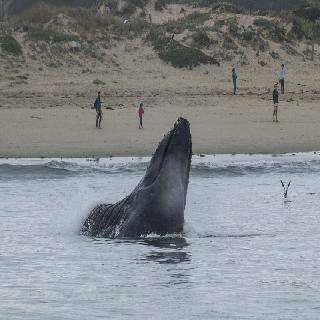}} &
\fcolorbox{family}{family}{\iminatevol{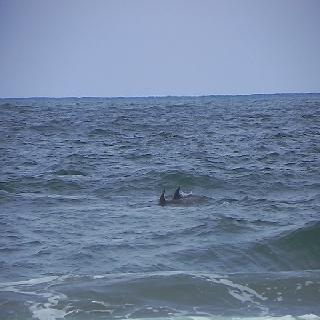}} &
\fcolorbox{species}{species}{\iminatevol{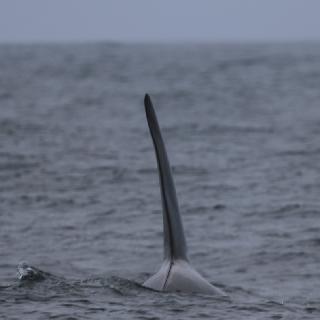}} \\
& \rotatebox{90}{\ours} & &
\fcolorbox{species}{species}{\iminatevol{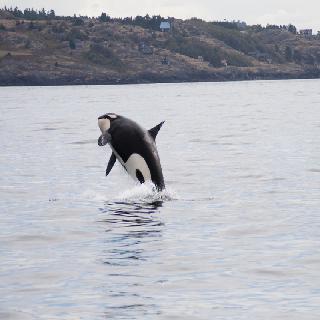}} &
\fcolorbox{species}{species}{\iminatevol{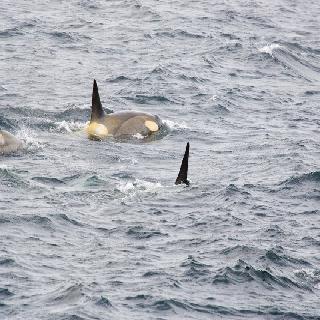}} &
\fcolorbox{species}{species}{\iminatevol{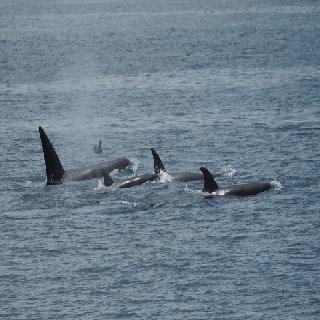}} &
\fcolorbox{species}{species}{\iminatevol{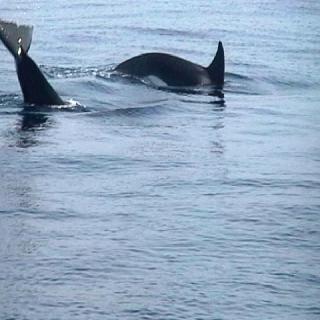}} &
\fcolorbox{species}{species}{\iminatevol{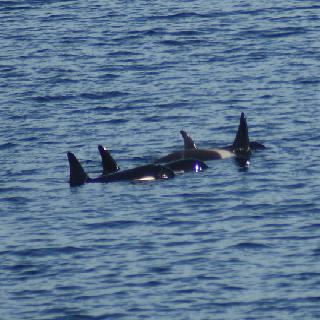}} &
\fcolorbox{species}{species}{\iminatevol{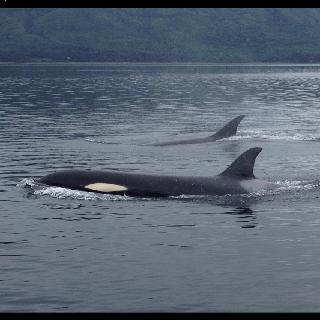}} &
\fcolorbox{species}{species}{\iminatevol{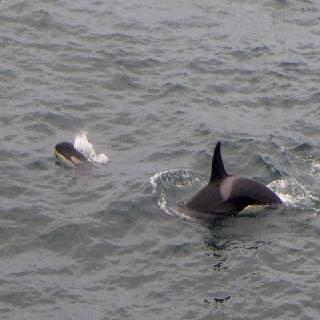}} &
\fcolorbox{species}{species}{\iminatevol{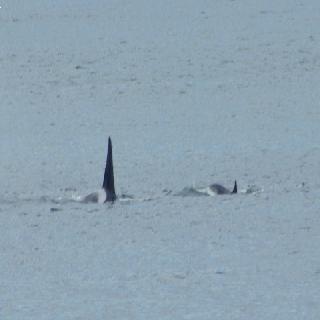}} &
\fcolorbox{species}{species}{\iminatevol{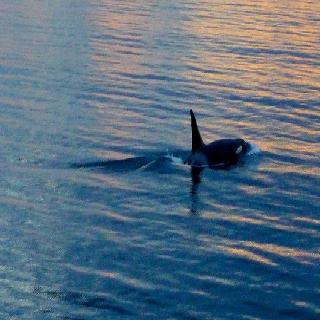}} &
\fcolorbox{species}{species}{\iminatevol{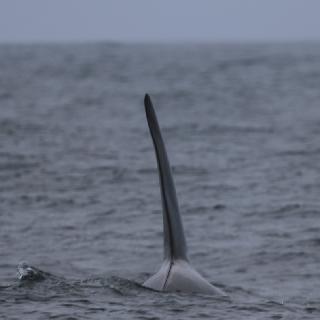}} \\
\midrule
\multirow{2}{2em}{\rotatebox{90}{\textcolor{species}{species}}} & \rotatebox{90}{Baseline} &
\iminatevol{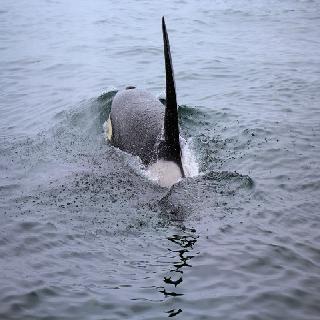} &
\fcolorbox{order}{order}{\iminatevol{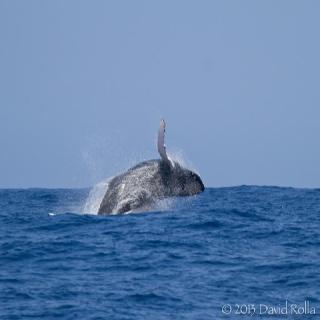}} &
\fcolorbox{order}{order}{\iminatevol{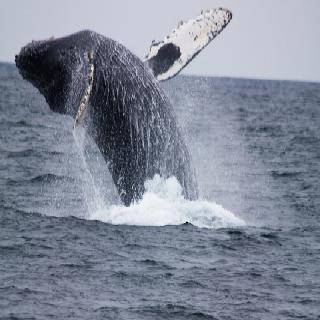}} &
\fcolorbox{order}{order}{\iminatevol{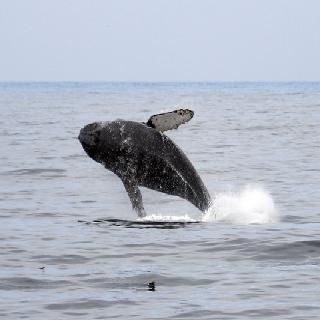}} &
\fcolorbox{order}{order}{\iminatevol{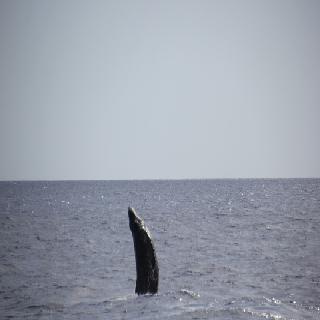}} &
\fcolorbox{order}{order}{\iminatevol{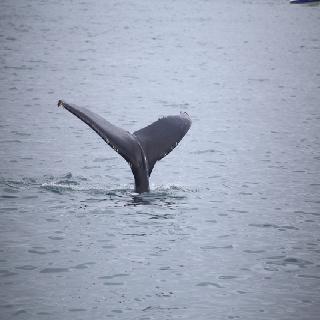}} &
\fcolorbox{order}{order}{\iminatevol{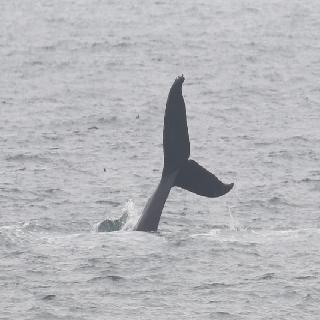}} &
\fcolorbox{family}{family}{\iminatevol{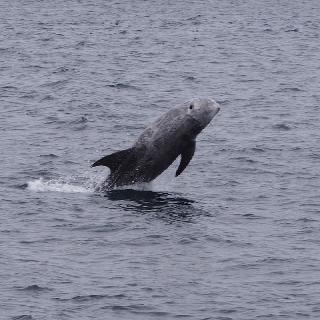}} &
\fcolorbox{order}{order}{\iminatevol{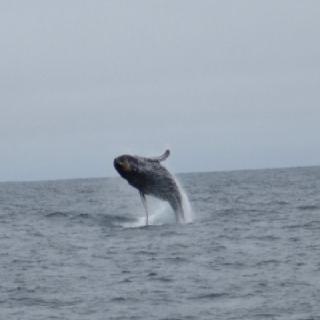}} &
\fcolorbox{order}{order}{\iminatevol{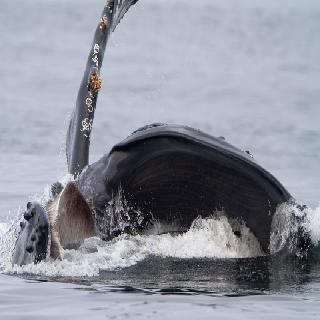}} &
\fcolorbox{class}{class}{\iminatevol{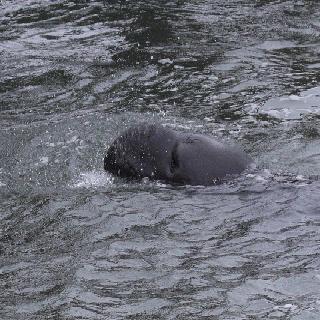}} \\
& \rotatebox{90}{\ours} & &
\fcolorbox{species}{species}{\iminatevol{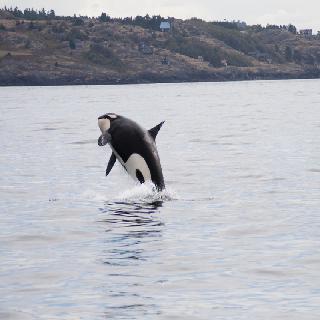}} &
\fcolorbox{species}{species}{\iminatevol{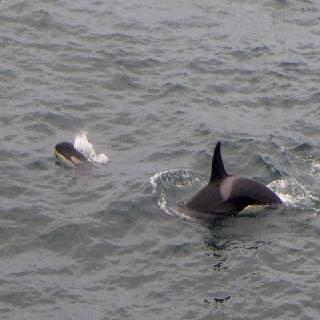}} &
\fcolorbox{species}{species}{\iminatevol{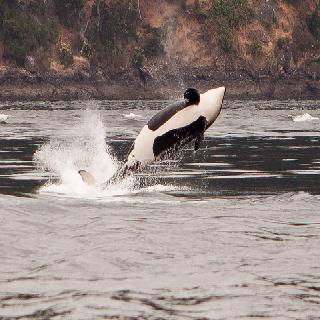}} &
\fcolorbox{species}{species}{\iminatevol{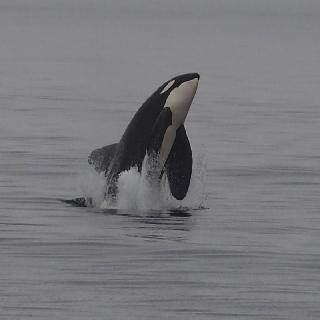}} &
\fcolorbox{species}{species}{\iminatevol{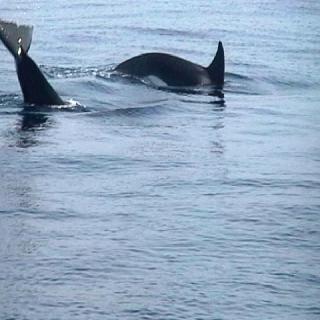}} &
\fcolorbox{species}{species}{\iminatevol{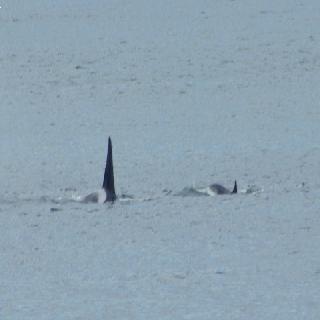}} &
\fcolorbox{species}{species}{\iminatevol{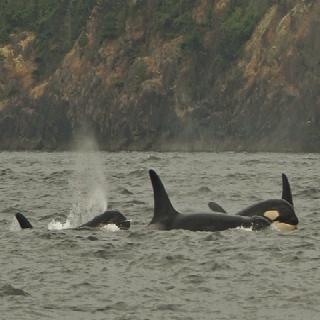}} &
\fcolorbox{species}{species}{\iminatevol{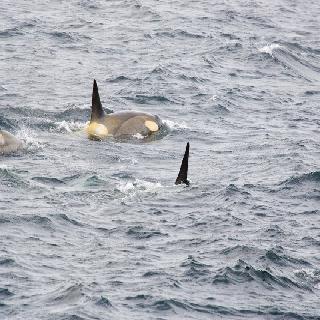}} &
\fcolorbox{species}{species}{\iminatevol{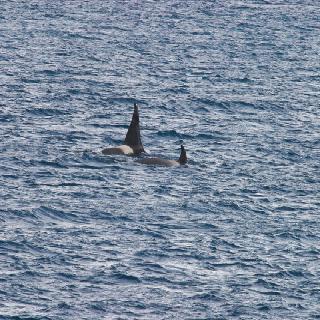}} &
\fcolorbox{species}{species}{\iminatevol{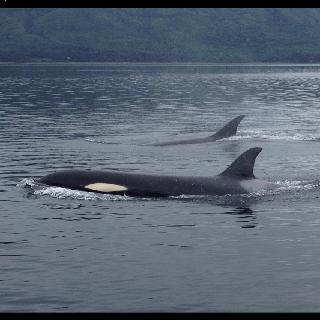}} \\
\bottomrule
\end{tabular}}

    \caption{We compare  \ours and Baseline for different training granularity. We rank the 10 closest images in the iNaturalist-2018 train set for a query image in the test set.
    The ranking is obtained with a cosine similarity on the features space of each of the two approaches. See Table~\ref{tab:copyright_appendix_3} for authors and image copyrights. \label{fig:inat_ranking_evolution_appendix_3}}
    
\end{figure*}

\FloatBarrier
\begin{table*}[p]
\caption{
Author and Creative Commons Copyright notice for images in Figure~\ref{fig:inat_ranking_evolution_appendix_1}.
\label{tab:copyright_appendix_1}}
\scalebox{0.64}{
\begin{minipage}{1.6\linewidth}
\noindent sequeirajluis: \href{http://creativecommons.org/licenses/by-nc/4.0/}{CC BY-NC 4.0}, 101724807574796438015: \href{http://creativecommons.org/licenses/by-nc/4.0/}{CC BY-NC 4.0}, Juan Carlos Pérez Magaña: \href{http://creativecommons.org/licenses/by-nc/4.0/}{CC BY-NC 4.0}, ONG OeBenin: \href{http://creativecommons.org/licenses/by-nc/4.0/}{CC BY-NC 4.0}, Damon Tighe: \href{http://creativecommons.org/licenses/by-nc/4.0/}{CC BY-NC 4.0}, colinmorita: \href{http://creativecommons.org/licenses/by-nc/4.0/}{CC BY-NC 4.0}, Bob Hislop: \href{http://creativecommons.org/licenses/by-nc/4.0/}{CC BY-NC 4.0}, Skyla Slemp: \href{http://creativecommons.org/licenses/by-nc/4.0/}{CC BY-NC 4.0}, Sam Kieschnick: \href{http://creativecommons.org/licenses/by-nc/4.0/}{CC BY-NC 4.0}, Derek Broman: \href{http://creativecommons.org/licenses/by-nc/4.0/}{CC BY-NC 4.0}, Donna Pomeroy: \href{http://creativecommons.org/licenses/by-nc/4.0/}{CC BY-NC 4.0}, dfwuw: \href{http://creativecommons.org/licenses/by-nc/4.0/}{CC BY-NC 4.0}, maractwin: \href{http://creativecommons.org/licenses/by-nc-sa/4.0/}{CC BY-NC-SA 4.0}, pfaucher: \href{http://creativecommons.org/licenses/by-nc/4.0/}{CC BY-NC 4.0}, Ryan Blankenship: \href{http://creativecommons.org/licenses/by-nc/4.0/}{CC BY-NC 4.0}, Gary Chang: \href{http://creativecommons.org/licenses/by-nc-sa/4.0/}{CC BY-NC-SA 4.0}, Linda Jo: \href{http://creativecommons.org/licenses/by-nc/4.0/}{CC BY-NC 4.0}, sequeirajluis: \href{http://creativecommons.org/licenses/by-nc/4.0/}{CC BY-NC 4.0}, pwiedenkeller: \href{http://creativecommons.org/licenses/by-nc/4.0/}{CC BY-NC 4.0}, dfwuw: \href{http://creativecommons.org/licenses/by-nc/4.0/}{CC BY-NC 4.0}, dfwuw: \href{http://creativecommons.org/licenses/by-nc/4.0/}{CC BY-NC 4.0}, sequeirajluis: \href{http://creativecommons.org/licenses/by-nc/4.0/}{CC BY-NC 4.0},  \href{http://creativecommons.org/licenses/by-nc/4.0/}{CC BY-NC 4.0}, Bruno Durand: \href{http://creativecommons.org/licenses/by-nc/4.0/}{CC BY-NC 4.0}, Kyle Jones: \href{http://creativecommons.org/licenses/by-nc/4.0/}{CC BY-NC 4.0}, Jakub Pełka: \href{http://creativecommons.org/licenses/by-nc/4.0/}{CC BY-NC 4.0}, ekoaraba: \href{http://creativecommons.org/licenses/by-nc/4.0/}{CC BY-NC 4.0}, Shane: \href{http://creativecommons.org/licenses/by-nc/4.0/}{CC BY-NC 4.0}, dfwuw: \href{http://creativecommons.org/licenses/by-nc/4.0/}{CC BY-NC 4.0}, Donna Pomeroy: \href{http://creativecommons.org/licenses/by-nc/4.0/}{CC BY-NC 4.0}, gyrrlfalcon: \href{http://creativecommons.org/licenses/by-nc/4.0/}{CC BY-NC 4.0}, dengland81: \href{http://creativecommons.org/licenses/by-nc/4.0/}{CC BY-NC 4.0}, mafzam: \href{http://creativecommons.org/licenses/by-nc/4.0/}{CC BY-NC 4.0}, ONG OeBenin: \href{http://creativecommons.org/licenses/by-nc/4.0/}{CC BY-NC 4.0}, Damon Tighe: \href{http://creativecommons.org/licenses/by-nc/4.0/}{CC BY-NC 4.0}, Cathy Bell: \href{http://creativecommons.org/licenses/by-nc-nd/4.0/}{CC BY-NC-ND 4.0}, ONG OeBenin: \href{http://creativecommons.org/licenses/by-nc/4.0/}{CC BY-NC 4.0}, ONG OeBenin: \href{http://creativecommons.org/licenses/by-nc/4.0/}{CC BY-NC 4.0}, fiddleman: \href{http://creativecommons.org/licenses/by-nc/4.0/}{CC BY-NC 4.0}, ONG OeBenin: \href{http://creativecommons.org/licenses/by-nc/4.0/}{CC BY-NC 4.0}, Brian Gratwicke: \href{http://creativecommons.org/licenses/by/4.0/}{CC BY 4.0}, ONG OeBenin: \href{http://creativecommons.org/licenses/by-nc/4.0/}{CC BY-NC 4.0}, sequeirajluis: \href{http://creativecommons.org/licenses/by-nc/4.0/}{CC BY-NC 4.0}, gyrrlfalcon: \href{http://creativecommons.org/licenses/by-nc/4.0/}{CC BY-NC 4.0}, Simon Kingston: \href{http://creativecommons.org/licenses/by-nc/4.0/}{CC BY-NC 4.0}, Mark Freeman: \href{http://creativecommons.org/licenses/by-nc/4.0/}{CC BY-NC 4.0}, Mike: \href{http://creativecommons.org/licenses/by-nc/4.0/}{CC BY-NC 4.0}, tiyumq: \href{http://creativecommons.org/licenses/by-nc/4.0/}{CC BY-NC 4.0}, Kyle Jones: \href{http://creativecommons.org/licenses/by-nc/4.0/}{CC BY-NC 4.0}, abn: \href{http://creativecommons.org/licenses/by-nc/4.0/}{CC BY-NC 4.0}, James Bailey: \href{http://creativecommons.org/licenses/by-nc/4.0/}{CC BY-NC 4.0}, Allan Finlayson: \href{http://creativecommons.org/licenses/by-nc/4.0/}{CC BY-NC 4.0}, joysglobal: \href{http://creativecommons.org/licenses/by-nc/4.0/}{CC BY-NC 4.0}, sea-kangaroo: \href{http://creativecommons.org/licenses/by-nc-nd/4.0/}{CC BY-NC-ND 4.0}, Ken-ichi Ueda: \href{http://creativecommons.org/licenses/by-nc/4.0/}{CC BY-NC 4.0}, driles5: \href{http://creativecommons.org/licenses/by-nc/4.0/}{CC BY-NC 4.0}, driles5: \href{http://creativecommons.org/licenses/by-nc/4.0/}{CC BY-NC 4.0}, 112692329998402018828: \href{http://creativecommons.org/licenses/by-nc-sa/4.0/}{CC BY-NC-SA 4.0}, Andy Jones: \href{http://creativecommons.org/licenses/by-nc-sa/4.0/}{CC BY-NC-SA 4.0}, almanzacamille: \href{http://creativecommons.org/licenses/by-nc/4.0/}{CC BY-NC 4.0}, 
\href{http://creativecommons.org/licenses/by-nc/4.0/}{CC BY-NC 4.0}, J. Maughn: \href{http://creativecommons.org/licenses/by-nc/4.0/}{CC BY-NC 4.0}, jameseatonbio1020: \href{http://creativecommons.org/licenses/by-nc/4.0/}{CC BY-NC 4.0}, sequeirajluis: \href{http://creativecommons.org/licenses/by-nc/4.0/}{CC BY-NC 4.0}, bio22003jeaton: \href{http://creativecommons.org/licenses/by-nc/4.0/}{CC BY-NC 4.0}, 112692329998402018828: \href{http://creativecommons.org/licenses/by-nc-sa/4.0/}{CC BY-NC-SA 4.0}, Sterling Sheehy: \href{http://creativecommons.org/licenses/by/4.0/}{CC BY 4.0}, Guillermo Debandi: \href{http://creativecommons.org/licenses/by-nc/4.0/}{CC BY-NC 4.0}, James Bailey: \href{http://creativecommons.org/licenses/by-nc/4.0/}{CC BY-NC 4.0}, sliverman: \href{http://creativecommons.org/licenses/by-nc/4.0/}{CC BY-NC 4.0}, Joanne Siderius: \href{http://creativecommons.org/licenses/by-nc/4.0/}{CC BY-NC 4.0}, janaohrner: \href{http://creativecommons.org/licenses/by-nc/4.0/}{CC BY-NC 4.0}, ONG OeBenin: \href{http://creativecommons.org/licenses/by-nc/4.0/}{CC BY-NC 4.0}, mcodellwildlife: \href{http://creativecommons.org/licenses/by-nc/4.0/}{CC BY-NC 4.0}, Johnny Wilson: \href{http://creativecommons.org/licenses/by-nc/4.0/}{CC BY-NC 4.0}, sea-kangaroo: \href{http://creativecommons.org/licenses/by-nc-nd/4.0/}{CC BY-NC-ND 4.0}, Brian Gratwicke: \href{http://creativecommons.org/licenses/by/4.0/}{CC BY 4.0}, 112692329998402018828: \href{http://creativecommons.org/licenses/by-nc-sa/4.0/}{CC BY-NC-SA 4.0}, ONG OeBenin: \href{http://creativecommons.org/licenses/by-nc/4.0/}{CC BY-NC 4.0}, summersilence: \href{http://creativecommons.org/licenses/by-nc/4.0/}{CC BY-NC 4.0}, efarias1: \href{http://creativecommons.org/licenses/by-nc/4.0/}{CC BY-NC 4.0}, redhat: \href{http://creativecommons.org/licenses/by-nc/4.0/}{CC BY-NC 4.0}, Dale Hameister: \href{http://creativecommons.org/licenses/by-nc/4.0/}{CC BY-NC 4.0}, 116916927065934112165: \href{http://creativecommons.org/licenses/by-nc-sa/4.0/}{CC BY-NC-SA 4.0}, sequeirajluis: \href{http://creativecommons.org/licenses/by-nc/4.0/}{CC BY-NC 4.0}, Zac Cota: \href{http://creativecommons.org/licenses/by-nc/4.0/}{CC BY-NC 4.0}, J Brew: \href{http://creativecommons.org/licenses/by-sa/4.0/}{CC BY-SA 4.0}, Cullen Hanks: \href{http://creativecommons.org/licenses/by-nc/4.0/}{CC BY-NC 4.0}, Allan Finlayson: \href{http://creativecommons.org/licenses/by-nc/4.0/}{CC BY-NC 4.0}, dfwuw: \href{http://creativecommons.org/licenses/by-nc/4.0/}{CC BY-NC 4.0}, Mark Freeman: \href{http://creativecommons.org/licenses/by-nc/4.0/}{CC BY-NC 4.0}, mmski303: \href{http://creativecommons.org/licenses/by-nc/4.0/}{CC BY-NC 4.0}, driles5: \href{http://creativecommons.org/licenses/by-nc/4.0/}{CC BY-NC 4.0}, driles5: \href{http://creativecommons.org/licenses/by-nc/4.0/}{CC BY-NC 4.0}, Marcus Garvie: \href{http://creativecommons.org/licenses/by-nc/4.0/}{CC BY-NC 4.0}, ryanubrown: \href{http://creativecommons.org/licenses/by-nc-nd/4.0/}{CC BY-NC-ND 4.0}, efarias1: \href{http://creativecommons.org/licenses/by-nc/4.0/}{CC BY-NC 4.0}, amarena: \href{http://creativecommons.org/licenses/by-nc/4.0/}{CC BY-NC 4.0}, ttempel: \href{http://creativecommons.org/licenses/by-nc/4.0/}{CC BY-NC 4.0}, tnewman: \href{http://creativecommons.org/licenses/by-nc/4.0/}{CC BY-NC 4.0}, Juan Cruzado Cortés: \href{http://creativecommons.org/licenses/by-nc-sa/4.0/}{CC BY-NC-SA 4.0}, lonnyholmes: \href{http://creativecommons.org/licenses/by-nc/4.0/}{CC BY-NC 4.0}, Audrey Kremer: \href{http://creativecommons.org/licenses/by-nc/4.0/}{CC BY-NC 4.0}, texasblonde: \href{http://creativecommons.org/licenses/by-nc/4.0/}{CC BY-NC 4.0}, Dale Hameister: \href{http://creativecommons.org/licenses/by-nc/4.0/}{CC BY-NC 4.0}, sequeirajluis: \href{http://creativecommons.org/licenses/by-nc/4.0/}{CC BY-NC 4.0}, Allan Finlayson: \href{http://creativecommons.org/licenses/by-nc/4.0/}{CC BY-NC 4.0}, pfaucher: \href{http://creativecommons.org/licenses/by-nc/4.0/}{CC BY-NC 4.0}, Edward George: \href{http://creativecommons.org/licenses/by-nc/4.0/}{CC BY-NC 4.0}, Cullen Hanks: \href{http://creativecommons.org/licenses/by-nc/4.0/}{CC BY-NC 4.0}, CK Kelly: \href{http://creativecommons.org/licenses/by/4.0/}{CC BY 4.0}, mustardlypig: \href{http://creativecommons.org/licenses/by-nc/4.0/}{CC BY-NC 4.0}, ONG OeBenin: \href{http://creativecommons.org/licenses/by-nc/4.0/}{CC BY-NC 4.0}, ONG OeBenin: \href{http://creativecommons.org/licenses/by-nc/4.0/}{CC BY-NC 4.0}, ONG OeBenin: \href{http://creativecommons.org/licenses/by-nc/4.0/}{CC BY-NC 4.0}, sequeirajluis: \href{http://creativecommons.org/licenses/by-nc/4.0/}{CC BY-NC 4.0}, sequeirajluis: \href{http://creativecommons.org/licenses/by-nc/4.0/}{CC BY-NC 4.0}, sequeirajluis: \href{http://creativecommons.org/licenses/by-nc/4.0/}{CC BY-NC 4.0}, sequeirajluis: \href{http://creativecommons.org/licenses/by-nc/4.0/}{CC BY-NC 4.0}, sequeirajluis: \href{http://creativecommons.org/licenses/by-nc/4.0/}{CC BY-NC 4.0}, sequeirajluis: \href{http://creativecommons.org/licenses/by-nc/4.0/}{CC BY-NC 4.0}, CK Kelly: \href{http://creativecommons.org/licenses/by/4.0/}{CC BY 4.0}, sequeirajluis: \href{http://creativecommons.org/licenses/by-nc/4.0/}{CC BY-NC 4.0}, sequeirajluis: \href{http://creativecommons.org/licenses/by-nc/4.0/}{CC BY-NC 4.0}, CK Kelly: \href{http://creativecommons.org/licenses/by/4.0/}{CC BY 4.0}, sequeirajluis: \href{http://creativecommons.org/licenses/by-nc/4.0/}{CC BY-NC 4.0}, sequeirajluis: \href{http://creativecommons.org/licenses/by-nc/4.0/}{CC BY-NC 4.0}, nighthawk0083: \href{http://creativecommons.org/licenses/by-nc/4.0/}{CC BY-NC 4.0}, pfaucher: \href{http://creativecommons.org/licenses/by-nc/4.0/}{CC BY-NC 4.0}, Johnny Wilson: \href{http://creativecommons.org/licenses/by-nc/4.0/}{CC BY-NC 4.0}, ncowey: \href{http://creativecommons.org/licenses/by-nc/4.0/}{CC BY-NC 4.0}, Johnny Wilson: \href{http://creativecommons.org/licenses/by-nc/4.0/}{CC BY-NC 4.0}, redhat: \href{http://creativecommons.org/licenses/by-nc/4.0/}{CC BY-NC 4.0}, ONG OeBenin: \href{http://creativecommons.org/licenses/by-nc/4.0/}{CC BY-NC 4.0}, owlentine: \href{http://creativecommons.org/licenses/by-nc/4.0/}{CC BY-NC 4.0}, sequeirajluis: \href{http://creativecommons.org/licenses/by-nc/4.0/}{CC BY-NC 4.0}, Jakub Pełka: \href{http://creativecommons.org/licenses/by-nc/4.0/}{CC BY-NC 4.0}, sequeirajluis: \href{http://creativecommons.org/licenses/by-nc/4.0/}{CC BY-NC 4.0}, sequeirajluis: \href{http://creativecommons.org/licenses/by-nc/4.0/}{CC BY-NC 4.0}, sequeirajluis: \href{http://creativecommons.org/licenses/by-nc/4.0/}{CC BY-NC 4.0}, sequeirajluis: \href{http://creativecommons.org/licenses/by-nc/4.0/}{CC BY-NC 4.0}, sequeirajluis: \href{http://creativecommons.org/licenses/by-nc/4.0/}{CC BY-NC 4.0}, sequeirajluis: \href{http://creativecommons.org/licenses/by-nc/4.0/}{CC BY-NC 4.0}, sequeirajluis: \href{http://creativecommons.org/licenses/by-nc/4.0/}{CC BY-NC 4.0}, sequeirajluis: \href{http://creativecommons.org/licenses/by-nc/4.0/}{CC BY-NC 4.0}, sequeirajluis: \href{http://creativecommons.org/licenses/by-nc/4.0/}{CC BY-NC 4.0}, Robert J. "Bob" Nuelle, Jr. AICEZS: \href{http://creativecommons.org/licenses/by-nc/4.0/}{CC BY-NC 4.0}
\end{minipage}}
\end{table*}

\begin{table*}[p]
\caption{Author and Creative Commons Copyright notice for images in Figure~\ref{fig:inat_ranking_evolution_appendix_2}. 
\label{tab:copyright_appendix_2}}
\scalebox{0.64}{
\begin{minipage}{1.6\linewidth}
\noindent stefanovet: \href{http://creativecommons.org/licenses/by-nc/4.0/}{CC BY-NC 4.0}, Donna Pomeroy: \href{http://creativecommons.org/licenses/by-nc/4.0/}{CC BY-NC 4.0}, selwynq: \href{http://creativecommons.org/licenses/by-nc/4.0/}{CC BY-NC 4.0}, emily-a: \href{http://creativecommons.org/licenses/by-nc/4.0/}{CC BY-NC 4.0}, carlaag: \href{http://creativecommons.org/licenses/by-nc/4.0/}{CC BY-NC 4.0}, carroll: \href{http://creativecommons.org/licenses/by-nc/4.0/}{CC BY-NC 4.0}, kevinhintsa: \href{http://creativecommons.org/licenses/by-nc/4.0/}{CC BY-NC 4.0}, Lindsey Smith: \href{http://creativecommons.org/licenses/by-nc/4.0/}{CC BY-NC 4.0}, kevinhintsa: \href{http://creativecommons.org/licenses/by-nc/4.0/}{CC BY-NC 4.0}, Bob Heitzman: \href{http://creativecommons.org/licenses/by-nc/4.0/}{CC BY-NC 4.0}, rmorgan: \href{http://creativecommons.org/licenses/by-nc/4.0/}{CC BY-NC 4.0}, Giuseppe Cagnetta: \href{http://creativecommons.org/licenses/by-nc/4.0/}{CC BY-NC 4.0}, cwwood: \href{http://creativecommons.org/licenses/by-sa/4.0/}{CC BY-SA 4.0}, Fernando de Juana: \href{http://creativecommons.org/licenses/by-nc/4.0/}{CC BY-NC 4.0}, Chris Evers: \href{http://creativecommons.org/licenses/by-nc/4.0/}{CC BY-NC 4.0}, Marion Zöller: \href{http://creativecommons.org/licenses/by-nc/4.0/}{CC BY-NC 4.0}, Marion Zöller: \href{http://creativecommons.org/licenses/by-nc/4.0/}{CC BY-NC 4.0}, Chuck Sexton: \href{http://creativecommons.org/licenses/by-nc/4.0/}{CC BY-NC 4.0}, Ronald Werson: \href{http://creativecommons.org/licenses/by-nc-nd/4.0/}{CC BY-NC-ND 4.0}, Chris Evers: \href{http://creativecommons.org/licenses/by-nc/4.0/}{CC BY-NC 4.0}, Chris van Swaay: \href{http://creativecommons.org/licenses/by-nc/4.0/}{CC BY-NC 4.0}, stefanovet: \href{http://creativecommons.org/licenses/by-nc/4.0/}{CC BY-NC 4.0}, Richard Barnes: \href{http://creativecommons.org/licenses/by-nc/4.0/}{CC BY-NC 4.0}, Fluff Berger: \href{http://creativecommons.org/licenses/by-sa/4.0/}{CC BY-SA 4.0}, Cheryl Harleston: \href{http://creativecommons.org/licenses/by-nc-sa/4.0/}{CC BY-NC-SA 4.0}, Steven Chong: \href{http://creativecommons.org/licenses/by-nc/4.0/}{CC BY-NC 4.0}, juliayl: \href{http://creativecommons.org/licenses/by-nc/4.0/}{CC BY-NC 4.0}, Kent McFarland: \href{http://creativecommons.org/licenses/by-nc/4.0/}{CC BY-NC 4.0}, Bruno Durand: \href{http://creativecommons.org/licenses/by-nc/4.0/}{CC BY-NC 4.0}, José Belem Hernández Díaz: \href{http://creativecommons.org/licenses/by-nc/4.0/}{CC BY-NC 4.0}, Nolan Eggert: \href{http://creativecommons.org/licenses/by-nc/4.0/}{CC BY-NC 4.0}, Lee Elliott: \href{http://creativecommons.org/licenses/by-nc-sa/4.0/}{CC BY-NC-SA 4.0}, Ronald Werson: \href{http://creativecommons.org/licenses/by-nc-nd/4.0/}{CC BY-NC-ND 4.0}, Giuseppe Cagnetta: \href{http://creativecommons.org/licenses/by-nc/4.0/}{CC BY-NC 4.0}, cwwood: \href{http://creativecommons.org/licenses/by-sa/4.0/}{CC BY-SA 4.0}, Donna Pomeroy: \href{http://creativecommons.org/licenses/by-nc/4.0/}{CC BY-NC 4.0}, martinswarren: \href{http://creativecommons.org/licenses/by-nc/4.0/}{CC BY-NC 4.0}, Fernando de Juana: \href{http://creativecommons.org/licenses/by-nc/4.0/}{CC BY-NC 4.0}, Chris van Swaay: \href{http://creativecommons.org/licenses/by-nc/4.0/}{CC BY-NC 4.0}, Monica Krancevic: \href{http://creativecommons.org/licenses/by-nc/4.0/}{CC BY-NC 4.0}, brentano: \href{http://creativecommons.org/licenses/by-nc/4.0/}{CC BY-NC 4.0}, martinswarren: \href{http://creativecommons.org/licenses/by-nc/4.0/}{CC BY-NC 4.0}, stefanovet: \href{http://creativecommons.org/licenses/by-nc/4.0/}{CC BY-NC 4.0}, Mark Rosenstein: \href{http://creativecommons.org/licenses/by-nc-sa/4.0/}{CC BY-NC-SA 4.0}, Mark Rosenstein: \href{http://creativecommons.org/licenses/by-nc-sa/4.0/}{CC BY-NC-SA 4.0}, Mark Nenadov: \href{http://creativecommons.org/licenses/by-nc/4.0/}{CC BY-NC 4.0}, Robin Agarwal: \href{http://creativecommons.org/licenses/by-nc/4.0/}{CC BY-NC 4.0}, Bryan: \href{http://creativecommons.org/licenses/by-nc/4.0/}{CC BY-NC 4.0}, Ronald Werson: \href{http://creativecommons.org/licenses/by-nc-nd/4.0/}{CC BY-NC-ND 4.0}, Chris van Swaay: \href{http://creativecommons.org/licenses/by-nc/4.0/}{CC BY-NC 4.0}, Donna Pomeroy: \href{http://creativecommons.org/licenses/by-nc/4.0/}{CC BY-NC 4.0}, Mark Rosenstein: \href{http://creativecommons.org/licenses/by-nc-sa/4.0/}{CC BY-NC-SA 4.0}, Monica Krancevic: \href{http://creativecommons.org/licenses/by-nc/4.0/}{CC BY-NC 4.0}, Chris Evers: \href{http://creativecommons.org/licenses/by-nc/4.0/}{CC BY-NC 4.0}, Ronald Werson: \href{http://creativecommons.org/licenses/by-nc-nd/4.0/}{CC BY-NC-ND 4.0}, Don Loarie: \href{http://creativecommons.org/licenses/by/4.0/}{CC BY 4.0}, Fernando de Juana: \href{http://creativecommons.org/licenses/by-nc/4.0/}{CC BY-NC 4.0}, Chris Evers: \href{http://creativecommons.org/licenses/by-nc/4.0/}{CC BY-NC 4.0}, Donna Pomeroy: \href{http://creativecommons.org/licenses/by-nc/4.0/}{CC BY-NC 4.0}, Donna Pomeroy: \href{http://creativecommons.org/licenses/by-nc/4.0/}{CC BY-NC 4.0}, Donna Pomeroy: \href{http://creativecommons.org/licenses/by-nc/4.0/}{CC BY-NC 4.0}, Chris van Swaay: \href{http://creativecommons.org/licenses/by-nc/4.0/}{CC BY-NC 4.0}, cwwood: \href{http://creativecommons.org/licenses/by-sa/4.0/}{CC BY-SA 4.0}, stefanovet: \href{http://creativecommons.org/licenses/by-nc/4.0/}{CC BY-NC 4.0}, Chris Evers: \href{http://creativecommons.org/licenses/by-nc/4.0/}{CC BY-NC 4.0}, J. Maughn: \href{http://creativecommons.org/licenses/by-nc/4.0/}{CC BY-NC 4.0}, Liam O'Brien: \href{http://creativecommons.org/licenses/by-nc/4.0/}{CC BY-NC 4.0}, Greg Lasley: \href{http://creativecommons.org/licenses/by-nc/4.0/}{CC BY-NC 4.0}, Steven Bach: \href{http://creativecommons.org/licenses/by-nc-sa/4.0/}{CC BY-NC-SA 4.0}, Donna Pomeroy: \href{http://creativecommons.org/licenses/by-nc/4.0/}{CC BY-NC 4.0}, Judith Lopez Sikora: \href{http://creativecommons.org/licenses/by-nc/4.0/}{CC BY-NC 4.0}, Chris van Swaay: \href{http://creativecommons.org/licenses/by-nc/4.0/}{CC BY-NC 4.0}, J. Maughn: \href{http://creativecommons.org/licenses/by-nc/4.0/}{CC BY-NC 4.0}, 112692329998402018828: \href{http://creativecommons.org/licenses/by-nc-sa/4.0/}{CC BY-NC-SA 4.0}, martinswarren: \href{http://creativecommons.org/licenses/by-nc/4.0/}{CC BY-NC 4.0}, Giuseppe Cagnetta: \href{http://creativecommons.org/licenses/by-nc/4.0/}{CC BY-NC 4.0}, Chris Evers: \href{http://creativecommons.org/licenses/by-nc/4.0/}{CC BY-NC 4.0}, Chris van Swaay: \href{http://creativecommons.org/licenses/by-nc/4.0/}{CC BY-NC 4.0}, Donna Pomeroy: \href{http://creativecommons.org/licenses/by-nc/4.0/}{CC BY-NC 4.0}, Ronald Werson: \href{http://creativecommons.org/licenses/by-nc-nd/4.0/}{CC BY-NC-ND 4.0}, Donna Pomeroy: \href{http://creativecommons.org/licenses/by-nc/4.0/}{CC BY-NC 4.0}, Monica Krancevic: \href{http://creativecommons.org/licenses/by-nc/4.0/}{CC BY-NC 4.0}, smwhite: \href{http://creativecommons.org/licenses/by-nc/4.0/}{CC BY-NC 4.0}, dwest: \href{http://creativecommons.org/licenses/by-nc/4.0/}{CC BY-NC 4.0}, stefanovet: \href{http://creativecommons.org/licenses/by-nc/4.0/}{CC BY-NC 4.0}, Donna Pomeroy: \href{http://creativecommons.org/licenses/by-nc/4.0/}{CC BY-NC 4.0}, greglasley: \href{http://creativecommons.org/licenses/by-nc/4.0/}{CC BY-NC 4.0}, Ronald Werson: \href{http://creativecommons.org/licenses/by-nc-nd/4.0/}{CC BY-NC-ND 4.0}, Donna Pomeroy: \href{http://creativecommons.org/licenses/by-nc/4.0/}{CC BY-NC 4.0}, Donna Pomeroy: \href{http://creativecommons.org/licenses/by-nc/4.0/}{CC BY-NC 4.0}, Donna Pomeroy: \href{http://creativecommons.org/licenses/by-nc/4.0/}{CC BY-NC 4.0}, Chris van Swaay: \href{http://creativecommons.org/licenses/by-nc/4.0/}{CC BY-NC 4.0}, Donna Pomeroy: \href{http://creativecommons.org/licenses/by-nc/4.0/}{CC BY-NC 4.0}, Chris van Swaay: \href{http://creativecommons.org/licenses/by-nc/4.0/}{CC BY-NC 4.0}, beschwar: \href{http://creativecommons.org/licenses/by-nc/4.0/}{CC BY-NC 4.0}, Chris van Swaay: \href{http://creativecommons.org/licenses/by-nc/4.0/}{CC BY-NC 4.0}, Donna Pomeroy: \href{http://creativecommons.org/licenses/by-nc/4.0/}{CC BY-NC 4.0}, Donna Pomeroy: \href{http://creativecommons.org/licenses/by-nc/4.0/}{CC BY-NC 4.0}, Fernando de Juana: \href{http://creativecommons.org/licenses/by-nc/4.0/}{CC BY-NC 4.0}, Marion Zöller: \href{http://creativecommons.org/licenses/by-nc/4.0/}{CC BY-NC 4.0}, Chuck Sexton: \href{http://creativecommons.org/licenses/by-nc/4.0/}{CC BY-NC 4.0}, Chris van Swaay: \href{http://creativecommons.org/licenses/by-nc/4.0/}{CC BY-NC 4.0}, martinswarren: \href{http://creativecommons.org/licenses/by-nc/4.0/}{CC BY-NC 4.0}, Kathy Richardson: \href{http://creativecommons.org/licenses/by-nc/4.0/}{CC BY-NC 4.0}, cwwood: \href{http://creativecommons.org/licenses/by-sa/4.0/}{CC BY-SA 4.0}, stefanovet: \href{http://creativecommons.org/licenses/by-nc/4.0/}{CC BY-NC 4.0}, Ronald Werson: \href{http://creativecommons.org/licenses/by-nc-nd/4.0/}{CC BY-NC-ND 4.0}, Donna Pomeroy: \href{http://creativecommons.org/licenses/by-nc/4.0/}{CC BY-NC 4.0}, Donna Pomeroy: \href{http://creativecommons.org/licenses/by-nc/4.0/}{CC BY-NC 4.0}, Chris van Swaay: \href{http://creativecommons.org/licenses/by-nc/4.0/}{CC BY-NC 4.0}, Donna Pomeroy: \href{http://creativecommons.org/licenses/by-nc/4.0/}{CC BY-NC 4.0}, Fernando de Juana: \href{http://creativecommons.org/licenses/by-nc/4.0/}{CC BY-NC 4.0}, Ronald Werson: \href{http://creativecommons.org/licenses/by-nc-nd/4.0/}{CC BY-NC-ND 4.0}, martinswarren: \href{http://creativecommons.org/licenses/by-nc/4.0/}{CC BY-NC 4.0}, Donna Pomeroy: \href{http://creativecommons.org/licenses/by-nc/4.0/}{CC BY-NC 4.0}, Marion Zöller: \href{http://creativecommons.org/licenses/by-nc/4.0/}{CC BY-NC 4.0}, Ronald Werson: \href{http://creativecommons.org/licenses/by-nc-nd/4.0/}{CC BY-NC-ND 4.0}, Chris van Swaay: \href{http://creativecommons.org/licenses/by-nc/4.0/}{CC BY-NC 4.0}, Giuseppe Cagnetta: \href{http://creativecommons.org/licenses/by-nc/4.0/}{CC BY-NC 4.0}, martinswarren: \href{http://creativecommons.org/licenses/by-nc/4.0/}{CC BY-NC 4.0}, Robin Agarwal: \href{http://creativecommons.org/licenses/by-nc/4.0/}{CC BY-NC 4.0}, Giuseppe Cagnetta: \href{http://creativecommons.org/licenses/by-nc/4.0/}{CC BY-NC 4.0}, Marion Zöller: \href{http://creativecommons.org/licenses/by-nc/4.0/}{CC BY-NC 4.0}, Ronald Werson: \href{http://creativecommons.org/licenses/by-nc-nd/4.0/}{CC BY-NC-ND 4.0}, Donna Pomeroy: \href{http://creativecommons.org/licenses/by-nc/4.0/}{CC BY-NC 4.0}, martinswarren: \href{http://creativecommons.org/licenses/by-nc/4.0/}{CC BY-NC 4.0}, stefanovet: \href{http://creativecommons.org/licenses/by-nc/4.0/}{CC BY-NC 4.0}, Ronald Werson: \href{http://creativecommons.org/licenses/by-nc-nd/4.0/}{CC BY-NC-ND 4.0}, Ronald Werson: \href{http://creativecommons.org/licenses/by-nc-nd/4.0/}{CC BY-NC-ND 4.0}, martinswarren: \href{http://creativecommons.org/licenses/by-nc/4.0/}{CC BY-NC 4.0}, Giuseppe Cagnetta: \href{http://creativecommons.org/licenses/by-nc/4.0/}{CC BY-NC 4.0}, Fernando de Juana: \href{http://creativecommons.org/licenses/by-nc/4.0/}{CC BY-NC 4.0}, Giuseppe Cagnetta: \href{http://creativecommons.org/licenses/by-nc/4.0/}{CC BY-NC 4.0}, rada: \href{http://creativecommons.org/licenses/by-nc/4.0/}{CC BY-NC 4.0}, Philip Mark Osso: \href{http://creativecommons.org/licenses/by-nc/4.0/}{CC BY-NC 4.0}, Fernando de Juana: \href{http://creativecommons.org/licenses/by-nc/4.0/}{CC BY-NC 4.0}, Giuseppe Cagnetta: \href{http://creativecommons.org/licenses/by-nc/4.0/}{CC BY-NC 4.0}, martinswarren: \href{http://creativecommons.org/licenses/by-nc/4.0/}{CC BY-NC 4.0}, Giuseppe Cagnetta: \href{http://creativecommons.org/licenses/by-nc/4.0/}{CC BY-NC 4.0}, Chris van Swaay: \href{http://creativecommons.org/licenses/by-nc/4.0/}{CC BY-NC 4.0}, Chris van Swaay: \href{http://creativecommons.org/licenses/by-nc/4.0/}{CC BY-NC 4.0}, Giuseppe Cagnetta: \href{http://creativecommons.org/licenses/by-nc/4.0/}{CC BY-NC 4.0}, rada: \href{http://creativecommons.org/licenses/by-nc/4.0/}{CC BY-NC 4.0}, martinswarren: \href{http://creativecommons.org/licenses/by-nc/4.0/}{CC BY-NC 4.0}, martinswarren: \href{http://creativecommons.org/licenses/by-nc/4.0/}{CC BY-NC 4.0}, martinswarren: \href{http://creativecommons.org/licenses/by-nc/4.0/}{CC BY-NC 4.0}, Fernando de Juana: \href{http://creativecommons.org/licenses/by-nc/4.0/}{CC BY-NC 4.0}. 
\end{minipage}}
\end{table*}

\begin{table*}[p]
\caption{Author and Creative Commons Copyright notice for images in Figure~\ref{fig:inat_ranking_evolution_appendix_3}. 
\label{tab:copyright_appendix_3}}
\scalebox{0.64}{
\begin{minipage}{1.6\linewidth}
\noindent slsfirefight: \href{http://creativecommons.org/licenses/by-nc/4.0/}{CC BY-NC 4.0}, J. Maughn: \href{http://creativecommons.org/licenses/by-nc/4.0/}{CC BY-NC 4.0}, Tim Hite: \href{http://creativecommons.org/licenses/by/4.0/}{CC BY 4.0}, Mikael Behrens: \href{http://creativecommons.org/licenses/by-nc/4.0/}{CC BY-NC 4.0}, colinmorita: \href{http://creativecommons.org/licenses/by-nc/4.0/}{CC BY-NC 4.0}, Mary Joyce: \href{http://creativecommons.org/licenses/by-nc/4.0/}{CC BY-NC 4.0}, Donna Pomeroy: \href{http://creativecommons.org/licenses/by-nc/4.0/}{CC BY-NC 4.0}, icosahedron: \href{http://creativecommons.org/licenses/by/4.0/}{CC BY 4.0}, Donna Pomeroy: \href{http://creativecommons.org/licenses/by-nc/4.0/}{CC BY-NC 4.0}, tnewman: \href{http://creativecommons.org/licenses/by-nc/4.0/}{CC BY-NC 4.0}, phylocode: \href{http://creativecommons.org/licenses/by-nc/4.0/}{CC BY-NC 4.0}, Marisa or Robin Agarwal: \href{http://creativecommons.org/licenses/by-nc/4.0/}{CC BY-NC 4.0}, Donna Pomeroy: \href{http://creativecommons.org/licenses/by-nc/4.0/}{CC BY-NC 4.0}, Donna Pomeroy: \href{http://creativecommons.org/licenses/by-nc/4.0/}{CC BY-NC 4.0}, Tom Benson: \href{http://creativecommons.org/licenses/by-nc-nd/4.0/}{CC BY-NC-ND 4.0}, tam topes: \href{http://creativecommons.org/licenses/by-nc/4.0/}{CC BY-NC 4.0}, James Maughn: \href{http://creativecommons.org/licenses/by-nc/4.0/}{CC BY-NC 4.0}, Mark Rosenstein: \href{http://creativecommons.org/licenses/by-nc-sa/4.0/}{CC BY-NC-SA 4.0}, Marisa or Robin Agarwal: \href{http://creativecommons.org/licenses/by-nc/4.0/}{CC BY-NC 4.0}, Donna Pomeroy: \href{http://creativecommons.org/licenses/by-nc/4.0/}{CC BY-NC 4.0}, Robin Agarwal: \href{http://creativecommons.org/licenses/by-nc/4.0/}{CC BY-NC 4.0}, slsfirefight: \href{http://creativecommons.org/licenses/by-nc/4.0/}{CC BY-NC 4.0}, Thomas Koffel: \href{http://creativecommons.org/licenses/by-nc/4.0/}{CC BY-NC 4.0}, Chris Evers: \href{http://creativecommons.org/licenses/by-nc/4.0/}{CC BY-NC 4.0}, jaliya: \href{http://creativecommons.org/licenses/by-nc/4.0/}{CC BY-NC 4.0}, Chris Evers: \href{http://creativecommons.org/licenses/by-nc/4.0/}{CC BY-NC 4.0}, Victor W Fazio III: \href{http://creativecommons.org/licenses/by-nc-nd/4.0/}{CC BY-NC-ND 4.0}, Chris Evers: \href{http://creativecommons.org/licenses/by-nc/4.0/}{CC BY-NC 4.0}, J. Maughn: \href{http://creativecommons.org/licenses/by-nc/4.0/}{CC BY-NC 4.0}, Andrew Cannizzaro: \href{http://creativecommons.org/licenses/by/4.0/}{CC BY 4.0}, 116916927065934112165: \href{http://creativecommons.org/licenses/by-nc-nd/4.0/}{CC BY-NC-ND 4.0}, gyrrlfalcon: \href{http://creativecommons.org/licenses/by-nc/4.0/}{CC BY-NC 4.0}, kolasafamily: \href{http://creativecommons.org/licenses/by-nc/4.0/}{CC BY-NC 4.0}, summermule: \href{http://creativecommons.org/licenses/by-nc/4.0/}{CC BY-NC 4.0}, Donna Pomeroy: \href{http://creativecommons.org/licenses/by-nc/4.0/}{CC BY-NC 4.0}, Donna Pomeroy: \href{http://creativecommons.org/licenses/by-nc/4.0/}{CC BY-NC 4.0}, Robin Agarwal: \href{http://creativecommons.org/licenses/by-nc/4.0/}{CC BY-NC 4.0}, Donna Pomeroy: \href{http://creativecommons.org/licenses/by-nc/4.0/}{CC BY-NC 4.0}, Robin Agarwal: \href{http://creativecommons.org/licenses/by-nc/4.0/}{CC BY-NC 4.0}, Robin Agarwal: \href{http://creativecommons.org/licenses/by-nc/4.0/}{CC BY-NC 4.0}, David R: \href{http://creativecommons.org/licenses/by-nc-nd/4.0/}{CC BY-NC-ND 4.0}, summermule: \href{http://creativecommons.org/licenses/by-nc/4.0/}{CC BY-NC 4.0}, slsfirefight: \href{http://creativecommons.org/licenses/by-nc/4.0/}{CC BY-NC 4.0}, Donna Pomeroy: \href{http://creativecommons.org/licenses/by-nc/4.0/}{CC BY-NC 4.0}, petecorradino: \href{http://creativecommons.org/licenses/by-nc/4.0/}{CC BY-NC 4.0}, Mike Leveille: \href{http://creativecommons.org/licenses/by-nc/4.0/}{CC BY-NC 4.0}, greglasley: \href{http://creativecommons.org/licenses/by-nc/4.0/}{CC BY-NC 4.0}, tegmort: \href{http://creativecommons.org/licenses/by-nc/4.0/}{CC BY-NC 4.0}, Donna Pomeroy: \href{http://creativecommons.org/licenses/by-nc/4.0/}{CC BY-NC 4.0}, Tom Benson: \href{http://creativecommons.org/licenses/by-nc-nd/4.0/}{CC BY-NC-ND 4.0}, flyfisherking: \href{http://creativecommons.org/licenses/by-nc/4.0/}{CC BY-NC 4.0}, Jean-Lou Justine: \href{http://creativecommons.org/licenses/by/4.0/}{CC BY 4.0}, Judith Lopez Sikora: \href{http://creativecommons.org/licenses/by-nc/4.0/}{CC BY-NC 4.0}, kolasafamily: \href{http://creativecommons.org/licenses/by-nc/4.0/}{CC BY-NC 4.0}, nudibranchmom: \href{http://creativecommons.org/licenses/by-nc/4.0/}{CC BY-NC 4.0}, Robin Agarwal: \href{http://creativecommons.org/licenses/by-nc/4.0/}{CC BY-NC 4.0}, R.J. Adams: \href{http://creativecommons.org/licenses/by-nc/4.0/}{CC BY-NC 4.0}, monicamares: \href{http://creativecommons.org/licenses/by-nc/4.0/}{CC BY-NC 4.0}, Donna Pomeroy: \href{http://creativecommons.org/licenses/by-nc/4.0/}{CC BY-NC 4.0}, nudibranchmom: \href{http://creativecommons.org/licenses/by-nc/4.0/}{CC BY-NC 4.0}, Donna Pomeroy: \href{http://creativecommons.org/licenses/by-nc/4.0/}{CC BY-NC 4.0}, Mike Leveille: \href{http://creativecommons.org/licenses/by-nc/4.0/}{CC BY-NC 4.0}, nudibranchmom: \href{http://creativecommons.org/licenses/by-nc/4.0/}{CC BY-NC 4.0}, slsfirefight: \href{http://creativecommons.org/licenses/by-nc/4.0/}{CC BY-NC 4.0}, J. Maughn: \href{http://creativecommons.org/licenses/by-nc/4.0/}{CC BY-NC 4.0}, Donna Pomeroy: \href{http://creativecommons.org/licenses/by-nc/4.0/}{CC BY-NC 4.0}, Tom Benson: \href{http://creativecommons.org/licenses/by-nc-nd/4.0/}{CC BY-NC-ND 4.0}, JJ Johnson: \href{http://creativecommons.org/licenses/by-nc/4.0/}{CC BY-NC 4.0}, James Maughn: \href{http://creativecommons.org/licenses/by-nc/4.0/}{CC BY-NC 4.0}, petecorradino: \href{http://creativecommons.org/licenses/by-nc/4.0/}{CC BY-NC 4.0}, summermule: \href{http://creativecommons.org/licenses/by-nc/4.0/}{CC BY-NC 4.0}, pfaucher: \href{http://creativecommons.org/licenses/by-nc/4.0/}{CC BY-NC 4.0}, Amy: \href{http://creativecommons.org/licenses/by-nc/4.0/}{CC BY-NC 4.0}, enzedfred: \href{http://creativecommons.org/licenses/by-nc/4.0/}{CC BY-NC 4.0}, Donna Pomeroy: \href{http://creativecommons.org/licenses/by-nc/4.0/}{CC BY-NC 4.0}, BJ Stacey: \href{http://creativecommons.org/licenses/by-nc/4.0/}{CC BY-NC 4.0}, Donna Pomeroy: \href{http://creativecommons.org/licenses/by-nc/4.0/}{CC BY-NC 4.0}, Donna Pomeroy: \href{http://creativecommons.org/licenses/by-nc/4.0/}{CC BY-NC 4.0}, Robin Agarwal: \href{http://creativecommons.org/licenses/by-nc/4.0/}{CC BY-NC 4.0}, Robin Agarwal: \href{http://creativecommons.org/licenses/by-nc/4.0/}{CC BY-NC 4.0}, Ken-ichi Ueda: \href{http://creativecommons.org/licenses/by-nc/4.0/}{CC BY-NC 4.0}, Robin Agarwal: \href{http://creativecommons.org/licenses/by-nc/4.0/}{CC BY-NC 4.0}, Robin Agarwal: \href{http://creativecommons.org/licenses/by-nc/4.0/}{CC BY-NC 4.0}, Brian Gratwicke: \href{http://creativecommons.org/licenses/by/4.0/}{CC BY 4.0}, slsfirefight: \href{http://creativecommons.org/licenses/by-nc/4.0/}{CC BY-NC 4.0}, Robin Agarwal: \href{http://creativecommons.org/licenses/by-nc/4.0/}{CC BY-NC 4.0}, Robin Agarwal: \href{http://creativecommons.org/licenses/by-nc/4.0/}{CC BY-NC 4.0}, Donna Pomeroy: \href{http://creativecommons.org/licenses/by-nc/4.0/}{CC BY-NC 4.0}, summermule: \href{http://creativecommons.org/licenses/by-nc/4.0/}{CC BY-NC 4.0}, KK: \href{http://creativecommons.org/licenses/by-nc/4.0/}{CC BY-NC 4.0}, John Karges: \href{http://creativecommons.org/licenses/by-nc/4.0/}{CC BY-NC 4.0}, Donna Pomeroy: \href{http://creativecommons.org/licenses/by-nc/4.0/}{CC BY-NC 4.0}, Javier Solís: \href{http://creativecommons.org/licenses/by-nc/4.0/}{CC BY-NC 4.0}, Donna Pomeroy: \href{http://creativecommons.org/licenses/by-nc/4.0/}{CC BY-NC 4.0}, Mike Leveille: \href{http://creativecommons.org/licenses/by-nc/4.0/}{CC BY-NC 4.0}, Ken-ichi Ueda: \href{http://creativecommons.org/licenses/by-nc/4.0/}{CC BY-NC 4.0}, J. Maughn: \href{http://creativecommons.org/licenses/by-nc/4.0/}{CC BY-NC 4.0}, Robin Agarwal: \href{http://creativecommons.org/licenses/by-nc/4.0/}{CC BY-NC 4.0}, David J Barton: \href{http://creativecommons.org/licenses/by-nc/4.0/}{CC BY-NC 4.0}, James Maughn: \href{http://creativecommons.org/licenses/by-nc/4.0/}{CC BY-NC 4.0}, greglasley: \href{http://creativecommons.org/licenses/by-nc/4.0/}{CC BY-NC 4.0}, Donna Pomeroy: \href{http://creativecommons.org/licenses/by-nc/4.0/}{CC BY-NC 4.0}, Robin Agarwal: \href{http://creativecommons.org/licenses/by-nc/4.0/}{CC BY-NC 4.0}, James Maughn: \href{http://creativecommons.org/licenses/by-nc/4.0/}{CC BY-NC 4.0}, Donna Pomeroy: \href{http://creativecommons.org/licenses/by-nc/4.0/}{CC BY-NC 4.0}, slsfirefight: \href{http://creativecommons.org/licenses/by-nc/4.0/}{CC BY-NC 4.0}, Donna Pomeroy: \href{http://creativecommons.org/licenses/by-nc/4.0/}{CC BY-NC 4.0}, John Karges: \href{http://creativecommons.org/licenses/by-nc/4.0/}{CC BY-NC 4.0}, 104623964081378888743: \href{http://creativecommons.org/licenses/by-nc-nd/4.0/}{CC BY-NC-ND 4.0}, Judith Lopez Sikora: \href{http://creativecommons.org/licenses/by-nc/4.0/}{CC BY-NC 4.0}, Amy: \href{http://creativecommons.org/licenses/by-nc/4.0/}{CC BY-NC 4.0}, Marisa or Robin Agarwal: \href{http://creativecommons.org/licenses/by-nc/4.0/}{CC BY-NC 4.0}, summermule: \href{http://creativecommons.org/licenses/by-nc/4.0/}{CC BY-NC 4.0}, Donna Pomeroy: \href{http://creativecommons.org/licenses/by-nc/4.0/}{CC BY-NC 4.0}, Jennifer Rycenga: \href{http://creativecommons.org/licenses/by-nc/4.0/}{CC BY-NC 4.0}, David J Barton: \href{http://creativecommons.org/licenses/by-nc/4.0/}{CC BY-NC 4.0}, thylacine: \href{http://creativecommons.org/licenses/by-nc/4.0/}{CC BY-NC 4.0}, greglasley: \href{http://creativecommons.org/licenses/by-nc/4.0/}{CC BY-NC 4.0}, J. Maughn: \href{http://creativecommons.org/licenses/by-nc/4.0/}{CC BY-NC 4.0}, Javier Solís: \href{http://creativecommons.org/licenses/by-nc/4.0/}{CC BY-NC 4.0}, redhat: \href{http://creativecommons.org/licenses/by-nc/4.0/}{CC BY-NC 4.0}, timputtre: \href{http://creativecommons.org/licenses/by-nc/4.0/}{CC BY-NC 4.0}, icosahedron: \href{http://creativecommons.org/licenses/by/4.0/}{CC BY 4.0}, rbbrummitt: \href{http://creativecommons.org/licenses/by-nc/4.0/}{CC BY-NC 4.0}, icosahedron: \href{http://creativecommons.org/licenses/by/4.0/}{CC BY 4.0}, David J Barton: \href{http://creativecommons.org/licenses/by-nc/4.0/}{CC BY-NC 4.0}, slsfirefight: \href{http://creativecommons.org/licenses/by-nc/4.0/}{CC BY-NC 4.0}, 104623964081378888743: \href{http://creativecommons.org/licenses/by-nc-nd/4.0/}{CC BY-NC-ND 4.0}, Ken-ichi Ueda: \href{http://creativecommons.org/licenses/by-nc-sa/4.0/}{CC BY-NC-SA 4.0}, Donna Pomeroy: \href{http://creativecommons.org/licenses/by-nc/4.0/}{CC BY-NC 4.0}, sakuraisomi: \href{http://creativecommons.org/licenses/by-nc/4.0/}{CC BY-NC 4.0}, Donna Pomeroy: \href{http://creativecommons.org/licenses/by-nc/4.0/}{CC BY-NC 4.0}, Donna Pomeroy: \href{http://creativecommons.org/licenses/by-nc/4.0/}{CC BY-NC 4.0}, J. Maughn: \href{http://creativecommons.org/licenses/by-nc/4.0/}{CC BY-NC 4.0}, kestrel: \href{http://creativecommons.org/licenses/by-nc/4.0/}{CC BY-NC 4.0}, BJ Stacey: \href{http://creativecommons.org/licenses/by-nc/4.0/}{CC BY-NC 4.0}, summermule: \href{http://creativecommons.org/licenses/by-nc/4.0/}{CC BY-NC 4.0}, thylacine: \href{http://creativecommons.org/licenses/by-nc/4.0/}{CC BY-NC 4.0}, icosahedron: \href{http://creativecommons.org/licenses/by/4.0/}{CC BY 4.0}, KK: \href{http://creativecommons.org/licenses/by-nc/4.0/}{CC BY-NC 4.0}, James Maughn: \href{http://creativecommons.org/licenses/by-nc/4.0/}{CC BY-NC 4.0}, Javier Solís: \href{http://creativecommons.org/licenses/by-nc/4.0/}{CC BY-NC 4.0}, rbbrummitt: \href{http://creativecommons.org/licenses/by-nc/4.0/}{CC BY-NC 4.0}, J. Maughn: \href{http://creativecommons.org/licenses/by-nc/4.0/}{CC BY-NC 4.0}, greglasley: \href{http://creativecommons.org/licenses/by-nc/4.0/}{CC BY-NC 4.0}, greglasley: \href{http://creativecommons.org/licenses/by-nc/4.0/}{CC BY-NC 4.0}, timputtre: \href{http://creativecommons.org/licenses/by-nc/4.0/}{CC BY-NC 4.0}.
\end{minipage}}
\end{table*}

\end{document}